\documentclass[jair-,twoside,11pt,theapa]{article}
\usepackage{jair-, theapa, rawfonts}

\usepackage{amsmath}
\usepackage{color}
\usepackage{url}
\usepackage{graphicx}
\usepackage{subfigure}

\usepackage[vlined, ruled, boxed]{algorithm2e}
\usepackage{multirow}

\usepackage{tikz}
\usetikzlibrary{shapes,arrows,shadows}

\tikzstyle{decision} = [diamond, draw, fill=blue!20, text width=4.5em, text badly centered, node distance=3cm, inner sep=0pt]
\tikzstyle{block} = [rectangle, draw, fill=blue!20, text width=5em, text centered, rounded corners, minimum height=4em]
\tikzstyle{line} = [draw, -latex']
\tikzstyle{cloud} = [draw, ellipse,fill=red!20, node distance=3cm, minimum height=2em]

\makeatletter
\newcommand\footnoteref[1]{\protected@xdef\@thefnmark{\ref{#1}}\@footnotemark}
\makeatother

\newcommand{\Aindent}{\leavevmode\hphantom{\textbf{A: }}}
\newcommand{\aindent}{\leavevmode\hphantom{(a) }}
\newcommand\tinyPut[1]{\fontfamily{put}{\selectfont{\tiny{#1}}}}

\ShortHeadings{Understand, Compose and Respond}
{Vatashsky \& Ullman}
\firstpageno{1}

\begin{document}

\title{Understand, Compose and Respond - Answering Visual Questions by a Composition of Abstract Procedures}

\author{\name Ben Zion Vatashsky \email ben-zion.vatashsky@weizmann.ac.il \\
       \name Shimon Ullman \email shimon.ullman@weizmann.ac.il \\
       \addr Weizmann Institute of Science,
       234 Herzl St.
       PO Box 26, \\
       Rehovot, 7610001
       Israel}


\maketitle

\begin{abstract}

An image related question defines a specific visual task that is required in order to produce an appropriate answer. The answer may depend on a minor detail in the image and require complex reasoning and use of prior knowledge. When humans perform this task, they are able to do it in a flexible and robust manner, integrating modularly any novel visual capability with diverse options for various elaborations of the task. In contrast, current approaches to solve this problem by a machine are based on casting the problem as an end-to-end learning problem, which lacks such abilities.

We present a different approach, inspired by the aforementioned human capabilities. The approach is based on the compositional structure of the question. The underlying idea is that a question has an abstract representation based on its structure, which is compositional in nature. The question can consequently be answered by a composition of procedures corresponding to its substructures. The basic elements of the representation are logical patterns, which are put together to represent the question. These patterns include a parametric representation for object classes, properties and relations. Each basic pattern is mapped into a basic procedure that includes meaningful visual tasks, and the patterns are composed to produce the overall answering procedure.

The UnCoRd (Understand Compose and Respond) system, based on this approach, integrates existing detection and classification schemes for a set of object classes, properties and relations. These schemes are incorporated in a modular manner, typical also to human vision. The logical composition of real visual tasks allows using meaningful intermediate results to elaborate the answer (e.g. reasoning) and provide corrections and alternatives when answers are negative. In addition, an external knowledge base is also integrated into the process, to supply common-knowledge information that may be required to understand the question and produce an answer. We performed a qualitative analysis of the system, which demonstrates its representation capabilities and provide suggestions for future developments.

\end{abstract}

\section{Introduction}
\label{Introduction}

Human ability to answer a question related to an image is remarkable in several ways. Given a single image, a large number of different questions can be answered about it. Answering these questions may require the detection and analysis of subtle, non-salient cues. Prior information and data obtained through experience are also incorporated into the process, to enable answering the question, which may be highly complex. The answering process itself is open to reasoning, allowing for example elaborations on the answer, or explaining how it was reached. In the last few years, the problem of image question-answering by a machine was addressed by many studies \cite{teney2017tips,pandhre2017survey,wu2016visual,Kafle2016vqasurvey}, mostly by treating the problem as an end-to-end multi-class training problem. In these methods, image representation is based on the last convolutional layer of a pre-trained Convolutional Neural Network (CNN) \cite{hochreiter1997long}. It is fused with the question features (mostly represented using a Recurrent Neural Network (RNN), e.g. LSTM \cite{lecun1998gradient}) to generate embedded features that are used to predict the answer from common answers of a training set.

Though current existing methods show statistical success on the trained datasets (e.g. VQA  \cite{VQA}, VQA v2 \cite{balanced_vqa_v2}, CLEVR \cite{johnson2016clevr}), they do so by exploiting biases of the questions and the specific datasets \cite{xu2017can,agrawal2016analyzing}. The human abilities and understanding mentioned above are missing from these methods. Casting the problem into an end-to-end multi-class problem, makes it practically impossible to obtain ``human like'' understanding of the question and the answering process itself, a process that for humans can be broken into meaningful pieces, which are used to provide elaborations and analysis of the answer. An Additional characteristic of the human answering process is the use of modular independent structures, where novel detection abilities may be integrated in the process. For example, learning to identify a new object class allows integrating this object into a variety of questions without requiring an additional training procedure. Finally, the question may guide the answering procedure to focus on specific and subtle details that may be lost in a general features extraction. Such abilities are missing from current machine answering algorithms

In the approach described below, we develop a framework that proceeds along the following steps. It generates a meaningful representation of the question, maps the question representation into a corresponding answering procedure, and applies the answering procedure to the image. The answering process is determined by the question itself and the details of its composition. Our scheme includes a representation of the query's meaning, in which the query is broken into its components. The individual components are handled by procedures that correspond to the type of the component, using existing visual estimators. These procedures are then combined together to provide the final answer. The entire process and its components, including the required visual estimations (such as classification, detection, segmentation and others) their order and combination, depend on the question and are structured to produce an appropriate response. This process does not require any question answering training and is not biased towards the statistics of a specific visual question answering dataset, as current end-to-end approaches.

Our scheme is focused on visual aspects of the image and not on specific domain knowledge. Although we utilize external knowledge sources, it is mainly to assist in understanding the question, and not as a fundamental information source for the answer. A relevant question for our system is a question that can be answered by any human (who understand the question) with an intact visual system, but without depending on specific domain knowledge. For example, the question "What famous book did the man in the picture write?", requires domain-specific knowledge for answering the question (who is the man? what books did he write? which book is famous?). Such questions are not in the scope of this work but they could be answered, using a richer knowledge base.

The system we propose and describe in this work handles a wide range of questions about images, without training on any questions (zero-shot learning). We concentrate on designing a general process for this task and not on fitting results to the statistics of a specific dataset, as current end-to-end approaches. Our system uses many existing methods for different visual tasks, such as detection, classification, segmentation, or extracting objects' properties and relations. In some cases novel detection methods were developed, however this is not a main focus of the work, as our system is modular, enabling `plugging in' new detectors to enhance its capabilities.

\subsection{The structure of questions}

A central aspect of our scheme is that different questions share a similar structure or sub-components with similar structure. For instance, the following questions have components with a common structure:
\begin{quote}
What kind of pants is \textbf{\textit{the person on the bed}} wearing? $\to$ {\color{blue} person \color{red} on \color{green} bed}

Is the \textbf{\textit{giraffe behind a fence}}? $\to$ {\color{blue} giraffe \color{red}behind \color{green} fence}
\end{quote}

The part with common structure can be represented as:
\begin{quote}
There exist $X$ of class {\color{blue} $\boldsymbol{c_{x}}$} and $Y$ of class {\color{green} $\boldsymbol{c_y}$}, such that \textbf{\textit{\color{red} r}}($X, Y$)
\end{quote}

Such structures may serve as building blocks for a compositional question representation. All components with similar structures can be handled by the same procedure, performing part of the answering task.
In our analysis, questions could be represented by a combination of a few types of structures which we refer to as "basic patterns". These patterns are short parametric logical phrases that represent an atomic segment of the question structure. Each basic pattern dictates a particular implementation scheme utilizing a pool of implemented building blocks. 
The combination of basic patterns determines the entire procedure of answering the question. One advantage of such a scheme is that it is modular, allowing the addition of building blocks to increase the scope of the scheme, 
with no dependency on the statistics of a specific visual questions' dataset. A second advantage is that the coverage of queries grows exponentially with the number of building blocks without the need to encounter such queries as training examples. Additional advantage is "understanding" capabilities. The basic meaningful components breaks the process and allows a separate analysis of each component, including reasons of failure and explanations.

The aspect of questions' coverage is also addressed in other directions. Such a direction is increasing the recognizable vocabulary of the question using commonsense knowledge.

\subsection{Utilizing commonsense knowledge}
In many cases answering a question requires integration of prior commonsense knowledge, especially about semantic relations between concepts. For example when answering the question 'What animal is this?' detection capabilities of specific animals (e.g. horse, dog, cat) will not suffice, since the answer requires the general notion of `animal' and which particular instances belong to it. However, a query to an external knowledge database (e.g. ConceptNet \cite{Speer2013}), may provide subcategories of 'animal'. Consequently, specific detectors can be activated to seek these specific recognizable animal types. These knowledge databases are mostly based on information extracted from the internet and include commonsense information about the world. Querying such a database allows the completion of missing information such as semantic connections between object's classes (e.g. synonym, superordinate, subordinate) as in the example above, the typical usage of different objects, and more. Integrating this type of information is important when answering questions asked by humans, as it is common knowledge and treated as universally available.

\section{Related Work}
\label{relatedWork}

Visual Question answering has developed dramatically in the last few years \cite{pandhre2017survey,wu2016visual,Kafle2016vqasurvey}. Practically all current works are based on casting the problem into a multi class classification problem, where image features, retrieved by a Convolutional Neural Network, are fused with question features (mostly extracted by Recurrent Neural Network) and used to predict one of the common training set answers, mostly short and succinct answers. These methods have the advantage of not requiring to incorporate a complicated parsing and understanding process and may present decent results when trained and tested on current existing datasets, yet they lack some important human characteristics like using a compositional process, utilizing existing and meaningful sub processes. Using meaningful sub processes allows humans to focus on different aspects and scopes according to the specific task, utilize existing abilities and modularly integrate novel ones, understand limitation and provide elaborations including suggestions of alternatives.

Incorporating the question information is largely addressed by seeking mechanisms for image-language features fusion. A large focus in this line of works was in simplifying bilinear pulling (which is based on the outer product of the two feature vectors) by reducing dimensionality of the features \cite{akira16} or a low rank factorization \cite{ben2017mutan,yu2017multi}.

In order to extract image information that is more informative to the question and avoid the noise of irrelevant image areas, many works incorporated attention mechanisms. During the attention stage image areas, that are considered more relevant, are multiplied by higher weights and contribute more to answering the question. Attention may be stacked for multiple stages \cite{Yang2016CVPR} with the motivation of refining it for complicated questions. Extracting relevant areas was also performed by integrating regions of detected objects related to question words \cite{ilievski2016focused}. The attention concept was also extended to include both image features and the question representation \cite{lu2016hierarchical}, where both attention types effect each other. Additional attention mechanisms utilize CRF \cite{zhu2017structured}, consider all word-region interactions \cite{nguyen2018}, incorporate correlations between image, question and candidate answer \cite{schwartz2017high} and combine grid based and object detection based regions \cite{lu2018co-attending,anderson2017bottom}.

Combining results of meaningful tasks (other than using pre-trained networks as visual features) such as object detection was in the focus of several additional works. One such work uses object and attribute recognition tasks for proposed regions and combines them with corresponding representations from question and candidate answer \cite{Gupta_2017_ICCV}. The use of visual concepts (object class and attributes) of attended regions and comparing them to extracted concepts from the question was proposed as well \cite{agrawal2018dont}. In another work concatenating pairs of vectors representing two detected objects and their properties with the encoded question was used to allow relation reasoning \cite{desta2018}. Objects and relations between them was utilized in a work that used graph representation for both the image (synthetic images) and the question \cite{Teney_2017_CVPR}. For the image graph objects were the nodes and edges were the spatial relations between them and for the question graph words were the nodes and their dependencies were the edges. Representations were merged in an attention-like mechanism to fuse the features and predict the answer.

A work that uses "facts" extracted from the image, including scene type, detected objects, properties and relations \cite{wang2017vqa}, combined them with co-attention mechanism into a fused feature vector and used attention weights to provide the contributing facts. Facts extraction is not guided by the question and may result in low contribution to questions on non salient details. In addition, any modification of the "facts" detectors would force a full training of the answering module. Providing reasoning was also addressed by merging the answer with the most relevant image caption \cite{li2018-vqae}. Image caption based reasoning (comparing relations extracted from parsed caption and question) was also used to allow answer modification based on Probabilistic Soft Logic (providing contributing relations as evidence) \cite{aditya2018explicit}. In some cases the representation was based on image caption where relevant words (based on image caption data), a sentence describing the image and the question were fused to feed the answer classifier \cite{li2018tell}.

The compositionality concept in visual question answering was addressed by the Neural module Network (NMN) works that compose a dynamic network out of trained modules. The original layout of these modules is based on the dependency parsing of the question \cite{andreas2016neural} and was also enhanced to include learning for the selection of the layout \cite{andreas16naacl}. Following the release of the CLEVR dataset \cite{johnson2016clevr}, which includes annotations for the answering programs, the layout was also learned in a supervised manner according to these programs \cite{johnson2017inferring,hu2017learning}. It is important to note that even though meaningful programs may be learned and corresponding modules are assigned, the modules are not trained to perform any independent meaningful task and its learned function is only to serve as a component for the question answering network trained for a specific dataset. This means that there is no flexibility and options to incorporate exiting methods as in our approach or modularly modify and improve the modules. NMN requires a large amount of question-answer examples, unlike our approach, that requires none. As NMN provide answers by a classification, no elaboration or limitation aware answers (e.g. 'Unknown class: scissors') are possible. In addition there is no utilization of commonsense information.

Another aspect of question answering related to our work is integrating external prior knowledge. One approach focused on questions that require external knowledge in addition to the image. This was addressed by querying knowledge databases according to visual concepts (objects, image scenes and image attributes) detected in the image. The query was generated either by mapping the question to a template \cite{wang2015explicit} or, in a modified version, by a learned mapping \cite{wang2016fvqa}. Another approach merged external knowledge, extracted using detected image attributes, with image representation (detected attributes and generated captions) and question \cite{Wu2016CVPR}. Integrating external knowledge using a Dynamic Memory Network \cite{xiong2016dynamic} was proposed  where knowledge base queries are based on detected objects and question keywords \cite{li2017incorporating}. 

The common to all the above approaches is that they cast the problem as one learning problem (mostly end-to-end, multi class classification), tailored for a specific datasest. Incorporation of compositionality, reasoning, attention mechanisms, external knowledge and visual detection tasks can all be described as part of "improved" feature extraction for the final classification task. No meaningful, independent tasks are used in the answering process, as naturally done by humans. When humans learn to identify a new object, property or relation, they can immediately incorporate it in their answering mechanism. Such modularity does not exist in current visual question answering systems, where each change requires a full retrain of the answering system. It is also evident that while existing methods may provide reasonable statistical results on existing datasets, it does so by exploiting inherent biases, which leads to insensitivity to full question details and images, with a tendency to fail on novel characteristics \cite{agrawal2016analyzing}. A system that builds and runs a meaningful process, tailored for the question without "seeing" any question-answer example was not proposed as far as we know. Such a process, utilizing existing visual analyzers and external knowledge is completely modular, aware for its detection limitations and can elaborate and correct negative or ungrounded answers. Desired and important capabilities may be addressed, even if they are not statistically prominent.

\section{UnCoRd Answering System}
\label{sec:system}

\subsection{Approach Overview}

Our Understand, Compose and Respond (UnCoRd) approach is based on the following observations:

\begin{itemize}
\item There is a representation of the question in terms of objects, their classes, properties and relations, including quantifiers and logical connectives as well as non logical symbols: predicates and functions. The representation has an 'abstract' structure, i.e. independent of the particular objects, classes, properties and relations that are represented as parameters. A single abstract representation can represent many different concrete questions.

Our main thesis is that the procedure to be applied for obtaining the answer depends on the abstract structure of the question and not the particular elements. Hence, it is important to use the right kind of abstract representation, which will allow this mapping to procedures (where all questions with the same abstract structure require the same procedure). A proper parsing and mapping of the language question to its abstract representation should be obtained to use this method.

\item The question has a compositional structure: there are basic components put together in particular ways. The abstract representations are composed from 'basic patterns' and methods for putting them together into more complex compound structures. This compound structure determines how the procedures are constructed. There are basic procedures for the basic patterns, and methods of composing from them a more complex procedure to deal with the compound abstract structures. In other words, we get a procedure for the entire question by having procedures for the basic components and a procedure to put them together.
\end{itemize}

We would like our system to meet the following criteria:
\begin{list}{--}{}
\item Answer correctly and efficiently.
\item ``Understanding'' the question, in the sense of:
\begin{list}{$\circ$}{}
\item Breaking the answering procedure into a set of simple visual tasks.
\item Identify which tasks it can perform and what are its limitations. Indicate if something is missing or unknown.
\item Ability to explain and reason - elaboration of the answering process using the image and intermediate results, including error correction and alternative suggestion. 
\end{list}
\item Modularity and robustness: handling questions and image categories of various types, not limited by a training set. 
\item Though not using a human psychology model, the ability to handle questions that people answer easily (and may be "hard" for computers) is desired, e.g. 'odd man out'.
\end{list}

A question can be seen as a statement about the image that the answering system tries to make true or refute. Making the statement true requires an assignment of the particular classes, properties and relations to the image. Their identification in the image is based on pre-trained classifiers and detectors. The recognizable set is modular and can be increased by adding new detectors or switching to stronger ones. Logical operations will be used to generate logic sentences with a formulation that fits first order logic (including functions) with some extensions.


The answering procedure is generated according to the input question in the following manner:
\begin{quote}
Question $\rightarrow$ Question representation $\rightarrow$ procedure
\end{quote}
A proper representation is fundamental to allow a successful mapping of the question into the answering routine. This representation should be concise and support generating the same procedure when applied to similar structured questions with different choices of classes, properties and relations. To obtain that, the visual elements (object classes, object properties and object relations) would be parameters, integrated using logic operations (e.g. $\land, \lor$) and quantifiers (e.g. $\forall, \exists, \exists 5$ ) into basic logic patterns corresponding to specific structures. These patterns are combined and merged to compose a more complicated structures that create the representation of the question and can be mapped to the answering procedure.

We use a directed graph to describe the question which is a natural choice in our case and allows diverse compositions of substructures. In this graph each node represents an object entity and its description (e.g. a list of required properties). These nodes are linked by the graph edges which represents relation between objects. The graph is divided into small segments that relate either to one node and correspond to part of its information (e.g. object class and one property) or to an edge and the two classes of the nodes it connects. Each of these graph segments matches a basic pattern that is handled by a corresponding procedure, using the specific visual elements of this substructure. The graph representation allows to decompose the answering procedure into a set of elementary procedures and put them together to generate a modular answering procedure. The elementary procedures invoke visual analyzers, which are the basic modules of the process. Each class, property and relation, has a visual analyzer to establish it. More general visual operations that serve more than one particular visual element (e.g. depth estimation) are activated according to need and their results are available to all basic procedures. The overall routine is obtained by applying these procedures and operations at an appropriate order, to appropriate objects, where the amount of required assignments per object are set by the quantifier of the corresponding node. The visual elements may have 'types', such as classes that can be basic or subordinate (i.e. basic with additional properties), properties that may be comparative (e.g. 'older than') and relations which can be symmetric (e.g. 'beside') or not.

The entire process of answering a visual question is described in Figure \ref{fig:proc}. It starts by receiving the input language question and mapping it to a graph representation. The next stage is running a recursive procedure that follows the graph and invokes the procedures associated with the basic structures, using the specific visual elements as inputs. After the results are obtained, the answer is returned.\\

\tikzstyle{block} = [rectangle, draw, fill=yellow!20, text width=25em, text centered, rounded corners, minimum height=4em]
\tikzstyle{block_tt} = [rectangle, draw, fill=yellow!20, text width=25em, text depth=0.8cm, text centered, rounded corners, minimum height=4em]
\tikzstyle{small_block} = [rectangle, draw, fill=yellow!80, text width=3em, text centered, rounded corners, minimum height=3.5em]
\tikzstyle{small_block2} = [rectangle, draw, fill=yellow!80, text width=7em, text centered, rounded corners, minimum height=1em]
\tikzstyle{line} = [draw, very thick, color=black!50, -latex']
\tikzstyle{cloud} = [draw, ellipse,fill=orange!20, node distance=5cm, minimum height=2em, text centered, text width=3.5em]
\tikzstyle{cloud2} = [draw, ellipse,fill=cyan!20, node distance=1.6cm, minimum height=2em, text centered, text width=3.5em]

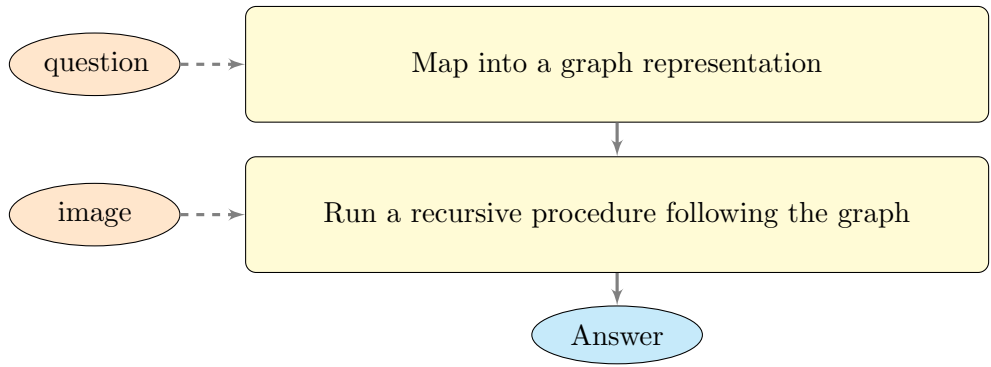
\begin{figure} [h!]
\begin{center}
\begin{tikzpicture}[scale=2, node distance = 2cm, auto]
    \node [block] (identify) {Map into a graph representation};
    \path (identify.west)+(-1, 0) node [cloud] (question) {question};
    \node [block, below of=identify] (run) {Run a recursive procedure following the graph};
    \path (run.west)+(-1, 0) node [cloud] (image) {image};
    \node [cloud2, below of=run] (answer) {Answer};
    \path [line] (identify) -- (run);
    \path [line, dashed] (question) -- (identify);
    \path [line, dashed] (image) -- (run);
    \path [line] (run) -- (answer);
\end{tikzpicture}
\end{center}
\caption[QA process]{A diagram for the process of answering a visual question}
\label{fig:proc}
\end{figure}

\tikzstyle{cir0} = [draw, circle,fill=yellow!20, align = left, inner sep=0.5ex, scale=0.7]
\tikzstyle{cir1} = [draw, circle,fill=yellow!20, text centered, text width=4.5em, inner sep=0.5ex, scale=0.8]
\tikzstyle{cir4} = [draw, circle,fill=yellow!20, text centered, text width=5.4em, inner sep=0.5ex, scale=0.8]
\tikzstyle{cir2} = [draw, circle,fill=yellow!20, text centered, text width=4.5em, inner sep=0.5ex, scale=0.6]

Questions with a simple structure (e.g. \textit{"Is there a red car?"}) can be represented by matching one specific pattern to a question. This covers a wide range of questions, however by allowing a composition of simple patterns, into a more complicated structures, the quantity of supported questions is raised substantially (from $\sim$60\% to $\sim$90\%, according to an analysis of 542 questions on images asked freely by people and using a set of 12 patterns). This composition is done using a graph. For example in the question \textit{"Is there a red car to the right of the yellow bus?}" there are two parts with a simple structure "Is there an object of class $\boldsymbol{c}$ with a property $\boldsymbol{p}$?" connected by the relation "to the right of", which corresponds to another simple structure: \textit{"Is there an object of class $\boldsymbol{c_1}$  and an object of class $\boldsymbol{c_2}$ that have the relation $\boldsymbol{r}$ between them?"}. The graph representing the question is:

\begin{center}
\begin{tikzpicture}[scale=3, node distance = 2cm, auto]
    \node [cir1, align = left] (node1) { $\; \boldsymbol{c}$: bus \\
                                         $\; \boldsymbol{p}$: yellow};

    \node [cir1, above of = node1, align = left] at (0, 0.2) (node2) {$\quad \boldsymbol{c}$: car \\
                                                                      $\quad \boldsymbol{p}$: red };
    \node at (-0.4, 0.37) {\scriptsize{'to the right of'}};
    \path [line] (node1) -- (node2);
\end{tikzpicture}
\end{center}

When a specific question is given, the question is parsed and mapped to a directed graph, where the visual elements are its parameters. This graph corresponds to a logic expression that is composed of simple expressions, that may share the object variables. Some of the parametric visual elements are variables that require estimation based on the image. Once the variables are estimated, the logic expression is evaluated (as true or false) and the query is answered accordingly. The formulation of the logic expression fit first order logic (including functions) with some extensions (e.g. a variable-sized set of arguments or outputs for some functions).

Each simple logic expression is related to a basic pattern, which corresponds to a basic procedure. The basic procedure obtains an answer to the expression by activating visual analyzers according to the types of object classes, properties and relations (which are inputs to the basic procedure). Such a system will have the ability of constant improvement by adding detectors for new classes, properties and relations according to requirements. Similar characteristics are also evident in human learning, where new learned details are integrated into the existing mechanism of world perception.

The UnCoRd system is implemented following the approach described above. It answers visual questions using a composed process that follows the graph representation of the question, activating real world visual analyzers. This system is described in the following section.

\subsection{System Description}

\subsubsection{Mapping to a Directed Graph}

One of the system's main tasks is to translate the query, given in natural language, into an abstract representation which will then be mapped into a procedure (the first step, described in Figure \ref{fig:proc}). We first use the START parser 
\cite{katz1988using,katz1997annotating} for transforming the question into a set of ternary expressions of the form \textbf{[subject relation object]}. For example the ternary expressions representing the question \textbf{\textit{"Is there a red car?"}} are \textbf{[car$_{[id=car\_id]}$ be null]} and \textbf{[car$_{[id=car\_id]}$ has\_property red]}.

\begin{figure} [h!]
\begin{center}
\begin{tikzpicture}[scale=3, node distance = 2cm, auto]
    \node [cir4, align = center] at ( 0.7, 0) (node4) { $\; \boldsymbol{c}$: car };

    \node [cir4, align = center] at (-0.7, 0) (node3) { $\; \boldsymbol{c}$: grass \\
                                            \scriptsize{$\; \boldsymbol{p}$: green}};

    \node [cir4, above of = node3, align = left] at (0, 0.2) (node2) {$      \quad \boldsymbol{c}$: cat \\
                                                                      \scriptsize{$\boldsymbol{p}$: \{small, red\} } \\
                                                                                  $\boldsymbol{q}$: 'all'};
    \node at (-0.5, 0.44) {\scriptsize{'on'}};
    \node at ( 0.6, 0.44) {\scriptsize{'behind'}};

    \node [cir4, above of = node2, align = left] at (0, 1.07) (node1) {     \quad $\boldsymbol{c}$: child  \\
                                                           \scriptsize{\quad\quad $\boldsymbol{p}$: tall } \\
                                                        			   \quad\quad $\boldsymbol{q}$: 2};
    \node at (-0.33, 1.165) {\scriptsize{'look\_at'}};
    \path [line] (node4) -- (node2);
    \path [line] (node3) -- (node2);
    \path [line] (node2) -- (node1);
\end{tikzpicture}
\end{center}
\caption[Q graph]{An example for the directed graph representing the question: \textit{'Are the two tall children looking at all the red small cats that are on the green grass and behind the car?'}, where $c$ is the object's class, $p$ is a list of its required properties and $q$ is the required quantifier.}
\label{fig:graph_ex}
\end{figure}
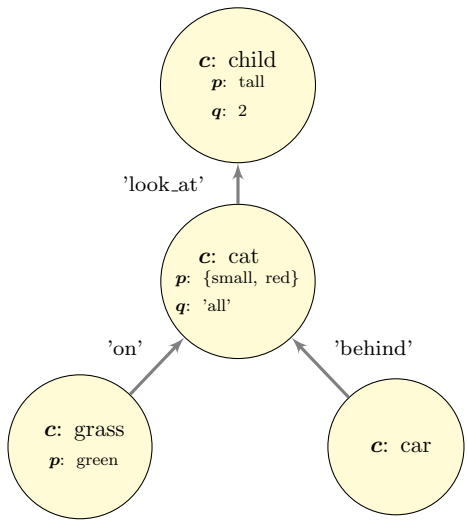

The generated set of ternary expressions is used for the generation of a graph representation, where nodes represent objects and edges represent relations between objects. The node include all of the object's requirements according to the question, mainly its class, properties that may be required (e.g. 'red') or queried (e.g. 'what color') and quantifiers that are not the default existence quantifier (e.g. 'all', 'two'). The directed edges correspond to relations between objects where the edge direction implies the direction of relation. Each edge is also assigned a direction of progress for the answering procedure. It is instantiated as the relation direction, but may be modified according to initial object detection to enhance detection abilities (see Section \ref{sec:qRelAtt} for details). An example for a mapping of a question to a directed graph can be seen in Figure \ref{fig:graph_ex}.

The graph representation is used to fit an answering procedure for each particular question. Fragments of information are extracted from subgraphs that include up to two connected nodes. A graph fragment includes a subset of elements (classes, properties, property functions and relations) that has a mapping to one of a few basic logic patterns. This mapping, combined with the particular accompanying visual elements defines a logic expression that selects and guides a component of the answering procedure. For example a fragment of the node's class and a required property is mapped to the pattern $\exists X \, (c_X(X) \land p_X(X))$.
The specific class $c_X$ and property $p_X$ define the particular logic expression that should be checked. Such mappings are done for the entire graph, where each fragment of it is mapped into a basic logic pattern and specific visual elements. These simple logic expressions, joined using logic operations, constitute one logic expression that represents the entire question.

Each basic logic patterns has a dedicated procedure that performs the evaluation required to confirm or refute it, using visual analysis according to the image. The procedure provide an answer according to an accompanying query.

We use the following notations for describing the basic logic patterns:
\par\nobreak

\begin{tabular}{l c p{11.9cm}}
$X, Y$    & --- & Objects \\
$c(X)$    & --- & A class, evaluated for object $X$ (as True/False), e.g. 'person', 'boy', 'bird', 'train'. \\
$p(X)$    & --- & A predicate property (predicate of arity 1), evaluated for object $X$, (as True/False), e.g. 'blue', 'male', 'big'. \\ 
$f(X)$    & --- & A property function. Returns properties of a specific type, e.g. 'color', 'age', 'size'.\\ 
$g(S_t)$  & --- & A global property function for a subset of objects of the same class: $S_t \subset \{X_t : c_t(X_t)\}$ . Returns properties of a specific type, e.g. 'quantity', 'difference', 'similarity'.\\
$p^f$     & --- & A predicate property, constrained to possible return values of $f(X)$ (e.g. $blue = color(X), male = gender(X), big = size(X)$).\\ 
$a^g$     & --- & One of the possible values returned by $g(X)$ (e.g. $3 = quantity(S_t)$, where $S_t = \{X_t : c_t(X_t)\}$). \\
$r(X, Y)$ & --- & Relation between objects $X$ and $Y$ (predicate of arity 2), e.g. $X$ below $Y$ $\to below(X, Y)$ and in the same manner $looking\_at(X, Y), near(X, Y)$. \\ 
$\textbf{?-}$ & --- & A query, the requested answer. \\ 
\end{tabular}
\\

Objects (or other elements) starting with a capital letter (e.g. $X, Y$) are unknown elements (variables) that should be estimated according to the image.\\

The particular used patterns were selected since they provide a small, simple and basic set that can naturally compose the logic representation of the question. This small set provides a high flexibility in composing a wide variety of logic expressions using the different visual elements. From a conducted survey and other checks it was evident that this set is empirically sufficient to represent the set of analyzed queries.

Following are the basic logic patterns that are mapped to basic procedures in the question answering process (followed by their corresponding graph fragment). The $\exists$ quantifier may be replaced by other quantifiers (e.g. $\forall$, $\exists 2$).

\begin{itemize}

\item \textbf{Property Existence:} $\exists X \, (c_X(X) \land p_X(X)) ; \; \textbf{?-} \exists /  c_X $ \; \begin{tikzpicture} [baseline=-\the\dimexpr\fontdimen22\textfont2\relax] \node [cir0] (node) { $\boldsymbol{c}$: $c_X$ \\ $\boldsymbol{p}$: $p_X$}; \end{tikzpicture}

Examples: \textit{'Is there a brown bear?'} (query for validity with a specific object class)\\
  \hphantom{Examples:} \textit{'What is the purple object?'} (unknown and queried object class)

An example for a modification due to a quantifier parameter: $\forall X \, (c_X(X) \land p_X(X)) ; \; \textbf{?-} \exists $, e.g.  \textit{'Are all bears brown?'}

\item \textbf{Function Property:} $\exists X \, (c_X(X)), f(X) = P^f ; \; \textbf{?-} P^f$ \; \begin{tikzpicture} [baseline=-\the\dimexpr\fontdimen22\textfont2\relax] \node [cir0] (node) { $\boldsymbol{c}$: $c_X$ \\ $\boldsymbol{f}$: $P^f$}; \end{tikzpicture}

Example: \textit{'what color is the chair?'}

\item \textbf{Property of a Set:} $\forall X_t \exists S_t \, (S_t = \{ X_t : c_t(X_t)\}), g(S_t) = A^g ; \; \textbf{?-} A^g$ \; \begin{tikzpicture} [baseline=-\the\dimexpr\fontdimen22\textfont2\relax] \node [cir0] (node) { $\boldsymbol{c}$: $c_{X_t}$ \\ $\boldsymbol{g}$: $A^g$}; \end{tikzpicture}

Example: \textit{'How many planes are in the photo?'}  

\item \textbf{Object Existence:} $\exists X \, (c_X(X)) ; \; \textbf{?-} \exists / c $ \; \begin{tikzpicture} [baseline=-\the\dimexpr\fontdimen22\textfont2\relax] \node [cir0] (node) {$\boldsymbol{c}$: ${c_X}$}; \end{tikzpicture}

Examples: \textit{'Is this a dog?'} \\
\hphantom{Examples:} \textit{'What is it?'}

\item \textbf{Relation Existence:} $\exists X \, \exists Y \, (c_{X}(X) \land c_{Y}(Y) \land r(X, Y)) ; \; \textbf{?-} \exists / c_{X} / c_{Y}$  \; \begin{tikzpicture} [baseline=-\the\dimexpr\fontdimen22\textfont2\relax] \node  at (0, -0.42) [cir0] (node1) { $\boldsymbol{c}$: $c_{Y}$}; \node at (0, 0.42) [cir0] (node2) {$\boldsymbol{c}$: $c_{X}$}; \node at (-0.2, 0) {\scriptsize{$r$}}; \draw [->, line width = 0.3mm] (node1) edge (node2); \end{tikzpicture}

Examples: \textit{'Is the man looking at the children?'} (validity query) \\
\hphantom{Examples:} \textit{'What is on top of the television?'} (query for one of the classes)
\end{itemize}

The combination and composition of these patterns has a powerful representation capabilities and provides a mapping to a set of basic procedures that constitute the full answering procedure. The procedure composition of ``real-world'' visual tasks allows both the use of existing detectors, including separate improvement of each task and explaining, elaborating and correcting questions. 

\begin{figure} [h!]
\begin{centering}
\begin{tabular}{cc}
\subfigure{
  \includegraphics[totalheight=0.23\textheight]{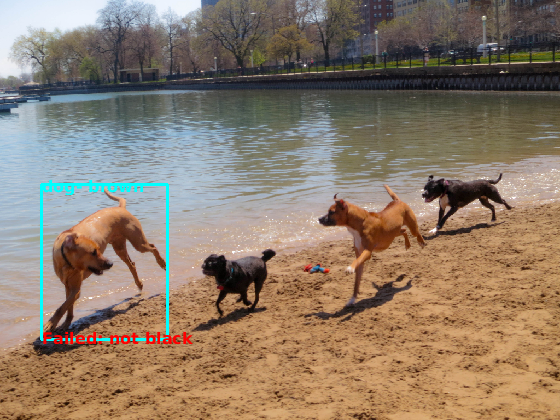}
} &
\subfigure{
  \includegraphics[totalheight=0.23\textheight]{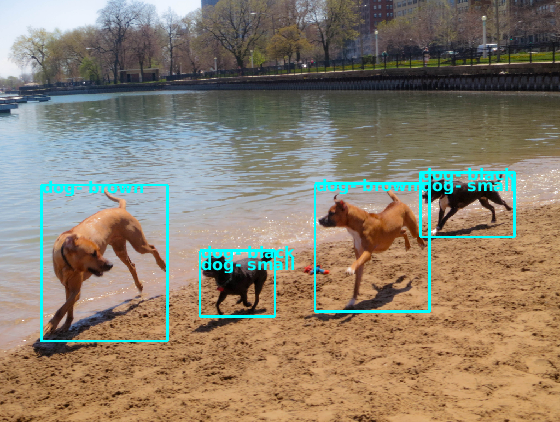}
}  \\
\multicolumn{1}{l}{ (a)  \textbf{Q}:  Are \textbf{all} dogs small and black?}                                                        & \multicolumn{1}{l}{(b) \textbf{Q}: Are \textbf{all} black dogs small?}  \\
\multicolumn{1}{l}{ \aindent \textbf{A}: no \scriptsize{[\textbf{Full}:  There are not enough proper dogs }}                          & \multicolumn{1}{l}{\aindent  \textbf{A}: yes }                          \\
\multicolumn{1}{l}{ \leavevmode\hphantom{(a) \textbf{A}: no \scriptsize{[\textbf{Full}:}} \scriptsize{(failed due to a brown dog)]}} &                                                                         \\[6pt]

\end{tabular}
\caption['all' quantifier example]{An example for 'all' quantifier: The question in (a) requires all 'dog' objects to be both black and small, hence the first dog that is not black renders the logic phrase false and the answer is ``no'' (failed object and reason are marked in the image). The question in (b) requires only that the black dogs would be small, hence all dogs are checked for color, and the size of the black ones is verified to be small. Since it is true, the answer is ``yes''.}
\label{fig:quant_ex}
\end{centering}
\end{figure}

As mentioned above, modified quantifiers may be added to nodes according to amount of objects required in the questions (see Figure \ref{fig:graph_ex}). These quantifiers may be either numbers (e.g. \textit{'Are there three guys?'}) or 'all' for entire group of objects. Setting the group may be according to subtle phrase differences which affect the answering procedure flow and results as can be seen in Figure \ref{fig:quant_ex}.


The graph naturally represents objects, their properties and binary connections between them. Though this covers a wide variety of questions, using global image information and some extensions to the basic graph increase the support to additional attributes. Property of a group is an example for such an extension. Properties that uses global information are 'closest' and 'size' (which is relative to other objects).

\begin{figure} [h!]
\begin{centering}
\includegraphics[totalheight=0.3\textheight]{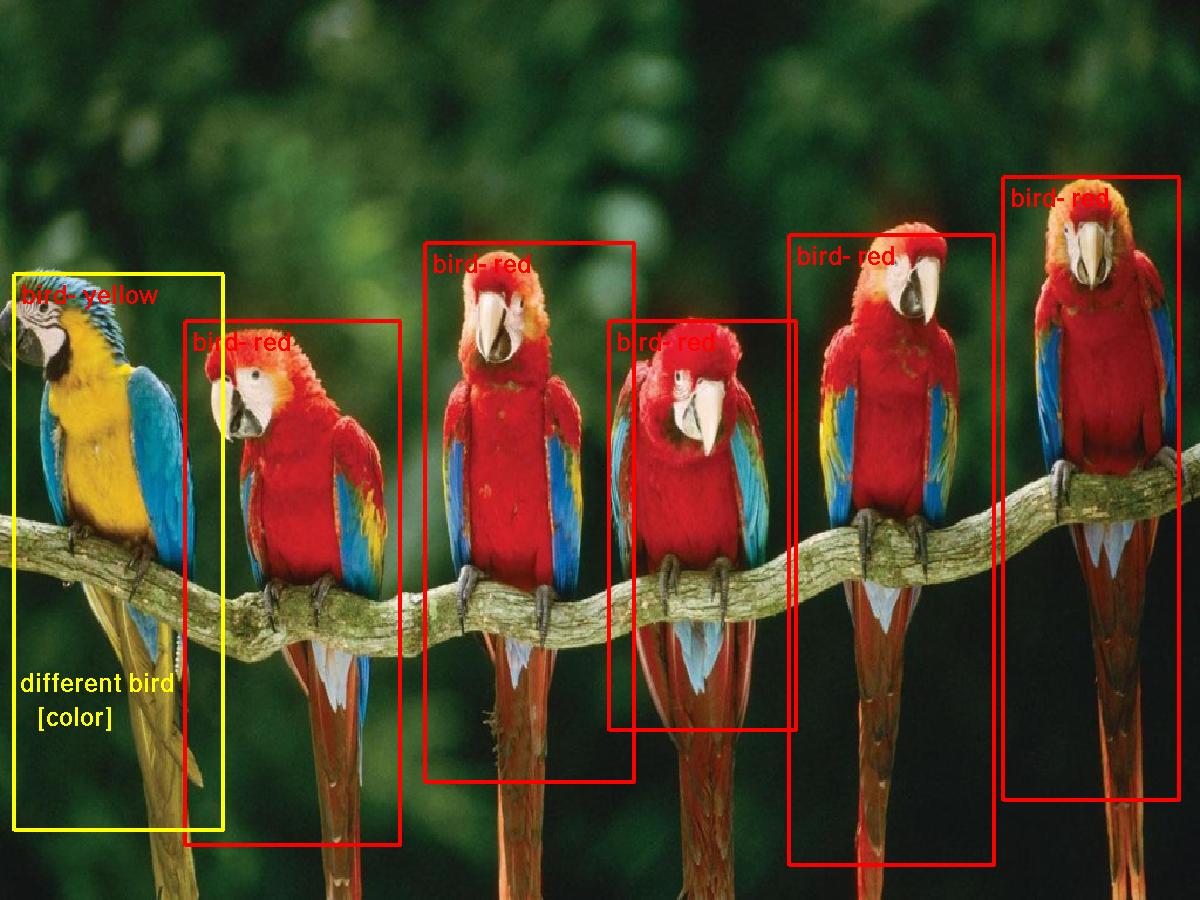}\\
\begin{tabular}{l}
\textbf{Q:} What difference does one bird have? \\
\textbf{A:} color (yellow), object center: (95, 325)\\
\end{tabular}
\caption[sub object example]{An 'odd man out' question for objects of class 'bird'. This is a complicated attribute that requires special treatment and mapping to the graph representation. Bounding boxes for birds with common property (red) and for 'odd man out' bird (yellow) are marked (in red and yellow correspondingly). [Object detection is based on faster R-CNN + DeepLab].}
\label{fig:diff}
\end{centering}
\end{figure}

Specific implementations for complicated attributes may be added as a dedicated tasks or by a preprocessing, braking it into graph plausible segments. An example for such an implementation in our system is 'odd man out' (e.g. ``How is one cow not like the others?''), where the relations 'diff\_\textless$f$\textgreater' and 'sim\_\textless$f$\textgreater' (for different and similar values of property $f$ correspondingly) are used to check and compare the properties of objects. An example is given in Figure \ref{fig:diff}.  The 'similarity' attribute (queries for a property that is similar for all objects in the group) is handled in the same manner.

The main building blocks of the question representation are the visual elements: object classes, object properties and object relations.

\begin{itemize}

\item \textbf{Object Classes}
Object class is the category of object required by the question. It does not necessarily match the used object detector. To enlarge the coverage of supported object classes we define a few categories of object classes and handle them accordingly.

\begin{itemize}
\item \textbf{Basic Classes}

These are the classes specifically covered by the main multi-class object detector. We currently use instance segmentation by mask R-CNN \cite{he2017maskrcnn} for the 80 classes of COCO dataset \cite{lin2014microsoft}. Having the segmented object is very useful as this accuracy is required in many cases (e.g. for the relation 'touch'). Other detection methods are also integrated and may be used instead. In many of the Figures, object detection is based on faster R-CNN \cite{ren2015faster} complemented by DeepLab's semantic segmentation \cite{chen14semantic,papandreou15weak,KrahenbuhlK11} (for the 20 classes of PASCAL VOC dataset \cite{everingham2014pascal}).

\item \textbf{Subordinate Classes}

When the requested class is a sub-group of a basic class, an object of this basic class should be detected and then additional properties are checked. It is used for the 'person' subordinate classes (e.g. 'woman'), where face detection is activated \cite{Mathias2014Eccv} for the detected 'person' objects, followed by age and gender classifier \cite{LH:CVPRw15:age} on the results (an example is demonstrated in Figure \ref{fig:classTypes}).

\item \textbf{Superordinate Classes}
\label{sec:superordinate}

Each category of a superordinate class includes a few basic classes (for example furniture, animal). To check this, we use ConceptNet \cite{Speer2013}, which is a commonsense knowledge database, based on data extracted from the internet (see also section \ref{sec:conceptNet}). It includes concepts and predefined relations between them. We use the relations: 'InstanceOf', 'IsA', 'MadeOf' and 'PartOf' with the requested class, and keep the results that fit our basic classes list. The detected objects of these classes are retrieved and used for the rest of the procedure. Also if the query is for the type of the requested superordinate class, the name of the detected basic class is given as an answer (see Figure \ref{fig:classTypes} for an example).

\item \textbf{Similar Classes}
A class that has a synonym or a very similar class in the basic classes set may be also searched as this corresponding class. These correspondences are extracted using the 'Synonym' and 'SimilarTo' relations in ConceptNet.

\item \textbf{A Group of Objects}
\par\nobreak
To identify a class that represents a group of objects (possibly of different optional basic classes), the ConceptNet relation 'MemberOf' is used (e.g. flock $\rightarrow$ bird, sheep; fleet $\rightarrow$ bus, ship...). A quantity requirement is added of at least two objects (demonstrated in Figure \ref{fig:classTypes}).


\item \textbf{Sub Objects}

Some objects are part of a 'known' objects and can be extracted according to the detection of the host object and additional processing. We apply human pose estimation \cite{ChenNIPS14} to obtain the different body parts when requested (e.g. \textit{'left/right hand', 'left/right foot'}). Relative areas of objects (e.g. \textit{'the middle of the bus'}) are also treated as sub objects. In these cases left and right are different than other uses of left/right as a location property (e.g. \textit{'the left box'}). A 'shirt' is also treated as a sub object, corresponding to the torso area, provided by human pose estimation results (an example is given in Figure \ref{fig:classTypes}). 

\begin{figure} [h!]
\begin{centering}
\setlength\tabcolsep{1.5pt} 
\renewcommand{\arraystretch}{0.8}
\begin{tabular}{p{3.7cm}p{3.7cm}p{3.7cm}p{3.7cm}}
\subfigure{
  \includegraphics[totalheight=0.13\textheight]{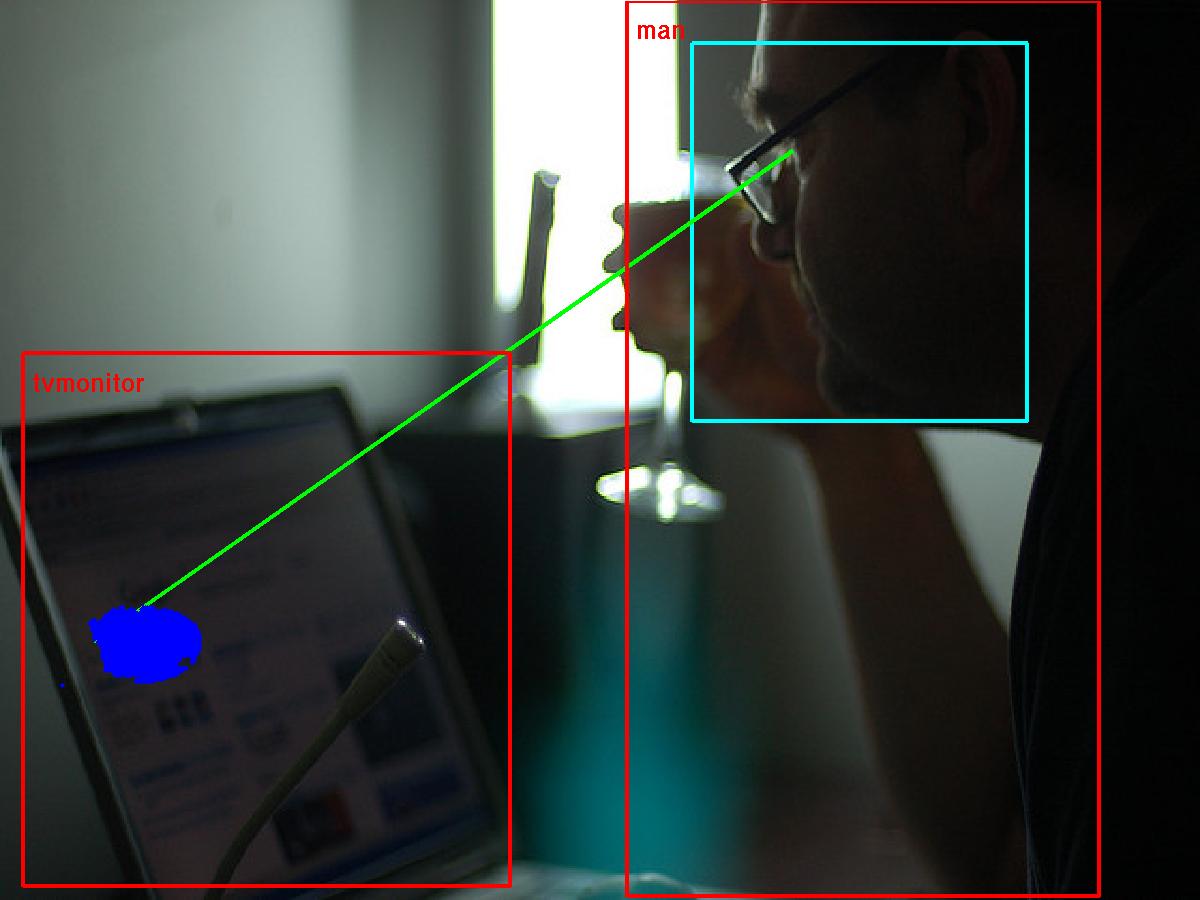}
} &
\subfigure{
  \includegraphics[totalheight=0.13\textheight]{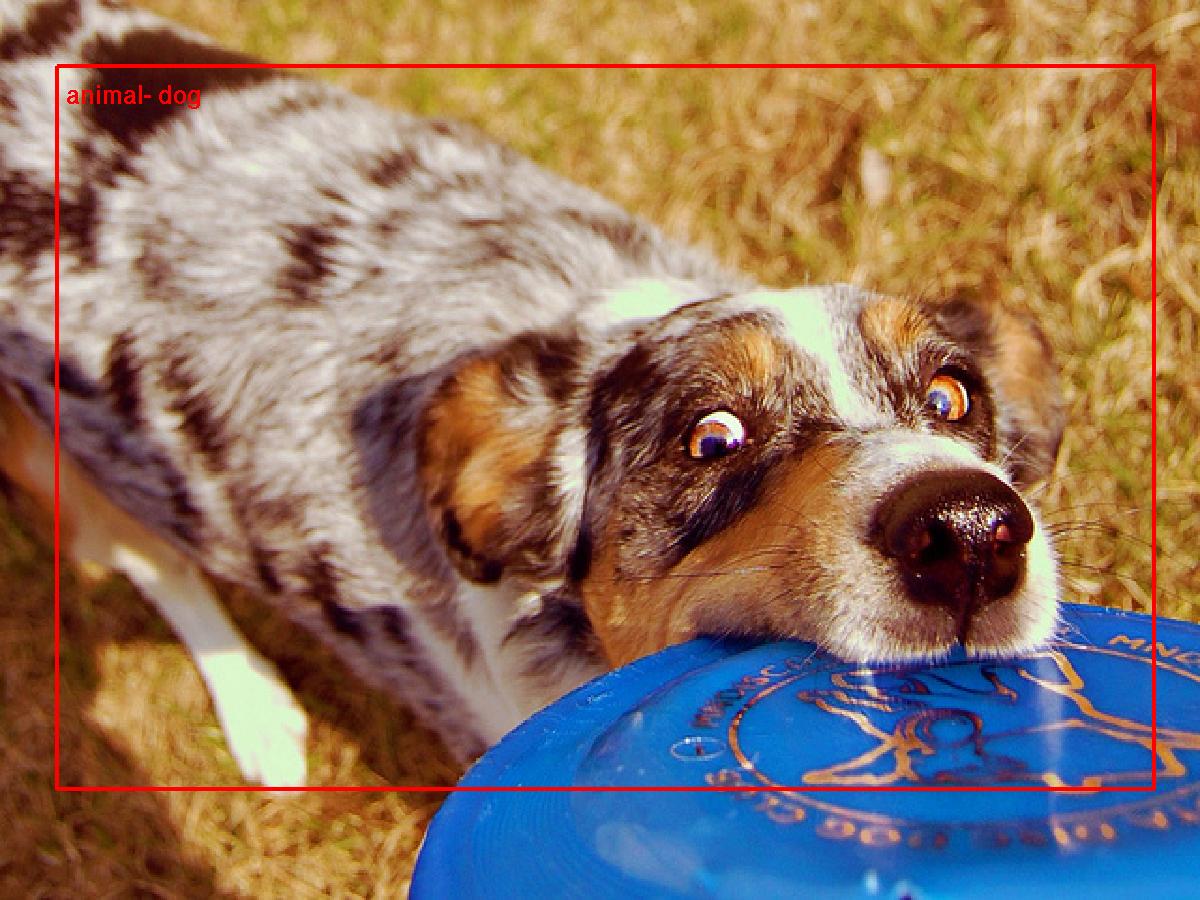}
} &
\subfigure{
  \includegraphics[totalheight=0.13\textheight]{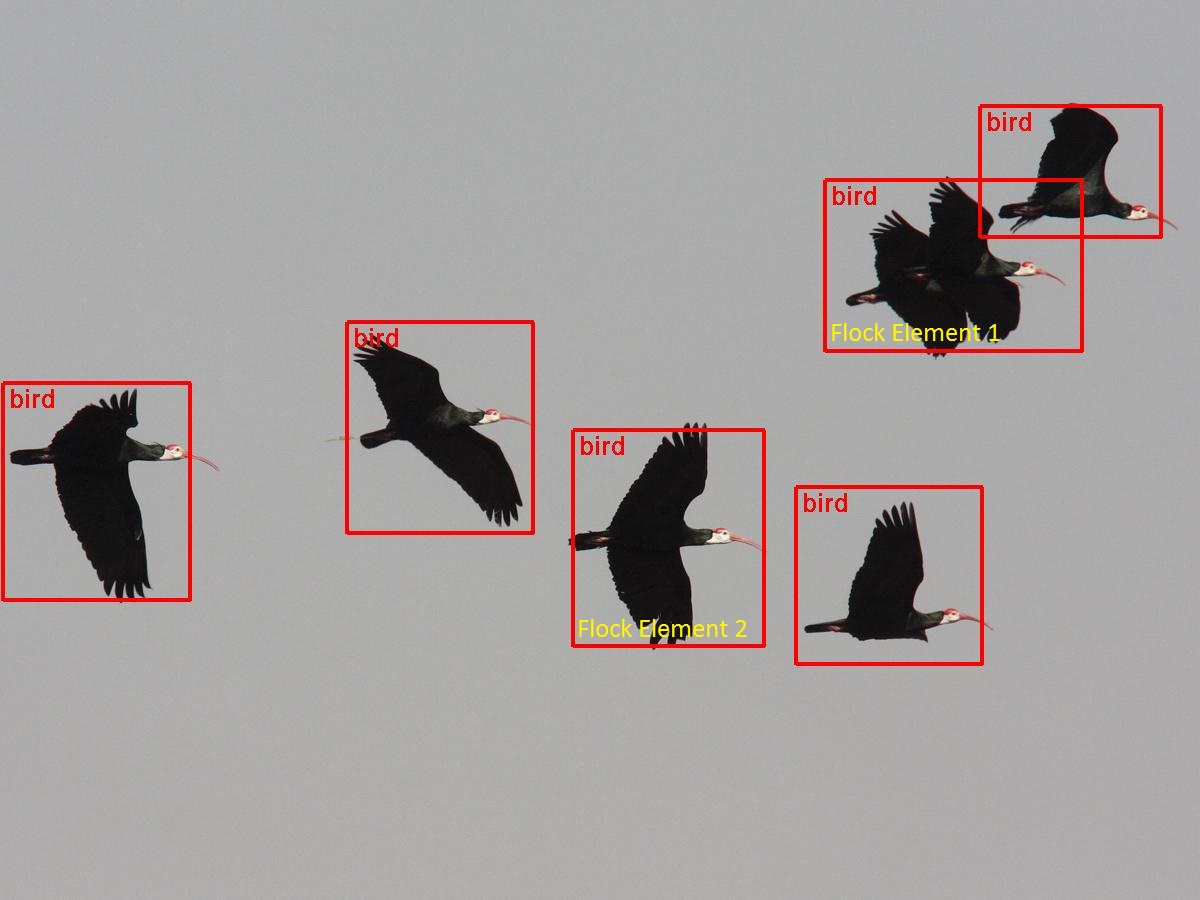}
} &
\subfigure{
  \includegraphics[totalheight=0.13\textheight]{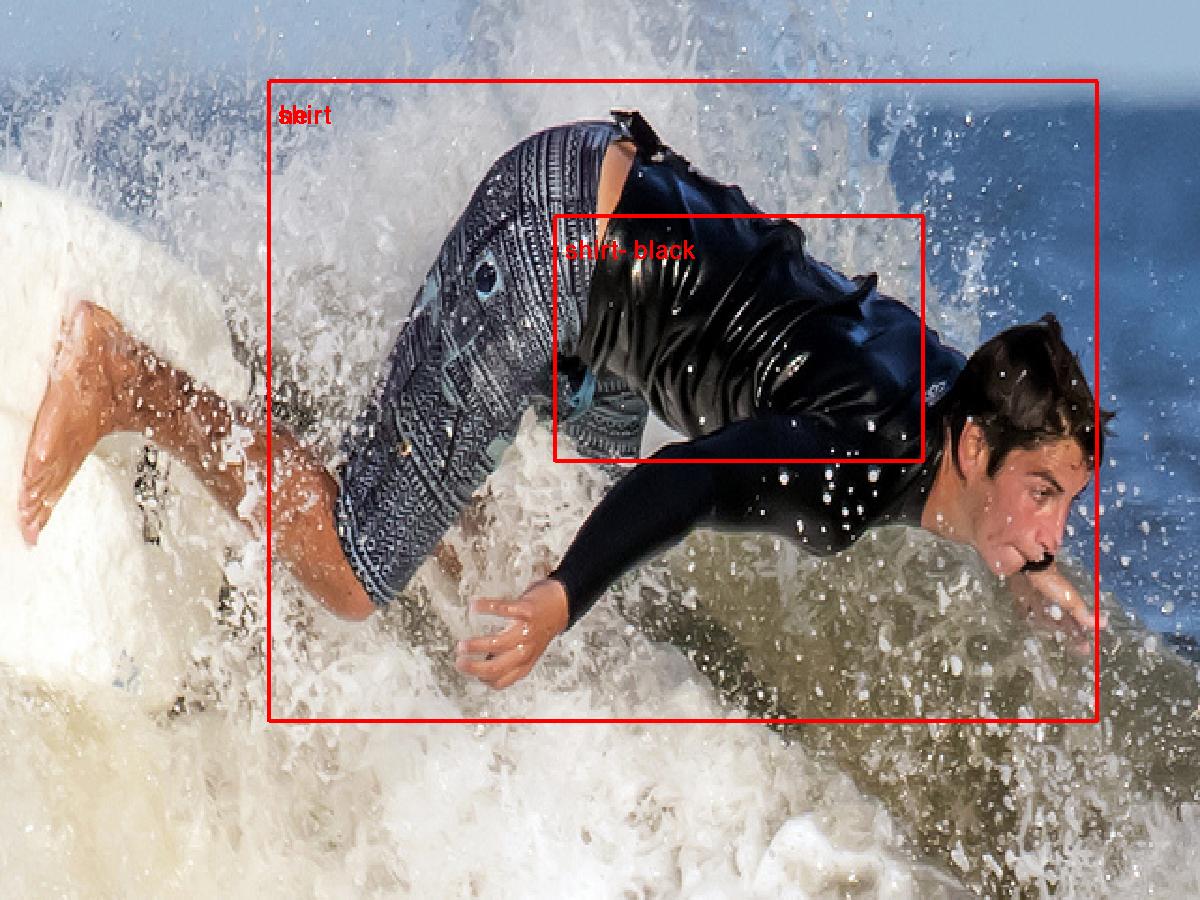}
} \\

\scriptsize{\textbf{Q:}} \tiny{What is the man looking at?} & \scriptsize{\textbf{Q:}} \tiny{What kind of animal is this?} & \scriptsize{\textbf{Q:} Is there a flock?} & \scriptsize{\textbf{Q:} What color is his shirt?} \\
\scriptsize{\textbf{A:} tvmonitor} &  \scriptsize{\textbf{A:} dog} &  \scriptsize{\textbf{A:} yes} & \scriptsize{\textbf{A:} black}

\end{tabular}
\caption[Object class types]{Examples for detection of different object class types. From left to right: subordinate class (person $\rightarrow$ man), subordinate class (dog $\rightarrow$ animal), group (multiple birds $\rightarrow$ flock), sub object (person $\rightarrow$ shirt). [Object detection is based on faster R-CNN + DeepLab].}
\label{fig:classTypes}
\end{centering}
\end{figure}

\end{itemize}

\item \textbf{Object Properties}

Objects have various visual properties. We differentiate between the \textbf{binary properties} (e.g. 'red') and a \textbf{function property} that returns the property of the object from a specific category (e.g. 'color'). Table \ref{table:usedProps} describes the used set of properties divided (most of them) to groups of function properties.

\begin{table}[h!]
\begin{minipage}{\textwidth}
\begin{centering}
    \begin{tabular}{ | l | l |} 
    \hline
    \textbf{Properties' Group}                                                              & \textbf{Predicate Properties} \\ \hline\hline
    color/colors                                                                            & 11 colors (e.g. \textit{'black', 'blue',}...) \\ \hline
    age\footnote{\label{note:FaceBased}Requires face detection.}                            & ages and ages inequalities (based on 8 age groups) \\ \hline
    gender\footnoteref{note:FaceBased}                                                      & female/male \\ \hline
    location\footnote{\label{note:LocAmb}Binary spatial properties are treated either as relative to other objects (e.g.$\,${\fontfamily{put}\selectfont {\scriptsize 'the right'}}), or as global (e.g.$\,${\fontfamily{put}\selectfont {\scriptsize 'top'}}).} (e.g. where)              & spatial image location (e.g. \textit{'bottom (of the image)'}) \\ \hline
    relative location\footnoteref{note:LocAmb}\footnote{Doesn't have a function property.}  & location relative to other objects (e.g. \textit{'the \textbf{left} dog'}) \\ \hline
    type & subclass (when available) \\ \hline
    size & \textit{'small', 'big', 'average'} \\ \hline
    quantity\footnote{\label{note:PropOfSet}A property of an objects' set} & number of objects \\ \hline
    difference\footnoteref{note:PropOfSet} (odd man out) & no direct binary property \\ \hline 
    similarity\footnoteref{note:PropOfSet} & no direct binary property  \\ \hline 
   \end{tabular}
\caption[Properties table]{Table of supported properties for single objects and objects' sets.}
\label{table:usedProps}
\end{centering}
\end{minipage}
\end{table}

\item \textbf{Object Relations}

Relations between two objects are represented by the directed graph edges. Detection of relations varies and require ``simple'' information for some (e.g. 'to the right of') and complicated visual features for others (e.g. 'wearing'). We combine specific rule based detection for some relations and a deep neural network for others.

\begin{itemize}
\item \textbf{Rule based relation classification:} Based on spatial checks, using (when needed) morphological methods, depth estimation \cite{Depth2015Liu}, face detection \cite{Mathias2014Eccv}, face key points detection \cite{zhu2012face} and gaze estimation \cite{nips15recasens}. Simplifications, compositions of relations are used, as well as exploiting commonsense knowledge (by querying ConceptNet \cite{Speer2013}). A special type of relations are the comparison relations, \textit{sim\_\textless$f$\textgreater} and \textit{diff\_\textless$f$\textgreater}, that checks similarity or difference of function property $f$ correspondingly.

\item \textbf{Deep neural network classifier:} Based on the DR-Net method \cite{dai2017detecting} for the relation predicate classification. This method, as other visual relation detectors, utilizes object detection. To avoid coupling of relation detection with object detection, which would reduce the robustness of our system, and yet exploit object detection when possible, we've added a layer that was trained to project a closeness measure based on the GloVe word embedding \cite{Pennington14glove:global} and generate a representation for any object class. This way, object classes that were not trained for the relation classification still have a representation projected on the DR-Net object classes vector. We use the version trained for the 70 VRD dataset \cite{lu2016visual} relations.

\end{itemize}

\end{itemize}

Since relations are also used as an attention for object detection (\ref{sec:GuidedDet}), inverse relations are matched to each relation, when possible. This way, attention can be used for both directions of the relation.

\subsubsection{Recursive Procedure}

The final stage of answering the question is activating a recursive procedure to follow the graph nodes and edges, invoke the relevant basic procedures and integrate all the information to provide the answer. A basic scheme of the procedure is given in Figure \ref{fig:procedure} and in Algorithm \ref{alg:procedure}.

\pgfdeclarelayer{background}
\pgfdeclarelayer{foreground}
\pgfsetlayers{background,main,foreground}


\tikzstyle{bp}  = [draw , fill=  green!20, text width=  5em, text centered, minimum height=2.5em,drop shadow]
\tikzstyle{vis} = [bp   , fill=   blue!20, text width=3.6em, minimum height=1.4em]
\tikzstyle{ann} = [above, text width= 5em, text centered]
\tikzstyle{op}  = [bp   , fill=    red!20, text width= 8em, minimum height=11em, rounded corners, drop shadow]
\tikzstyle{ex}  = [op   , fill= orange!20, text width= 5em, minimum height=5em]
\tikzstyle{im}  = [ex   , minimum height=6.7em, text width= 5.6em]
\tikzstyle{ch}  = [vis  ,fill= brown!15]
\tikzstyle{ec}  = [ch   ,fill= brown!25]

\tikzstyle{obj}   = [bp   ,rounded corners, minimum height=1.5em]
\tikzstyle{det}   = [vis  ,rounded corners, text width=5em, minimum height=1.5em]

\def\blockdist{2.3}
\def\edgedist{2.5}

\begin{figure}[ht!]
\begin{center}

\begin{tikzpicture}

    \node (op) [op] {};
    \path (op.north)+(0,-0.35) node (optxt) {\footnotesize{Object-wise Analysis}};

    \path (op.north)+(-0.1*\blockdist, \blockdist*0.9) node (det)   [det] {\tiny{object detection}};
    \path (op.north)+(1.2*\blockdist, \blockdist*0.9) node (curNode)   [det, fill= teal!15] {\tiny{set cur\_node}};

    \path (op.north)+(0, \blockdist*0.3) node (objs) [obj] {\footnotesize{get objects}};

    \path (op.north)+(0.1,  -\blockdist*0.5 + 0.1) node (p1)   [obj] {check $p_i$};
    \path (op.north)+(0.05,  -\blockdist*0.5 + 0.05) node (p2)   [obj] {check $p_i$};
    \path (op.north)+(0,  -\blockdist*0.5) node (p3)   [obj] {check $p_i$};

    \path (op.north)+(0.1,  -\blockdist*0.85 + 0.1) node (r1)   [obj] {check $r_i$};
    \path (op.north)+(0.05,  -\blockdist*0.85 + 0.05) node (r2)   [obj] {check $r_i$};
    \path (op.north)+(0,  -\blockdist*0.85) node (r3)   [obj] {check $r_i$};

    \path (op.north)+(\blockdist*1.3+0.1,  -\blockdist*0.85 + 0.1) node (d1)   [det, dashed] {\tiny{detect daughter}};
    \path (op.north)+(\blockdist*1.3+0.05,  -\blockdist*0.85 + 0.05) node (d2)   [det, dashed] {\tiny{detect daughter}};
    \path (op.north)+(\blockdist*1.3,  -\blockdist*0.85) node (d3)   [det, dashed] {\tiny{detect daughter}};

    \path (op.north)+(0,  -\blockdist*1.2) node (f)   [obj] {get $f$};
    \path (op.north)+(0,  -\blockdist*1.55) node (c)   [obj] {check $c$};
    \path (op.north)+(0,  -\blockdist*2.15) node (g)   [obj] {get $g$};

    \path (op.west)+(-\blockdist, 2.09) node (graph) [im, minimum height=1.4em] {\footnotesize{question graph}};

    \path (op.west)+(-\blockdist, -0.9) node (image) [im] {};
    \path (image.north)+(0,-0.4) node (imtxt) {\footnotesize{Extended Image}};

    \path (imtxt.north)+(0, -\blockdist*0.38) node (rgb)   [ch] {\footnotesize{{\color{red}R}, {\color{green}G}, {\color{blue}B}}};
    \path (imtxt.north)+(0, -\blockdist*0.615) node (depth) [ec] {\footnotesize{Depth}}; 
    \path (imtxt.north)+(0, -\blockdist*0.84) node (color) [ec] {\footnotesize{Color}};

    \path [draw, shorten >=0.5cm, ->, >=stealth] (image.east) -- (-1.68, -0.9) node [above] {} (op) ;
    \path [draw, shorten >=0.5cm, <-, >=stealth] (depth.east) -- (-1.68, -1.14) node [above] {} (op) ;
    \path [draw, shorten >=0.5cm, <-, >=stealth] (color.east) -- (-1.68, -1.67) node [above] {} (op) ;

    \path [draw, shorten >=0.5cm, ->, >=stealth] (graph.east) -- (-1.68, 2.09) node [above] {} (op) ;
    \draw [draw, shorten >=0.5cm, ->, >=stealth, rounded corners=2mm] (graph) + (1.48, 0) -- (-2.5, 5) -- (2.8, 5) -- (2.8, 4);

    \path (d1.east)+( \blockdist, 1.6) node (knowledge) [ex] {External Knowledge};
    \path (d1.east)+( \blockdist, -2.1) node (mem) [ex] {Working Memory};

    \path (op.south) +(3.55, 3.9) node[text width=2cm, minimum height=6em] (dummy1) {};
    \path [draw, ->, >=stealth] (knowledge) edge[bend right= 20] node [left] {} (dummy1);
    \path [draw, <-, >=stealth] (knowledge) edge[bend right=-20] node [left] {} (dummy1);

    \path (op.south) +(3.55, 0.2) node[text width=2cm, minimum height=6em] (dummy2) {};
    \path [draw, ->, >=stealth] (mem) edge[bend right= 20] node [left] {} (dummy2);
    \path [draw, <-, >=stealth] (mem) edge[bend right=-20] node [left] {} (dummy2);

    \path [line] (det) -- +(0, -0.77) (objs);
    \path [line] (curNode) -- +(0, -0.77) (objs);
    \path [line] (objs) -- (op);
    \path [line] (op) -- (g);
    \path [line] (r2) -- (d2);

    \begin{scope}[line, rounded corners=2mm]
        \draw (d2) -- (8, 0.2) -- (8, 4.2)  -- (curNode);
        \draw (op.south) + (1.3, 0) -- (1.3, -4) --  (8, -4) -- (8, 0.2) -- (8, 4.2)  -- (curNode);
    \end{scope}

    \begin{pgfonlayer}{background}
        \path (op.west |- objs.north)+(-0.5,0.3) node (a) {};
        \path (g.south -| d1.east)+(+0.5,-0.3) node (b) {};

        \path[fill=yellow!20,rounded corners, draw=black!50] 
            (a) rectangle (b);

    \end{pgfonlayer}

\end{tikzpicture}

\end{center}
\caption[procedure]{A scheme of the recursive answering procedure. At each step the current node (cur\_node) is set and the objects are examined according to node's requirements. If succeeded, a new cur\_node is set (according to a relation or next global parent node) and the function is called again to handle the subgraph starting from it.  The required visual elements: $c$: object class, $p_i$: an object property, $f$: function property, $g$: property of a set, $r_i$: a relation. The daughter object detection is activated only when none was detected in previous stages. Note that the estimated maps of depth and color names are calculated by the procedure according to needs.}
\label{fig:procedure}
\end{figure}
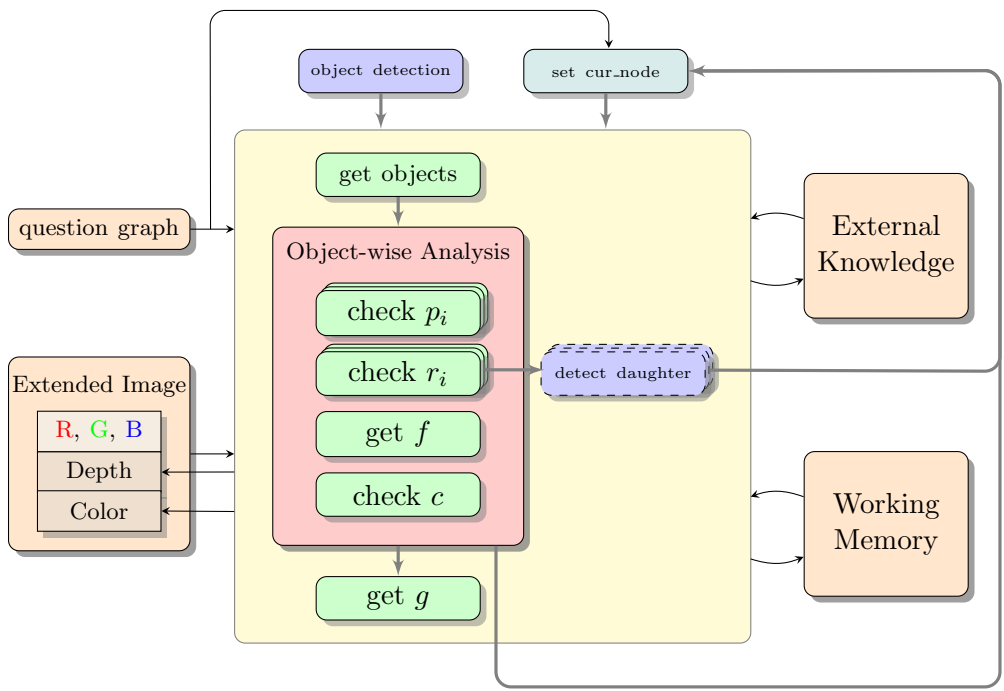


The first step is a preliminary object detection, carried out by applying instance segmentation \cite{he2017maskrcnn} on the image. Then, a recursive function ($getGraphAnswer$) is invoked for node handling (starting at a global parent node). It runs specific procedures that activate visual analyzers to check the requirements (properties, relations) and fetch required information (function property). The retrieved objects that fulfill the requirements are coupled to the corresponding question objects, so that next checks would be held on the same objects. The number of required objects is mainly according to quantifiers. Once a node checks are completed, the same function ($getGraphAnswer$) is invoked for next node. Next node is determined according to relation (graph edge) or next global parent node. Once all nodes are queried, the checks for entire set are activated (if needed). Answers are provided by all basic procedures and final answer is set according to precedence (e.g. queried property type has a priority over binary answers).


\begin{algorithm}[h!]
\scalebox{0.75}{
\begin{minipage}{\textwidth}
\SetAlgoLined
\KwIn{question\_graph, image}
\KwResult{Answer to question}
 initialization: run object detection, $current\_node = first\_parent\_node$\;
 $\boldsymbol{Run} [success, answer] = getGraphAnswer:$ \\
 \Begin{
 Node parameters: $\boldsymbol{p}$: properties, $\boldsymbol{r}$: relations, $f$: function property, $g$: property of a set, $\boldsymbol{obj}$: candidate objects\footnote{According to object detection and previous checks}\;
 \For{obj in \textbf{obj}}{
   \If{$\lnot$ empty($\boldsymbol{p}$)}{
     \For{$p$ in $\boldsymbol{p}$}{
       [$success, answer$] = $is\_p(obj)$ \;
       \textbf{if} $\lnot$ success \textbf{then} break \textbf{end}
     }
     \If{$\lnot$ success}{
     \eIf{\#possible\_objs \textless \#required\_objs\footnote{\label{note:requiredObjs}According to quantifiers and other requirements}} {return} {continue}
     }
   }
   \textbf{if} $\lnot$ empty($f$) \textbf{then} $answer = f(obj)$ \textbf{end}
   \eIf{empty($\boldsymbol{r}$)}{
     \If{exist(next\_parent\_node)}{
       $current\_node = next\_parent\_node$\;
       Run [$success, answer$] = getGraphAnswer\;
     }
   } {
      \For{$r$ in $\boldsymbol{r}$}{
        \textbf{if} empty($\boldsymbol{d\_objs}$) \textbf{then} Run detectObjsUsingRel($r$) \textbf{end}

        \For{$obj\_d$ in $\boldsymbol{d\_objs}$ (candidate objects for daughter nodes)}{
          [$success, answer$] =  $is\_r(obj\_d, obj)$ \;
          \If{success}{
            $current\_node = next\_node$\footnote{Either to daughter node or next global parent node}\;
            Run [$success, answer$] = getGraphAnswer\;
            \textbf{if} success $\land$ (\#success\_objs\_d == \#required\_objs\_d\footnoteref{note:requiredObjs}) \textbf{then} break \textbf{end}
          }
            \textbf{if} $\lnot$ success $\land$ (\#possible\_objs\_d \textless \#required\_objs\_d\footnoteref{note:requiredObjs}) \textbf{then} break \textbf{end}
          }
          \textbf{if} $\lnot$ success \textbf{then} break \textbf{end}
        }
     }
     \textbf{if} success \textbf{then} break \textbf{end}
   }
   \textbf{if} success $\land$ $\lnot$ empty($g$) \textbf{then} $answer = g(valid\_objs)$ \textbf{end}
   Return $answer$
 }
 \end{minipage}
 }
 \caption{Answering procedure according to graph}
 \label{alg:procedure}
\end{algorithm}


\paragraph{Working Memory}

The global information gathered through the answering process is stored in a "Working Memory" component. It stores the calculations that may be required at several stages of the process. This information is calculated only if needed and includes objects and their retrieved data, depth map, current node, currently used objects and more.

\paragraph{Common Knowledge}

When a person is answering a visual question, there is an important role to prior common knowledge. This includes connection between classes, famous brands and logos, knowing the role and characteristics of objects and actions, anticipation of the future, knowing to ignore details and more.

Some of the issues related to prior commonsense knowledge are addressed by our system. The main uses of prior knowledge are common relations in images (using the Visual Genome dataset \cite{krishna2017visual}) and  commonsense knowledge on categories of objects, as well as connections between them (using ConceptNet \cite{Speer2013}).

\begin{itemize}
\item \textbf{Visual Genome Dataset}
\label{sec:visGenome}

The Visual Genome dataset \cite{krishna2017visual} contains (among many others) annotations for objects and binary relations between them for a set of 108077 images. Common relations involving specific objects are extracted from this dataset (by demand) and used as prior knowledge to assist detection. It allows refining the search area when an object is not detected in the initial detection as described below and demonstrated in Figure \ref{fig:common_att}.

\item \textbf{ConceptNet}
\label{sec:conceptNet}
To obtain general commonsense knowledge we use ConceptNet database (version 5) \cite{Speer2013}. The source of information for this database is the internet (results from additional databases are also incorporated). It allows querying for concepts and relations between them of the form:
\begin{quote}
concept1 - \textit{relation} $\rightarrow$ concept2 \, (e.g. horse - \textit{IsA} $\rightarrow$ animal)
\end{quote}
 The query is performed by providing two of the triplet [relation, concept1, concept2] and querying for the third. These common knowledge relations provide complement capabilities for answering 'real world' questions in which such common knowledge is assumed. We currently use ConceptNet mainly to extend understanding of objects' classes (e.g. superordinate classes, similar classes) as described for example in section \ref{sec:superordinate}. Examples for questions are given in Figure \ref{fig:classTypes} for connections between classes.

\end{itemize}

\paragraph{Guided Object Detection}
\label{sec:GuidedDet}

A question may refer to specific objects in the image that may be hard to detect (e.g. due to size, occlusion, clutter). When a requested object is not detected on the first attempt (searching the entire image), additional attempts are made. These attempts focus on regions where the object has a higher probability to be found. We use relations with detected objects as an attention source. Two sources for such an attention are used.

\begin{itemize}

\item \textbf{Attention by common relations:} The source for this attention is from the Visual Genome dataset \cite{krishna2017visual}, where objects and relations between them are annotated in images. When a requested object is not detected on the first attempt (searching the entire image), additional attempts are made. These attempts focus on regions where the object has a higher probability to be found. This is done using the annotation of the Visual Genome dataset \cite{krishna2017visual}, where objects and relations between them are annotated in images (see also section \ref{sec:visGenome}). We seek the most common relation of the requested object (with an object from our known classes' set) and a corresponding relative location. Then, if the other object is found we activate the object detector on the relevant area. An additional search area is obtained by the relation's spatial constraints. An example of using common relations as an attention is given in Figure \ref{fig:common_att}.

\begin{figure} [h!]
\begin{centering}
\begin{tabular}{cc}
\subfigure{
  \includegraphics[totalheight=0.23\textheight]{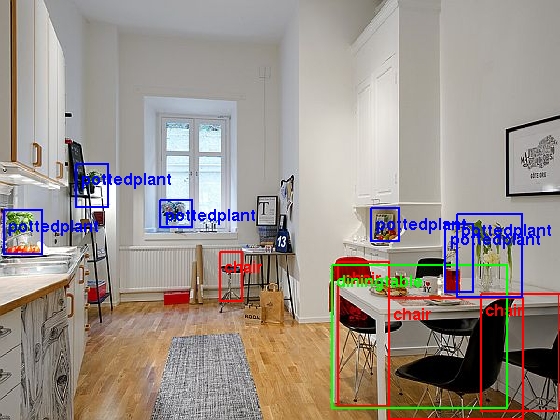}
} &
\subfigure{
  \includegraphics[totalheight=0.23\textheight]{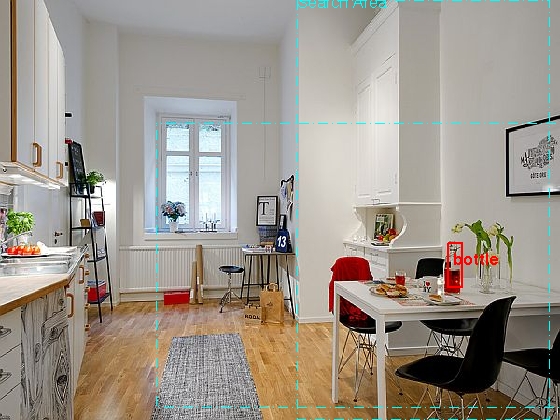}
}  \\
(a) Detection results on entire image & (b) Detected bottle using common relations \\[6pt]

\end{tabular}
\caption[common relation attention example]{Attention by common relations in answering the question \textit{``Is there a bottle?''} (a) Initial object detection on entire image did not detect the bottle. (b) An additional detection attempt is performed on search areas extracted by the common relation \textit{'bottle-on-diningtable (table)'}. [Object detection is based on faster R-CNN + DeepLab].}
\label{fig:common_att}
\end{centering}
\end{figure}

\item \textbf{Attention by question relations:} \label{sec:qRelAtt} The question itself may include relations that can assist detection by focussing on relevant areas. Since the processing is according to the question graph representation, relation edge directions are modified from detected to undetected objects. This allows using relations with a verified detected object as a detection guidance for undetected objects in the same manner described above. The usage of this type of attention is demonstrated in Figure \ref{fig:question_att}.

\begin{figure} [h!]
\begin{centering}
\begin{tabular}{cc}
\subfigure{
  \includegraphics[totalheight=0.23\textheight]{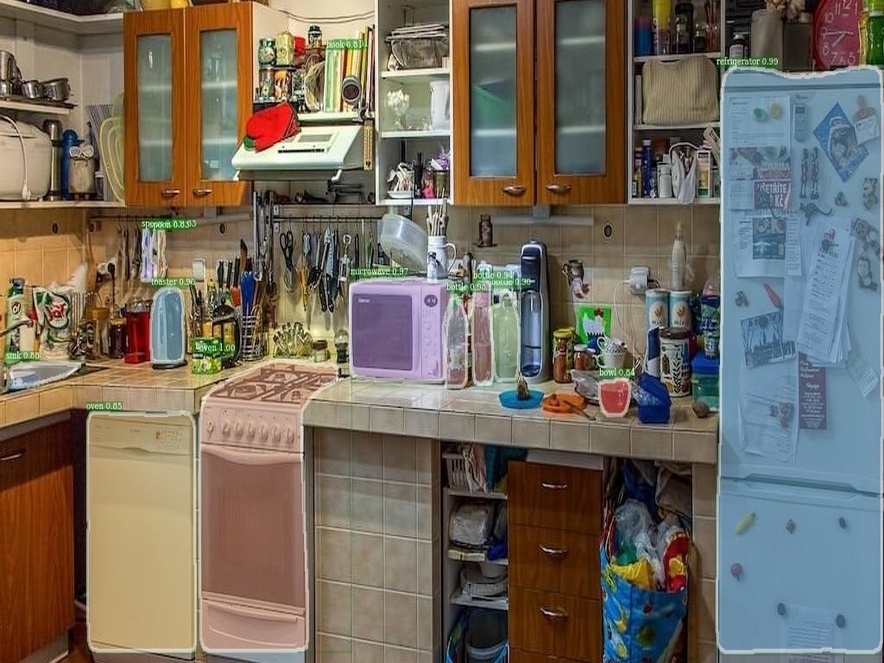}
} &
\subfigure{
  \includegraphics[totalheight=0.23\textheight]{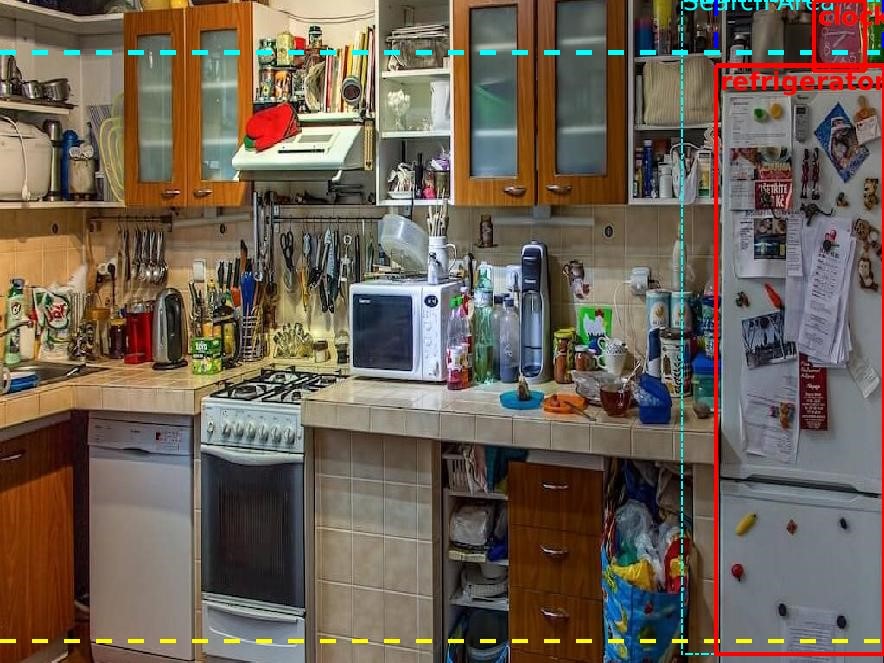}
}  \\
(a) Detection results on entire image & (b) Detected clock using question relation \\[6pt]

\end{tabular}
\caption[question attention example]{Attention by question relations in answering the question \textit{"Is there a clock above the refrigerator?"} and \textit{"Is there a refrigerator below the clock?"} (a) Initial object detection on entire image did not detect the clock. (b) An additional detection attempt is performed on search areas extracted by the question relation \textit{'clock-above-refrigerator'}.}
\label{fig:question_att}
\end{centering}
\end{figure}


\end{itemize}

\subsection{"Understanding" Capabilities} 

Having a system that breaks the visual answering task into real world sub tasks has many advantages. Other than abilities of modular modifications and improvements, the meaningful, compositional process is utilized and leveraged to provide information derived from internal processing. Failure reasons and verified alternatives are provided, as well as elaborations on detected objects.

\subsubsection{Provide Alternatives/Corrections}
\label{sec:alternatives}

When the logic expression representing the question is not valid for the given image, alternatives for the failed part are searched, such that a close expression may be validated and provided as a supplement to the answer. The checks include alternative objects, relations and also properties according to the following:
\begin{itemize}
  \item For failed object classes alternative classes are checked.
  \item Real properties are specified for objects with failed properties.
  \item For failed relations alternative relations are checked.
  \item Additional attempts with close person’s subordinate classes (e.g. when failed to classify a person as a woman, other sub-person classes are checked).
\end{itemize}

\begin{figure} [h!]
\begin{centering}
\scalebox{0.9}{
\setlength\tabcolsep{1.5pt} 
\renewcommand{\arraystretch}{0.8}
\begin{tabular}{p{3.7cm}p{3.7cm}p{3.7cm}p{3.7cm}}
\subfigure{
  \includegraphics[totalheight=0.13\textheight]{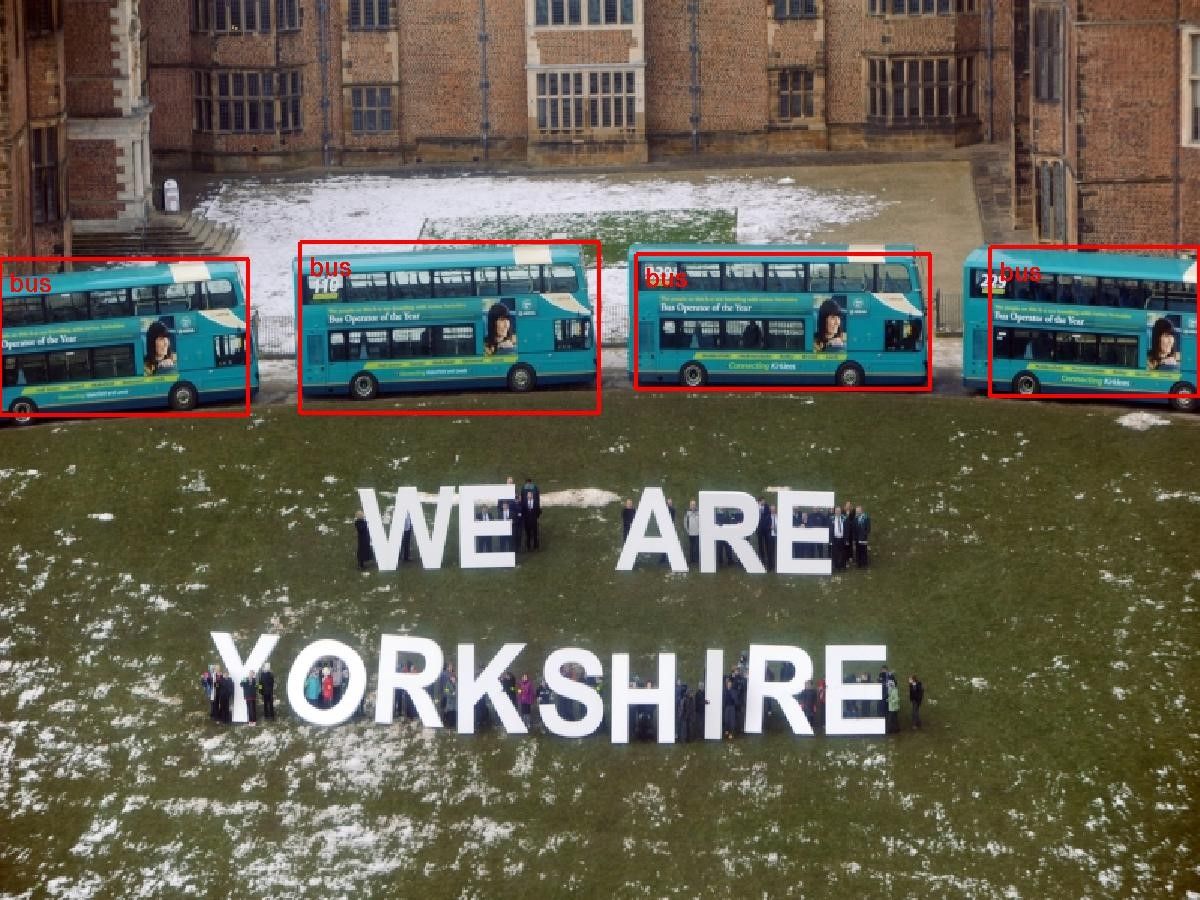}
} &
\subfigure{
  \includegraphics[totalheight=0.13\textheight]{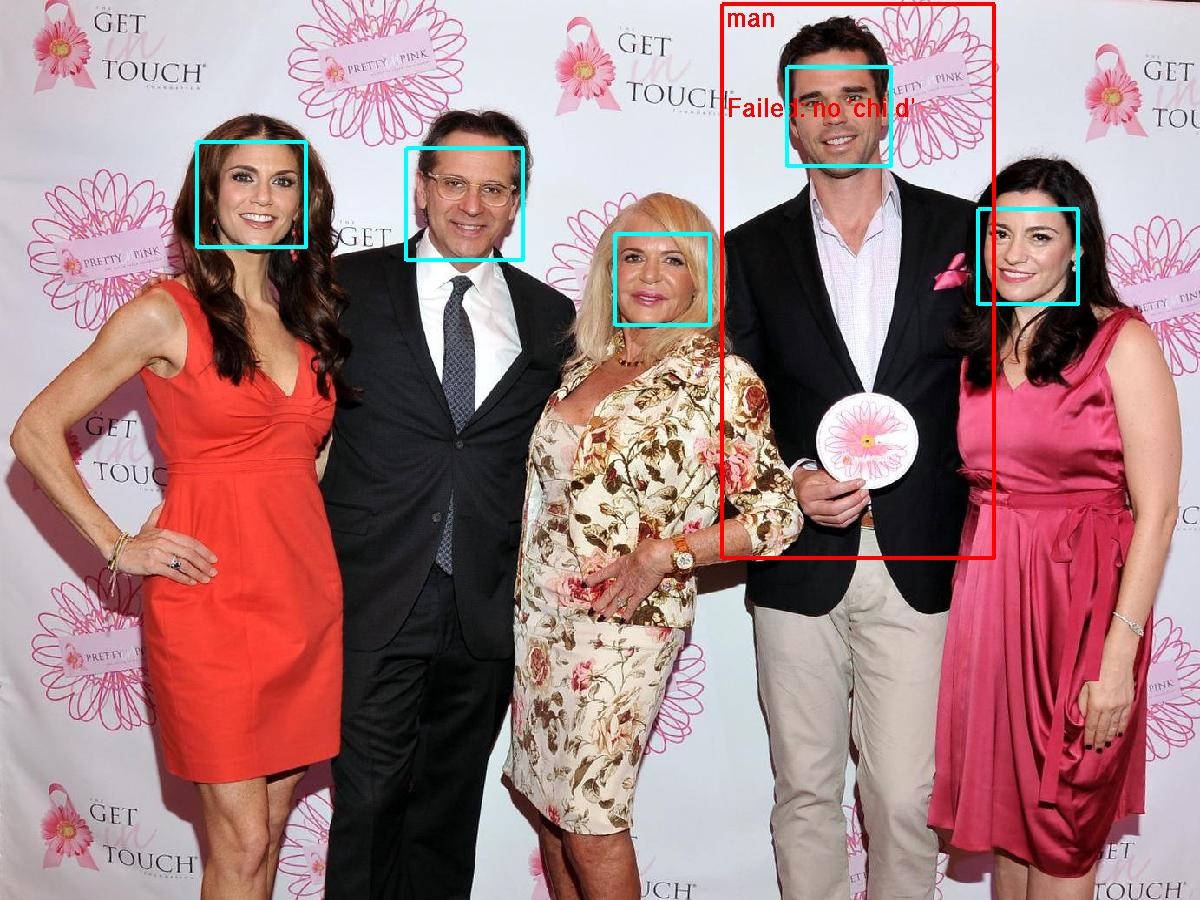}
} &
\subfigure{
  \includegraphics[totalheight=0.13\textheight]{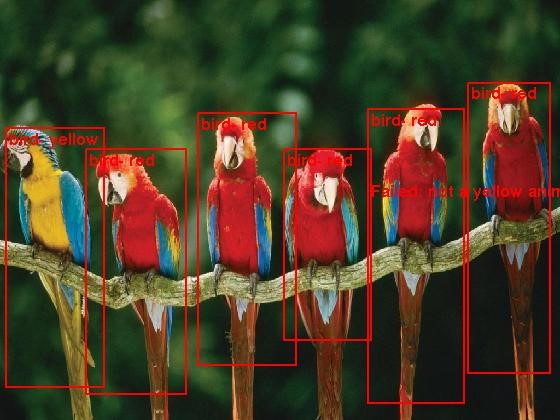}
} &
\subfigure{
  \includegraphics[totalheight=0.13\textheight]{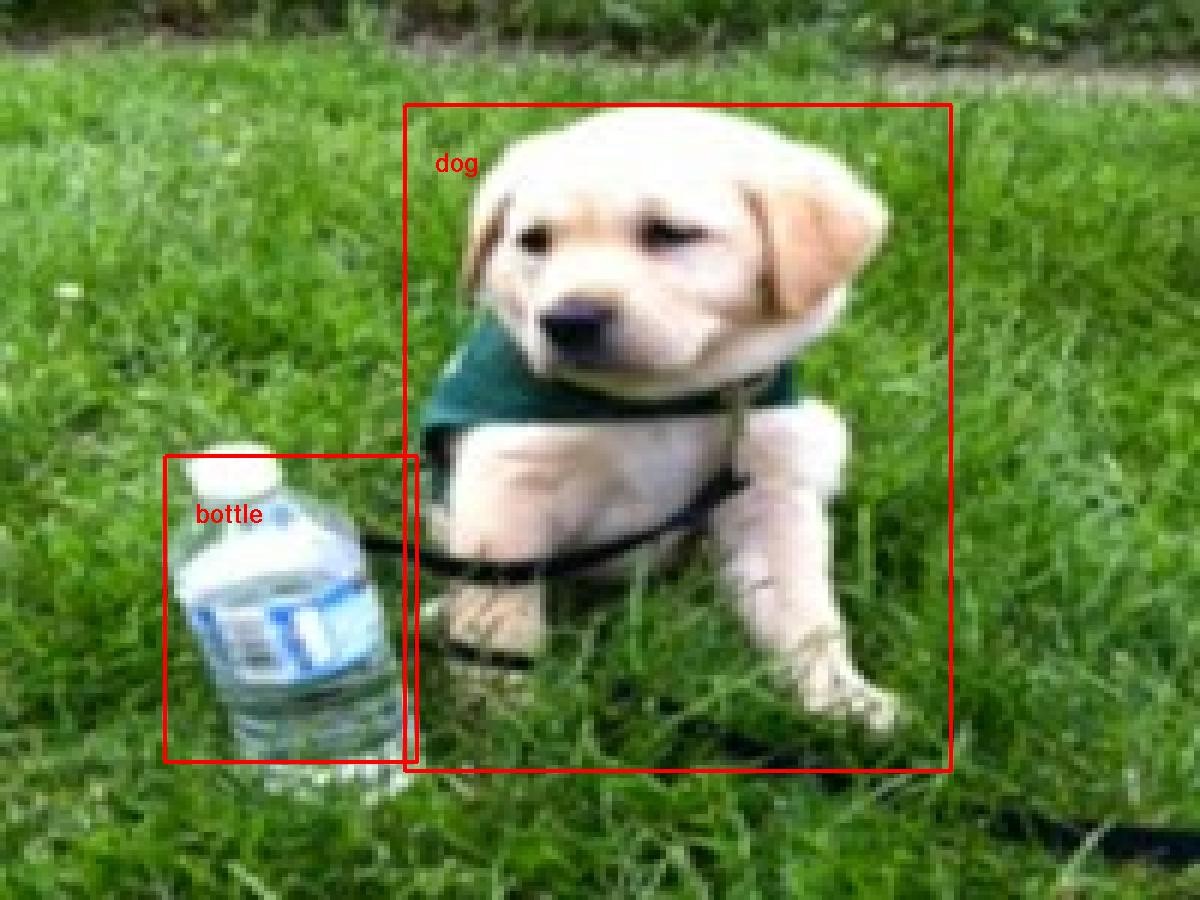}
}\\

\tinyPut{\textbf{Q:} Is there a train in the image?} & \tinyPut{\textbf{Q:} Is there a child in the image?} & \tinyPut{\textbf{Q:} Are there a two yellow animals?} & \tinyPut{\textbf{Q:} Is there a bottle to the right of a dog?} \\
\tinyPut{\textbf{A:} There is no train} & \tinyPut{\textbf{A:} Couldn't find any object of class:} & \tinyPut{\textbf{A:} There are not enough yellow animals} & \tinyPut{\textbf{A:} There are no bottles to the right of a } \\
\tinyPut{\Aindent There is a bus}       & \tinyPut{\Aindent child (failed subclasses: }            & \tinyPut{\Aindent (failed due to 5 red birds), }          & \tinyPut{\Aindent dog} \\
                                        & \tinyPut{\Aindent 3 women and 2 men)}                    & \tinyPut{\Aindent where bird is a subclass of animal}     & \tinyPut{\Aindent Existing alternative relations:} \\
                                        & \tinyPut{\Aindent There is a man}                        &                                                           & \tinyPut{\Aindent 'bottle to the left of a dog'} \\
                                        &                                                          &                                                           & \\

\subfigure{
  \includegraphics[totalheight=0.13\textheight]{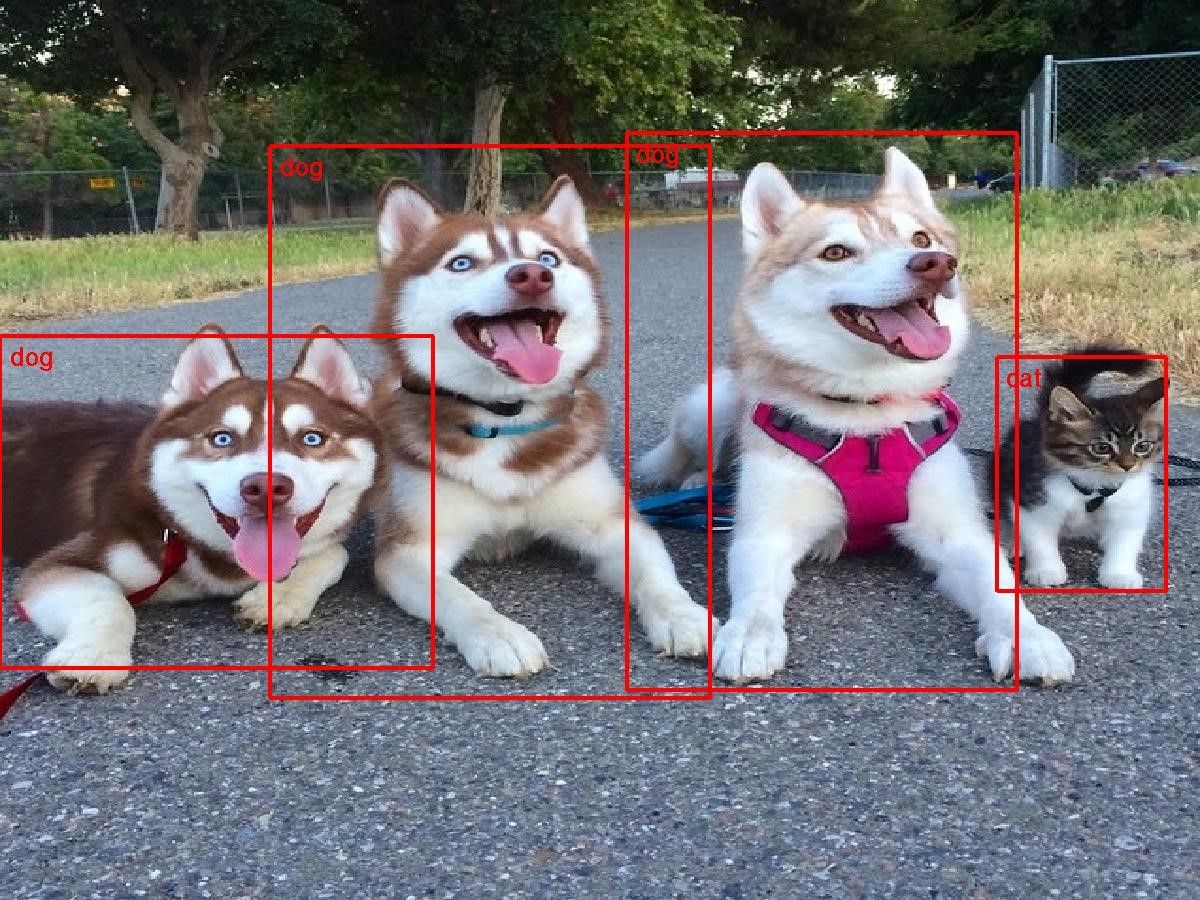}
} &
\subfigure{
  \includegraphics[totalheight=0.13\textheight]{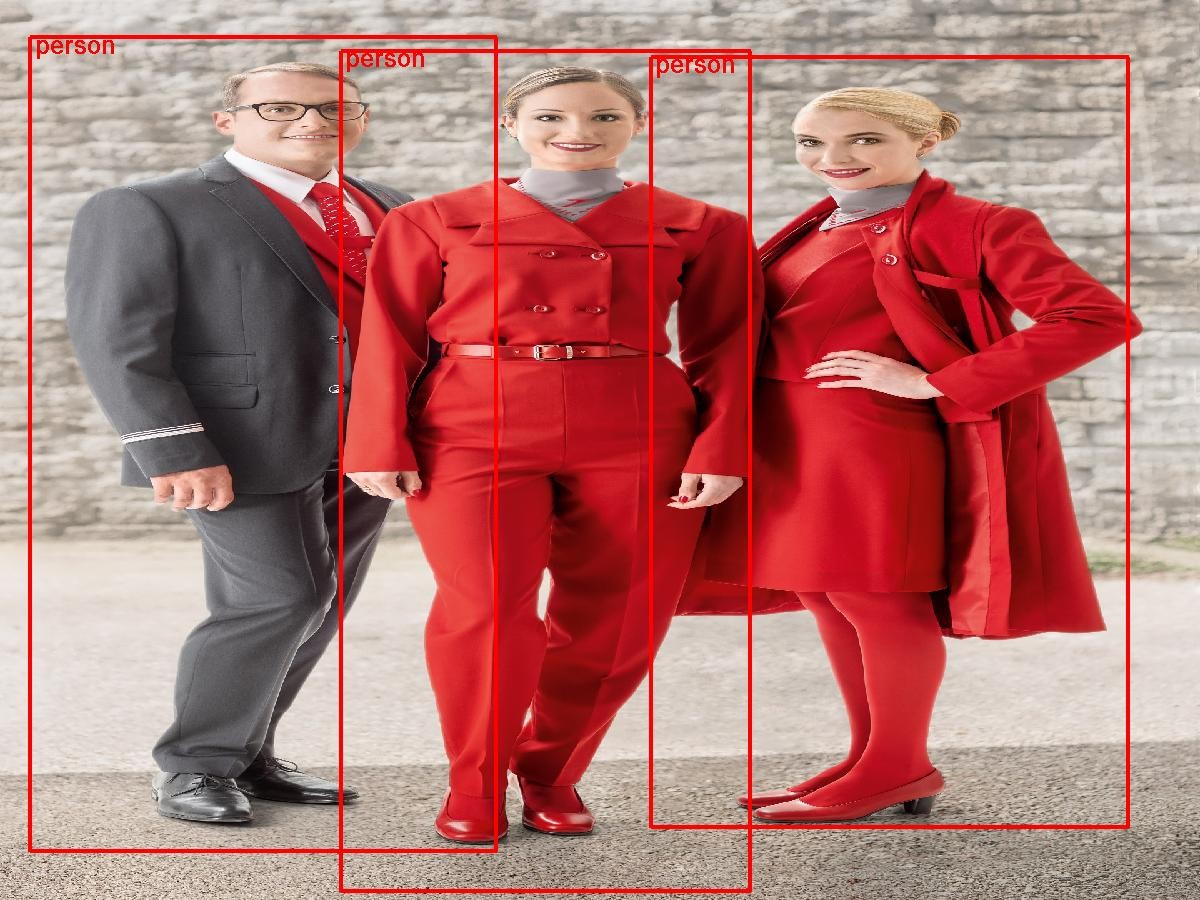}
} &
\subfigure{
  \includegraphics[totalheight=0.13\textheight]{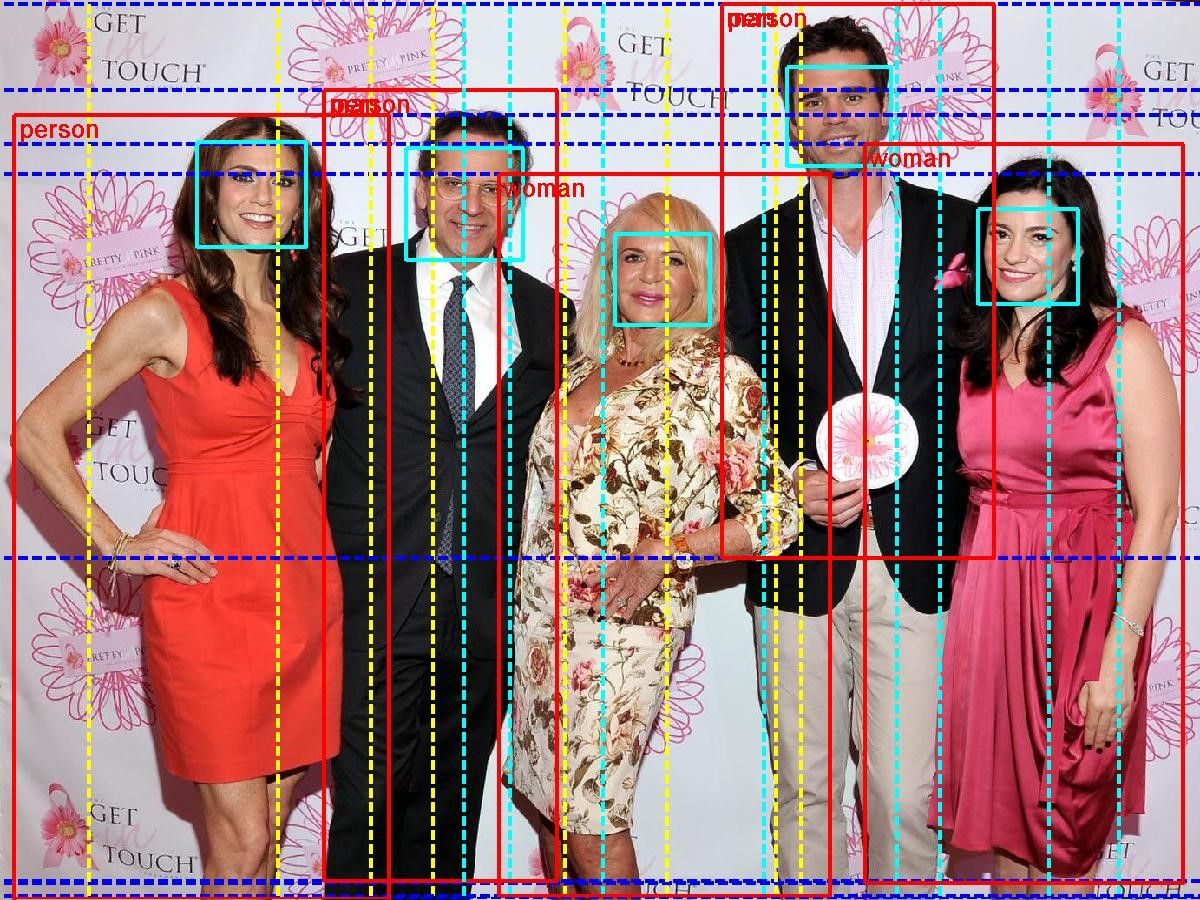}
} &
\subfigure{
  \includegraphics[totalheight=0.13\textheight]{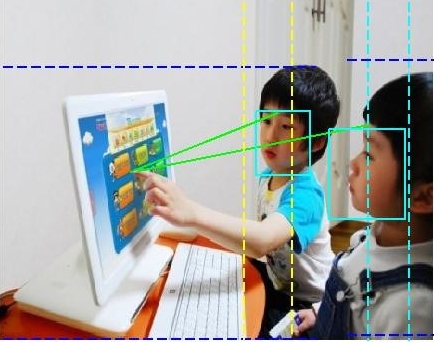}
} \\

\tinyPut{\textbf{Q:} Are there 2 dogs behind a cat?} & \tinyPut{\textbf{Q:} Is there a baby below a boy?} & \tinyPut{\textbf{Q:} Are there three women to the right} & \tinyPut{\textbf{Q:} Is there a woman looking at a man?}\\
&     &   \tinyPut{\Aindent  of a man?}  & \\
\tinyPut{\textbf{A:} There are not enough dogs behind} & \tinyPut{\textbf{A:} There are no babies below a boy} & \tinyPut{\textbf{A:} There are not enough women to the} & \tinyPut{\textbf{A:} There are no women looking at}\\
\tinyPut{\Aindent a cat}                           &  \tinyPut{\Aindent Existing alternative relations: }    & \tinyPut{\Aindent right of a man (failed subclasses: } & \tinyPut{\Aindent a man (failed subclasses: a boy)} \\
\tinyPut{\Aindent Existing alternative relations:} &  \tinyPut{\Aindent 2 'person to the right of a person'} & \tinyPut{\Aindent 2 men and 2 women)}                  &\tinyPut{\Aindent Couldn't find any object of class: }\\
\tinyPut{\Aindent 2 'dog to the left of a cat'}    &                                                         & \tinyPut{\Aindent There are not enough women to the }  & \tinyPut{\Aindent woman (failed subclasses: a boy)} \\
                                                   &                                                         & \tinyPut{\Aindent right of a person (superordinate  }  & \tinyPut{\Aindent There are no people (superordinate }\\
                                                   &                                                         & \tinyPut{\Aindent class)}                              & \tinyPut{\Aindent class) looking at a person } \\
                                                   &                                                         & \tinyPut{\Aindent There are 3 people (superordinate }  & \tinyPut{\Aindent (superordinate class)} \\
                                                   &                                                         & \tinyPut{\Aindent class) to the right of a person}     & \tinyPut{\Aindent Existing alternative relations:} \\
                                                   &                                                         & \tinyPut{\Aindent (superordinate class)}               & \tinyPut{\Aindent 'person to the right of a person'}
\end{tabular}
}
\caption[Answer alternative examples]{Answer alternative examples [Object detection is based on faster R-CNN + DeepLab].}
\label{fig:alternatives}
\end{centering}
\end{figure}

Examples are given in Figure \ref{fig:alternatives} (note that some include multiple rounds of attempts).

\subsubsection{Answer Elaboration}

During the answering process, related information may be accumulated for verifying the logical expression representing the question. This information is provided as part of the answer, explaining and elaborating it.
The following supplementals are included:
\begin{itemize}
  \item If object detection was by a related class (e.g. synonym, parts of a group, subordinate classes), it is specified in the answer (including numbers of each subclass).
  \item The hint relation used as an attention for object detection, is indicated (if used).
  \item If queried function properties (e.g. color) are different for different relevant objects, property for each object is specified.
\end{itemize}

\begin{figure} [h!]
\begin{centering}
\setlength\tabcolsep{1.5pt} 
\renewcommand{\arraystretch}{0.8}
\begin{tabular}{p{5cm}p{5cm}p{5cm}}
\subfigure{
  \includegraphics[totalheight=0.165\textheight]{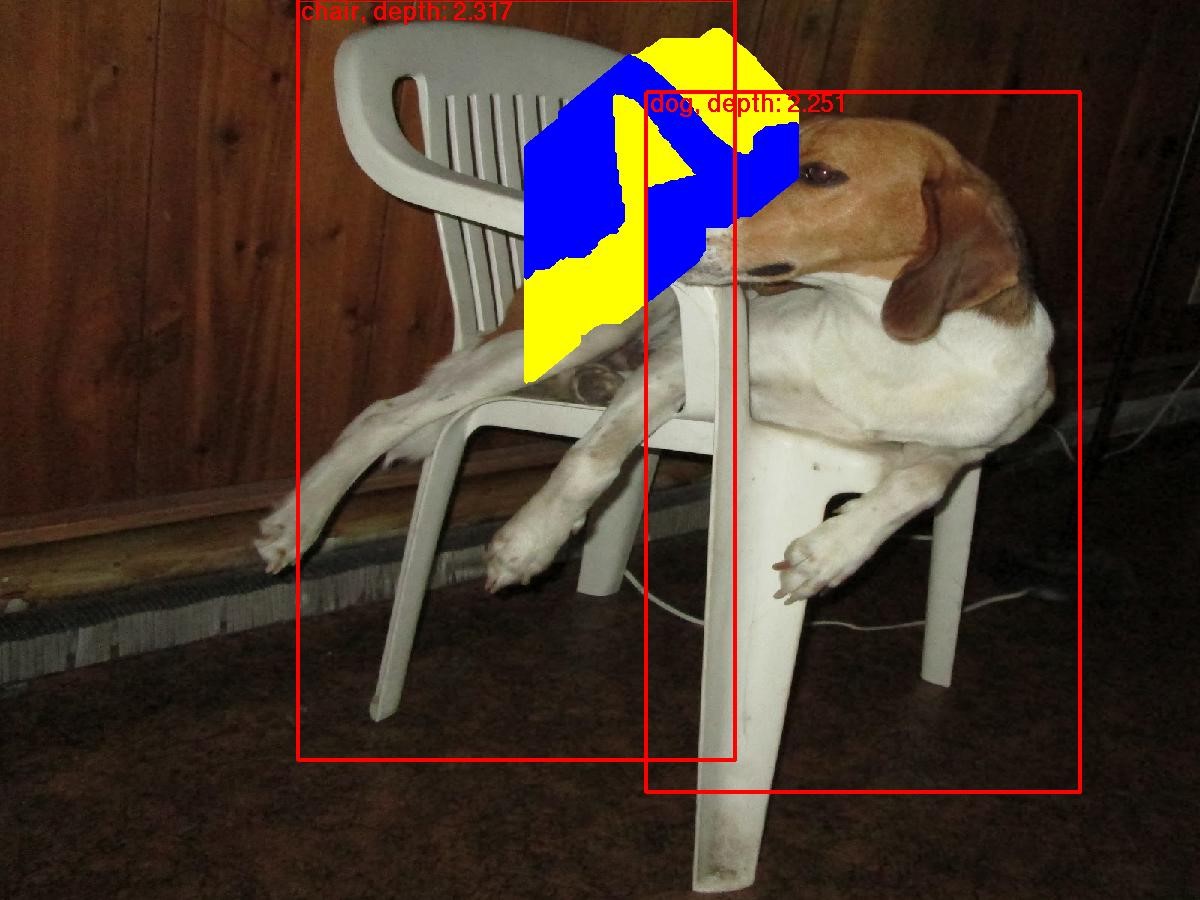}
} &
\subfigure{
  \includegraphics[totalheight=0.165\textheight]{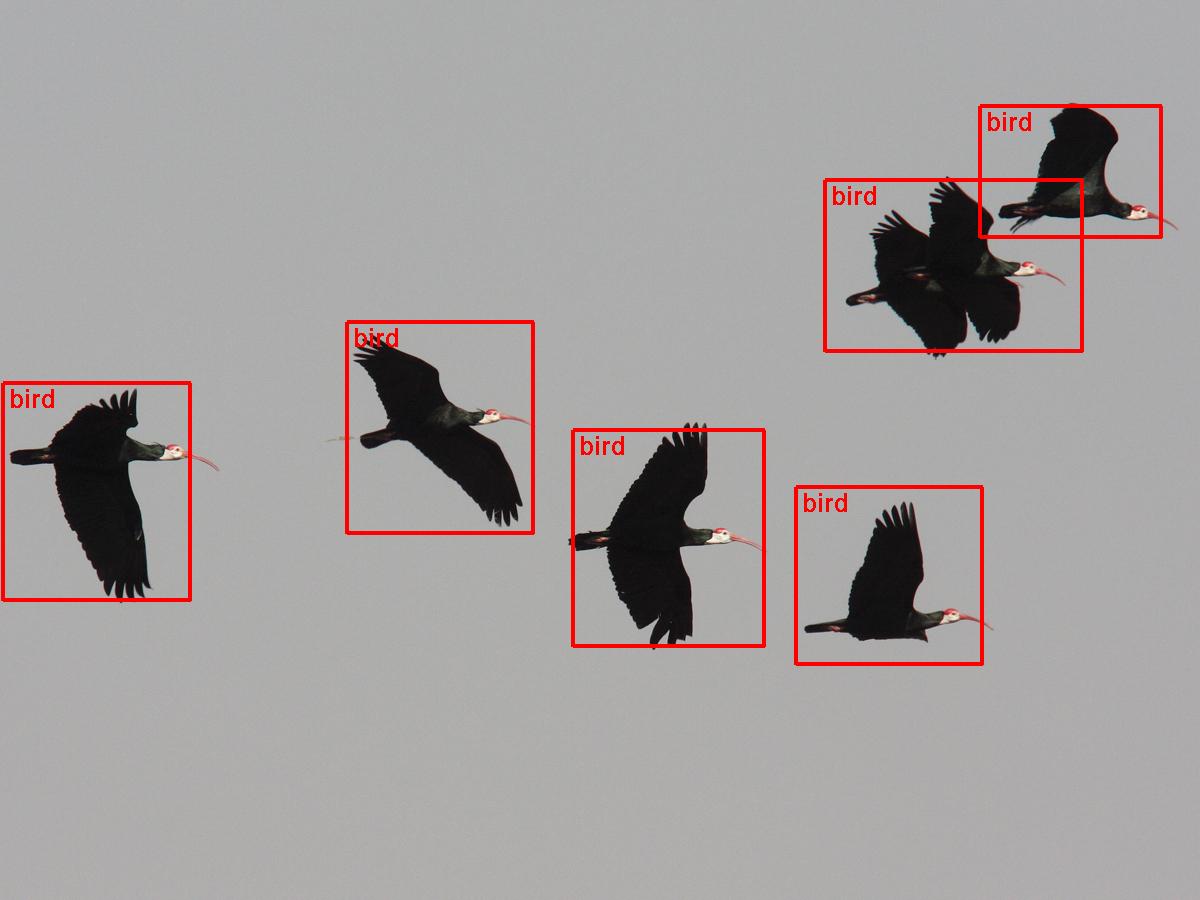}
} &
\subfigure{
  \includegraphics[totalheight=0.165\textheight]{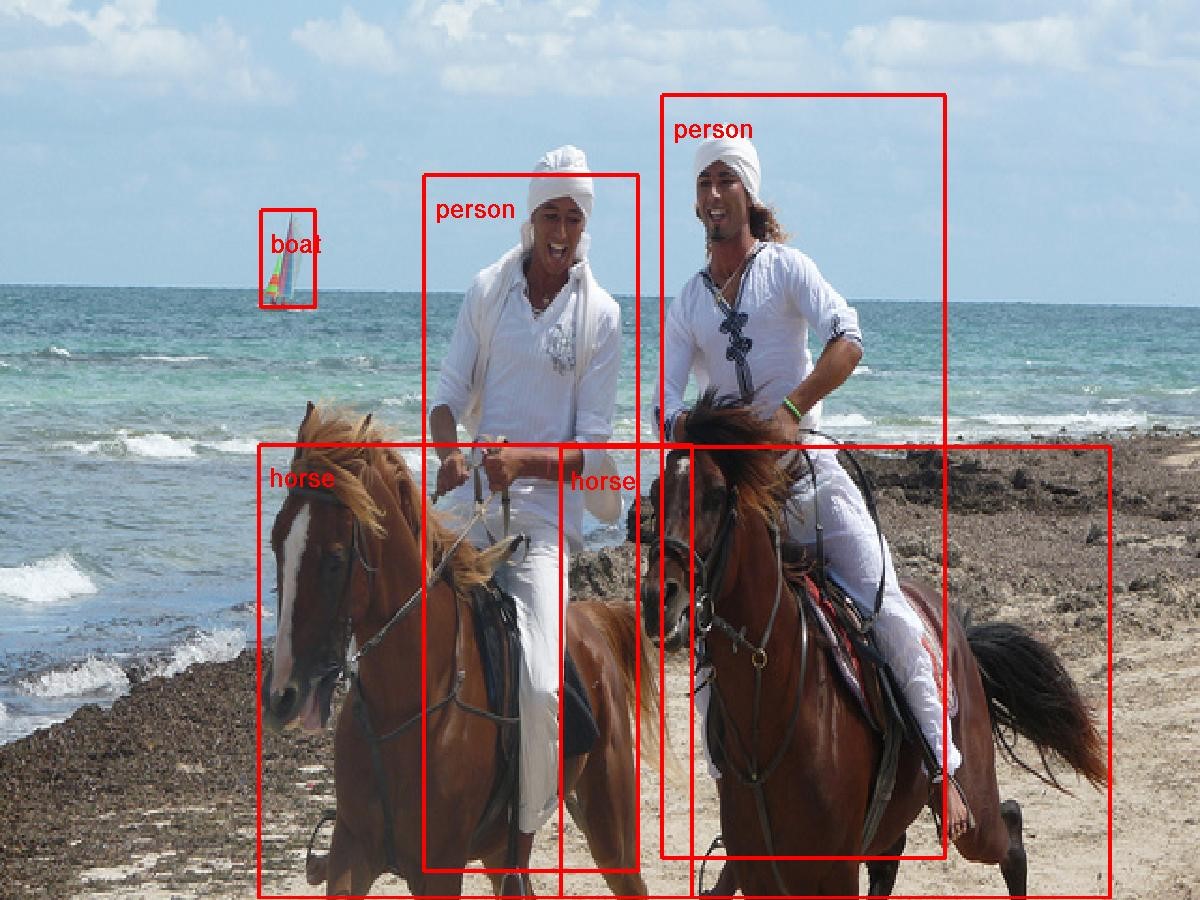}
} \\

\tiny{\textbf{Q:} Is there a hound near a chair?} & \tiny{\textbf{Q:} Is there a flock?} & \tiny{\textbf{Q:} Are there animals in the image?} \\
\tiny{\textbf{A:} Yes, there is a dog near a chair, } &  \tiny{\textbf{A:} Yes, there are 2 birds (at least 2 birds), } &  \tiny{\textbf{A:} Yes, there are horses, where horse is } \\
\tiny{\Aindent where dog is a synonym of hound} & \tiny{\Aindent where bird is a part of a flock} & \tiny{\Aindent a subclass of animal} \\

\subfigure{
  \includegraphics[totalheight=0.165\textheight]{catsAndDogs_animal}
} &
\subfigure{
  \includegraphics[totalheight=0.165\textheight]{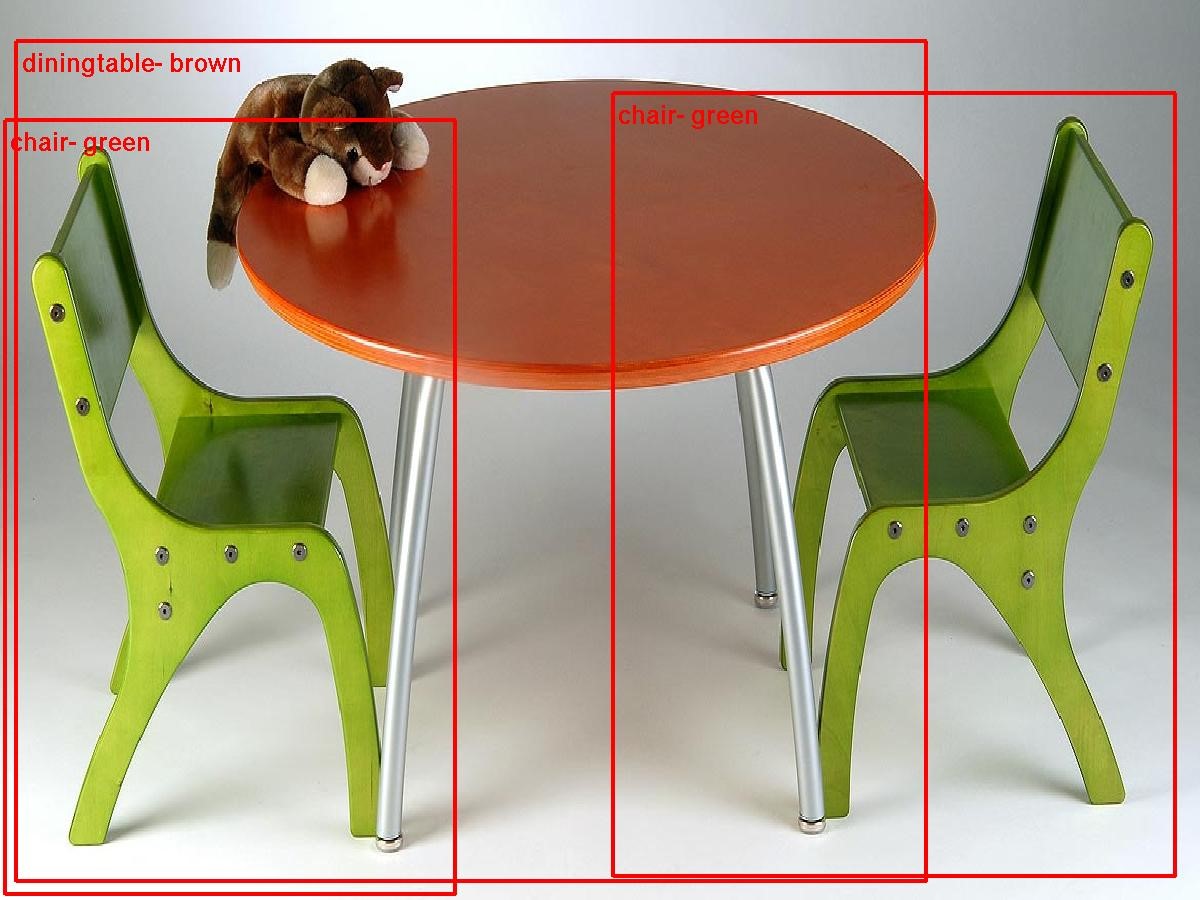}
} &
\subfigure{
  \includegraphics[totalheight=0.165\textheight]{bottle_on_table7_bottle-full}
} \\

\tiny{\textbf{Q:} How many animals are there?} & \tiny{\textbf{Q:} What is the color of the furniture?} & \tiny{\textbf{Q:} Is there a bottle?} \\
\tiny{\textbf{A:} The number of the animals: 4, } &  \tiny{\textbf{A:} The color of the chair: green} &  \tiny{\textbf{A:} Yes, there is a bottle ('bottle ' was } \\
\tiny{\Aindent number per sub group: dog: 3, cat: 1} &  \tiny{\Aindent The color of the chair: green} & \tiny{\Aindent detected according to "hint" relation: }\\
                                                     &  \tiny{\Aindent The color of the diningtable: brown} & \tiny{\Aindent 'on diningtable') }

\end{tabular}
\caption[Answer elaboration examples]{Answer elaboration examples [Object detection is based on faster R-CNN + DeepLab].}
\label{fig:elaboration}
\end{centering}
\end{figure}

Some examples can be seen in Figure \ref{fig:elaboration}.

\subsubsection{Integration in Related Applications} 

\begin{figure} [h!]
\begin{centering}
\begin{tabular}{cc}
\subfigure{
  \includegraphics[totalheight=0.23\textheight]{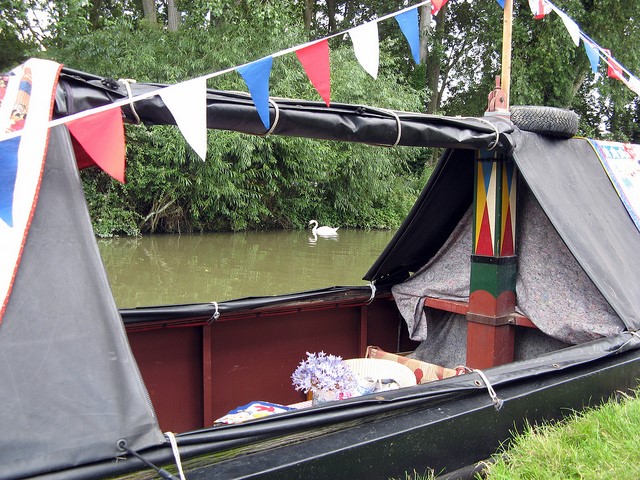}
} &
\subfigure{
  \includegraphics[totalheight=0.23\textheight]{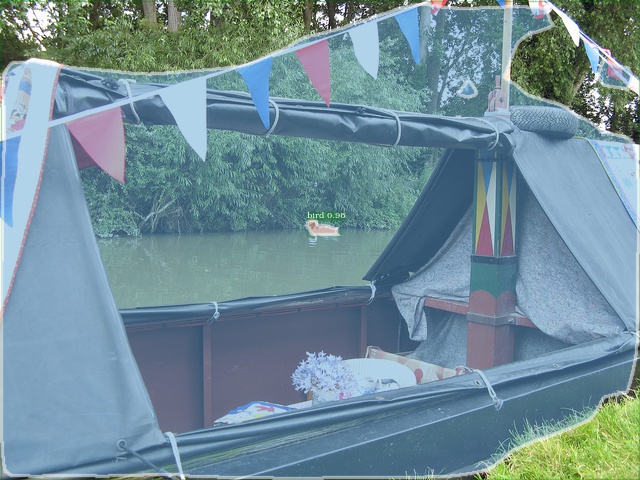}
}  \\
\end{tabular}
\end{centering}

 \textbf{Caption:} a man sitting on a bench with a large umbrella \\
 \textbf{Q:} Is there a man on a bench with a large umbrella? \\
 \textbf{A:} There is no bench \\
 \Aindent There is no man \\
 \Aindent There is no umbrella \\
 \Aindent There is nothing on a object \\
 \Aindent Existing alternative relations: 'boat in front of a bird' \\

\caption[image caption correction]{Example for image caption correction. Image caption is the result of the NeuralTalk model \cite{karpathy2015deep}}
\label{fig:caption}
\end{figure}

As the answering process accumulates real "knowledge" related to the image, it may be saved and used for extended applications. One of them may be a discourse on the image, where follow up questions may be answered. Additional application may be correction of image caption \cite{bernardi2016automatic}, where caption may be transformed into a question and the answer may verify it or correct it (as described in Section \ref{sec:alternatives}). An example for image caption correction is given in Figure \ref{fig:caption}.

\section{Results Analysis}
\label{sec:results}

Our system is currently limited by the visual elements it is able to recognize. It is not trained or optimized for any visual question answering dataset. Since our goals include question ``understanding'' and modularity, we first focus in basic capabilities that will be developed with time to be more comprehensive. We've checked our system for various aspects and specific examples and provide an analysis. We've examined graph representation for a random set of questions to see current status as well as potential. The performance of our full system was checked on a wide set of examples. We analyze sources of failures and present examples for correct and incorrect answers.

\subsection{Question Representation}
\label{sec:representation}

First we check the representation capabilities of our system. To do that we've sampled randomly 100 questions from the VQA dataset \cite{VQA} and checked their graph representation. Results are given in Table \ref{table:repRes}.

\begin{table}[h!]
\begin{minipage}{\textwidth}
\begin{centering}
    \begin{tabular}{ | l | l | c | c |} 
    \hline
    \multicolumn{2}{|l|}{}& {\textbf{Current}} & \textbf{Potential}\\ \hline\hline
    \multicolumn{2}{|l|}{\textbf{Fit}}                     & 72 & 100 \\ \hline
    \multirow{2}{*}{\textbf{No fit}} & \textbf{Vocabulary} & 12 & \\ \cline{2-4} 
                                     & \textbf{Other}      & 14 & \\ \hline
    \multicolumn{2}{|l|}{\textbf{Unparsed}}                &  2 & \\ \hline
   \end{tabular}
\caption[Representation results]{Representation results on a random set of 100 questions from the VQA dataset \cite{VQA}. The vocabulary no fit cases are miss representation due to fail in phrases recognition. 'Unparsed' are questions that START couldn't parse. The 'Potential' column represent questions that may be represented by the graph.}
\label{table:repRes}
\end{centering}
\end{minipage}
\end{table}

It is not always clear whether a representation is accurate, as in some cases a representation may fit the language structure but less accurate for the actual meaning. For example a simple representation of the question ``Is this picture in focus?'' may be:

\begin{center}
\begin{tikzpicture}[scale=3, node distance = 2cm, auto]
    \node [cir1] (node1) {$\boldsymbol{c}$: focus};
    \node [cir1, above of = node1] at (0, 0.1) (node2) {$\boldsymbol{c}$: picture};
    \node at (-0.1, 0.32) {\scriptsize{in}};
    \path [line] (node1) -- (node2);
\end{tikzpicture}
\end{center}

However, 'in focus' represents a single element and should be recognized as such. This demonstrates the importance of vocabulary knowledge. In another example, the following questions have a similar structure:
\begin{quote}
Are they all wearing the same \textit{color}?

Are they all wearing the same \textit{pants}?
\end{quote}
However, 'color' and 'pants' belong to two different types of visual elements and hence questions should have different representations.

Sometimes minor phrasing changes have a substantial effect on parsing and representation. The variation in phrasing may also include grammar inaccuracies and typos. This sensitivity reduce the consistency of the representation and adds noise and inaccuracies to the system.


For the two "Unparsed" questions in our representation test, simple corrections lead to successes. The corrections are (original $\rightarrow$ corrected):
\begin{quote}
\small{What season do these toy's represent? $\rightarrow$ What season do these toys represent?

Where are these items at? $\rightarrow$ Where are these items?}
\end{quote}

\begin{figure} [h!]
\begin{centering}
\begin{tabular}{cc}
\begin{tikzpicture}[scale=3, node distance = 2cm, auto]
    \node [cir1] (node1) {$\boldsymbol{c}$: room};
    \node [cir1, above of = node1] at (0, 0.15)  (node2) {$\boldsymbol{c}$: picture
    \scriptsize{\quad\quad $\boldsymbol{f}$: type }};
    \node at (-0.36, 0.33) {\scriptsize{'related\_to'}};
    \path [line] (node1) -- (node2);
\end{tikzpicture}
 &
\begin{tikzpicture}[scale=3, node distance = 2cm, auto]
    \node [cir1] (node1) {$\boldsymbol{c}$: room
    \scriptsize{\quad\quad $\boldsymbol{f}$: type }};
        \node [cir1, above of = node1] at (0, 0.15)  (node2) {$\boldsymbol{c}$: picture };
    \node at (-0.36, 0.34) {\scriptsize{'related\_to'}};
    \path [line] (node1) -- (node2);
\end{tikzpicture}
  \\
(a) What room is \textit{this} a picture of? & (b) What room is \textit{it} a picture of? \\[6pt]
\end{tabular}
\caption[phrasing change example]{Minor phrasing changes may effect START parser results and hence the graph representation. In this example replacement of the word \textit{'this'} with \textit{'it'} (keeping the question's meaning) changes representation to be a correct one.}
\label{fig:phrase_ex}
\end{centering}
\end{figure}
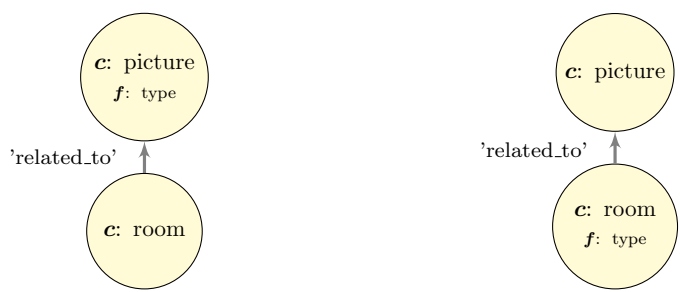

There are other cases where a minor phrasing change corrects the representation, as can be seen in Figure \ref{fig:phrase_ex}.

Additional parsing limitation is no indication coordinating conjunctions ('or', 'and') between phrases. Hence both are treated as 'and'.

As mentioned before, since the questions are free form, they may involve slang, typos or wrong grammar. The question meaning may even be not clear. For example the question 'How is the table design?' may be the correct intended question. However it may be that the intended question is ``How is the table \textit{designed}?''.

All the questions sampled in this analysis can be potentially represented using the suggested graph representation. This demonstrates that in general our scheme has a very high representation capabilities. However some require identification of complicated properties and related terms e.g. ``Is the refrigerator \textit{capacity greater than 22 cubic feet}?'' (similar comparisons of property's quantity already exist for age). The issue of adding description levels rises for complicated properties that may have a natural representation using properties of properties, e.g.
\begin{quote}
Is this the \textit{normal use} for the object holding the flowers?

\textit{How} is the table \textit{designed}?

\textit{Where} do these animals \textit{originate}?
\end{quote}

In some cases it may be reasonable to alter the exact meaning into a more reasonable one to handle, e.g
\begin{quote}
Does this truck have all of its original parts? $\rightarrow$ Are all the parts of this truck original?
\end{quote}

In other checks performed, there were (very few) cases where relations between multiple objects of different types were required (e.g. 'Does this image contain more mountain, sky or grass?'). A support for such cases may be added in the future.

\subsection{Question Answering}

Our current implementation is obviously limited by the number of recognizable visual elements, queried both explicitly and implicitly. It does not include any training or adaptation to any Visual Question Answering dataset. Also, some implementations maybe incomplete or arbitrary, e.g. 'location', which implementation is relative to image. Answers are, however mostly self aware. When running on the VQA \cite{VQA} dataset most answers indicate the unfamiliar visual element which prevents answering (e.g. "Unknown class: linoleum"). %

\begin{figure} [h!]
\begin{centering}
\scalebox{0.95}{
\setlength\tabcolsep{1.5pt} 
\renewcommand{\arraystretch}{0.8}
\begin{tabular}{p{5cm}p{5cm}p{5cm}}
\subfigure{
  \includegraphics[totalheight=0.165\textheight]{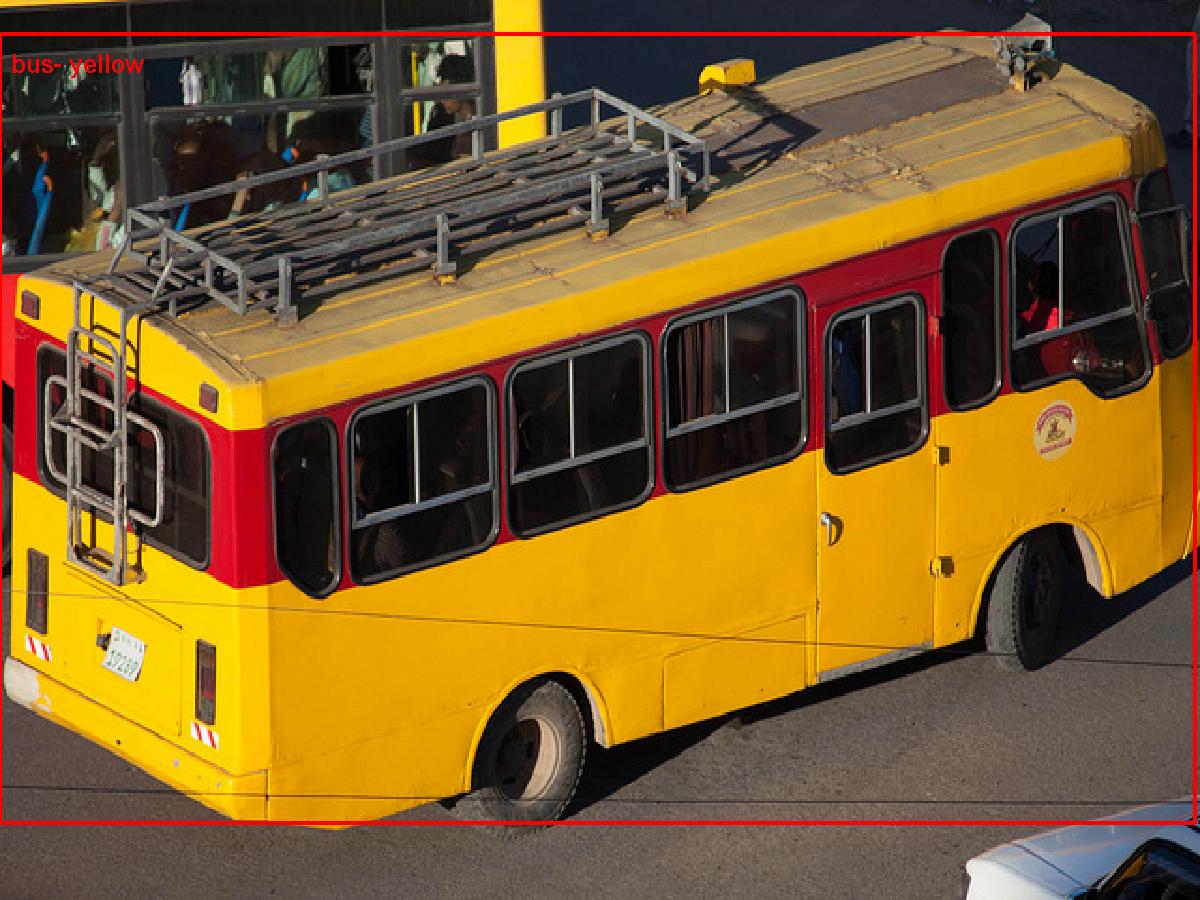}
} &
\subfigure{
  \includegraphics[totalheight=0.165\textheight]{COCO_val2014_000000545309_qa_type_transportation}
} &
\subfigure{
  \includegraphics[totalheight=0.165\textheight]{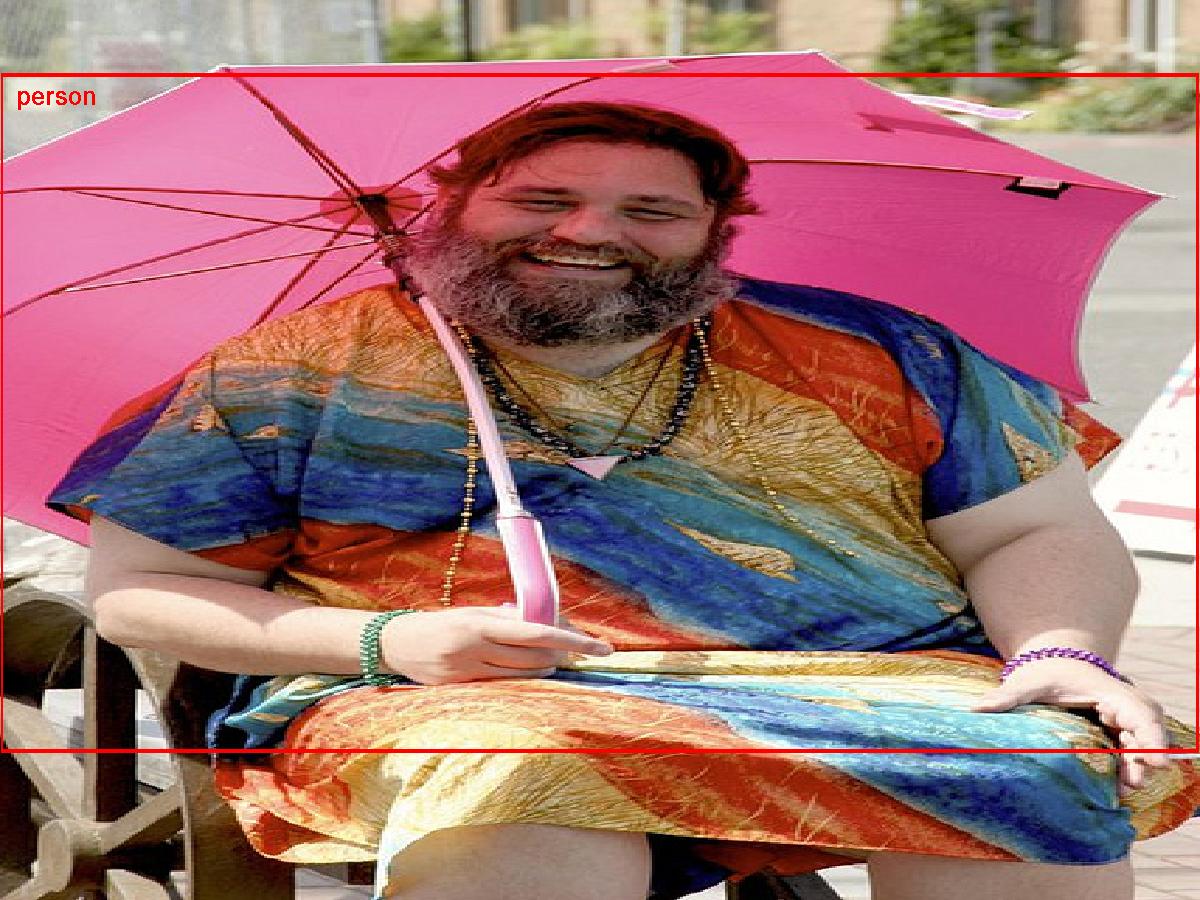}
} \\

\scriptsize{\textbf{Q:} What color is this ride?} & \scriptsize{\textbf{Q:}} \tinyPut{What type of transportation is this?} & \scriptsize{\textbf{Q:} Is this a lady?} \\
\scriptsize{\textbf{A:} yellow} &  \scriptsize{\textbf{A:} train} &  \scriptsize{\textbf{A:} no} \\

\subfigure{
  \includegraphics[totalheight=0.165\textheight]{COCO_val2014_000000364745-man_looking_at_children}
} &
\subfigure{
  \includegraphics[totalheight=0.165\textheight]{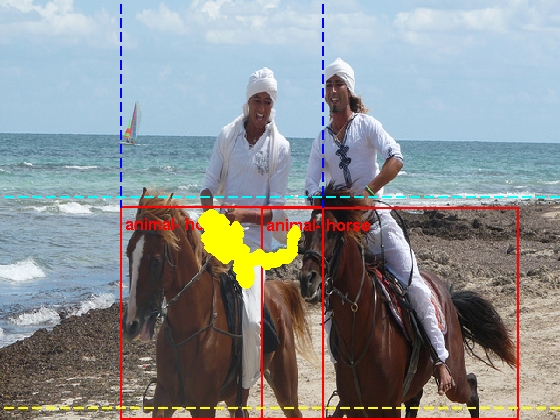}
} &
\subfigure{
  \includegraphics[totalheight=0.165\textheight]{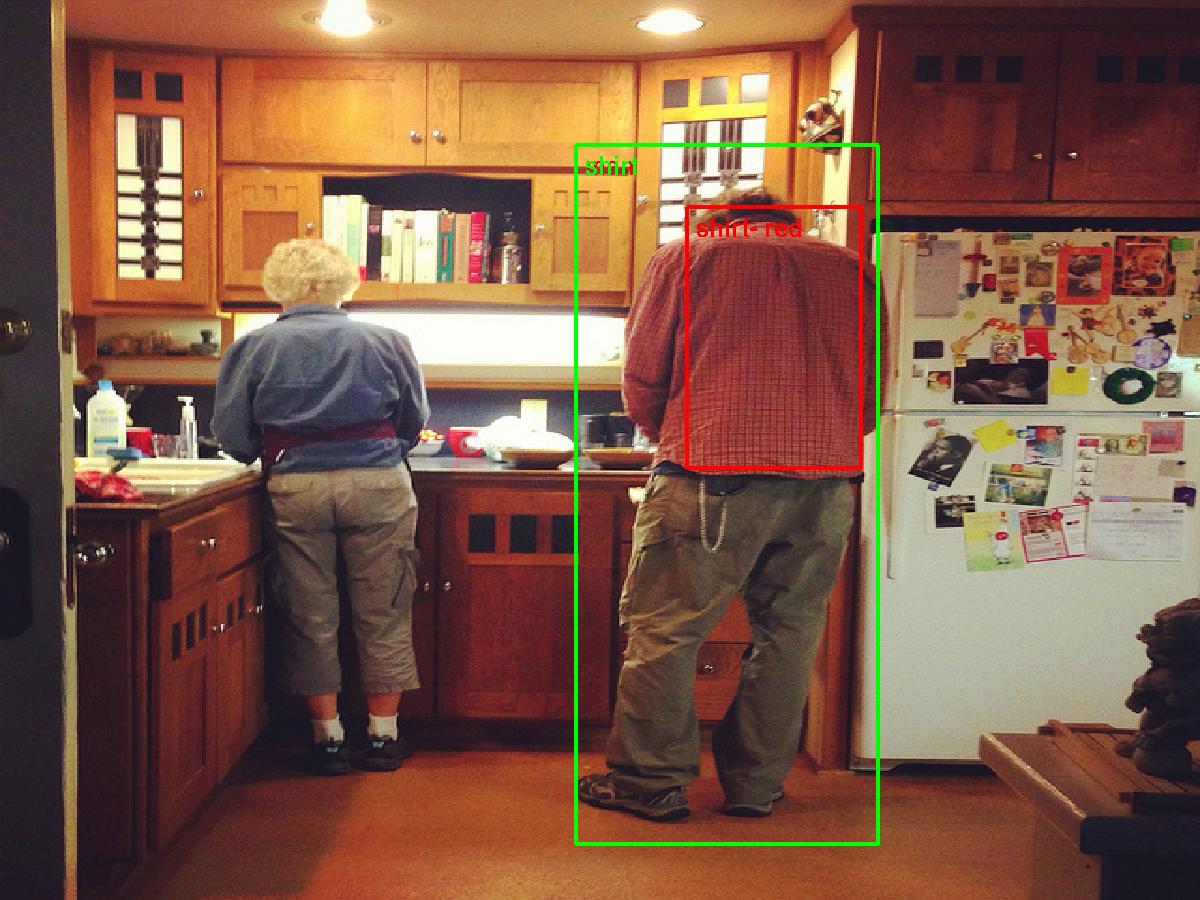}
} \\

\scriptsize{\textbf{Q:} Is the man looking at the children?} & \scriptsize{\textbf{Q:}} \tinyPut{What type of animal are both men riding? } & \scriptsize{\textbf{Q:} What color is the man's shirt?} \\
\scriptsize{\textbf{A:} no} & \scriptsize{\textbf{A:} horse} & \scriptsize{\textbf{A:} red} \\

\subfigure{
  \includegraphics[totalheight=0.165\textheight]{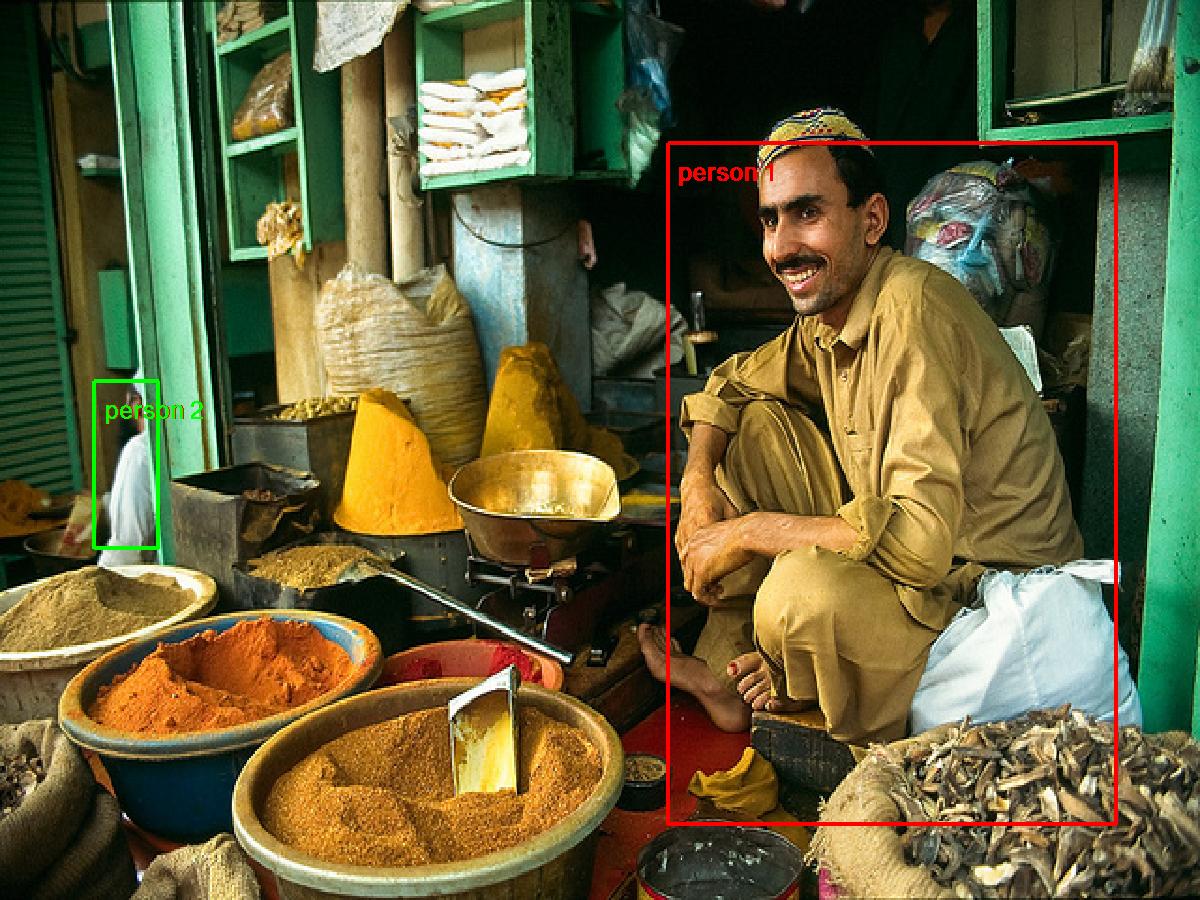}
} &
\subfigure{
  \includegraphics[totalheight=0.165\textheight]{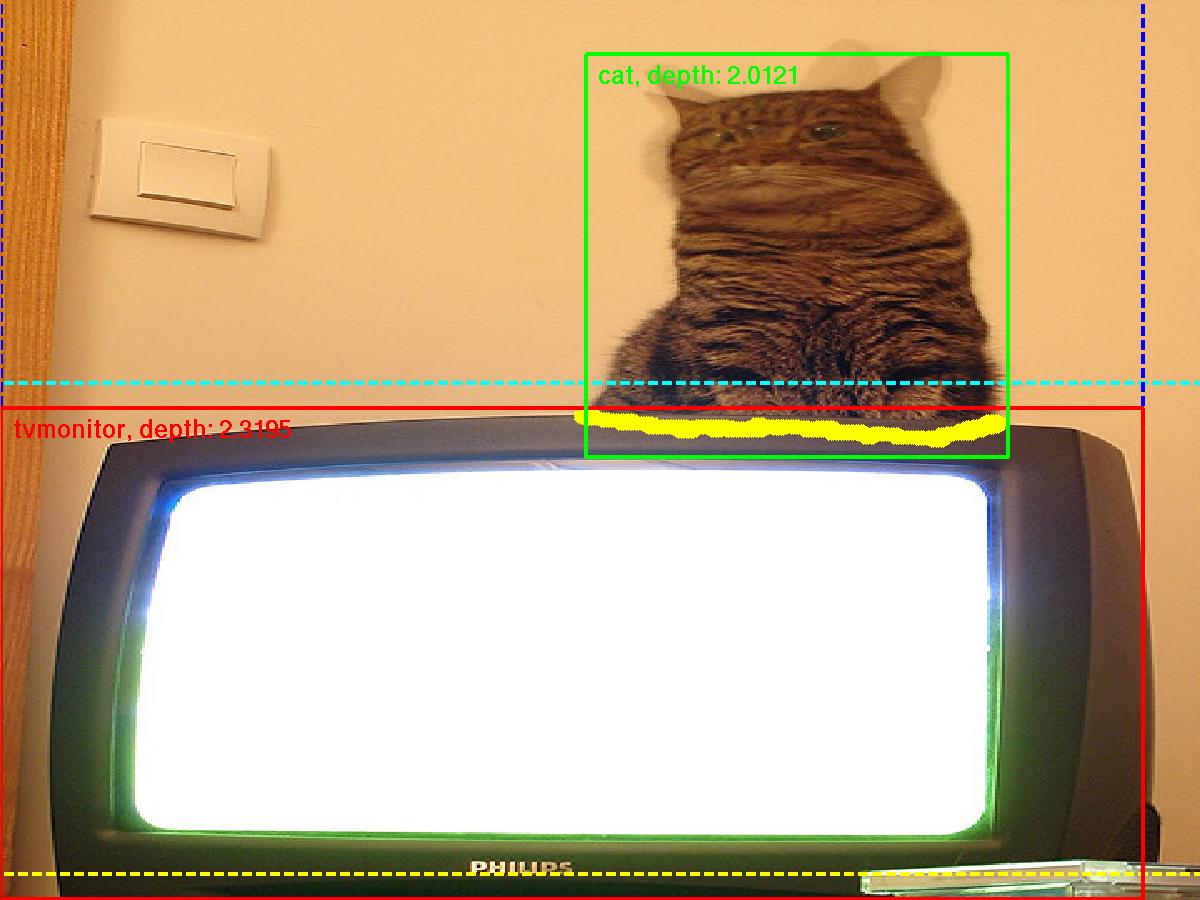}
} &
\subfigure{
  \includegraphics[totalheight=0.165\textheight]{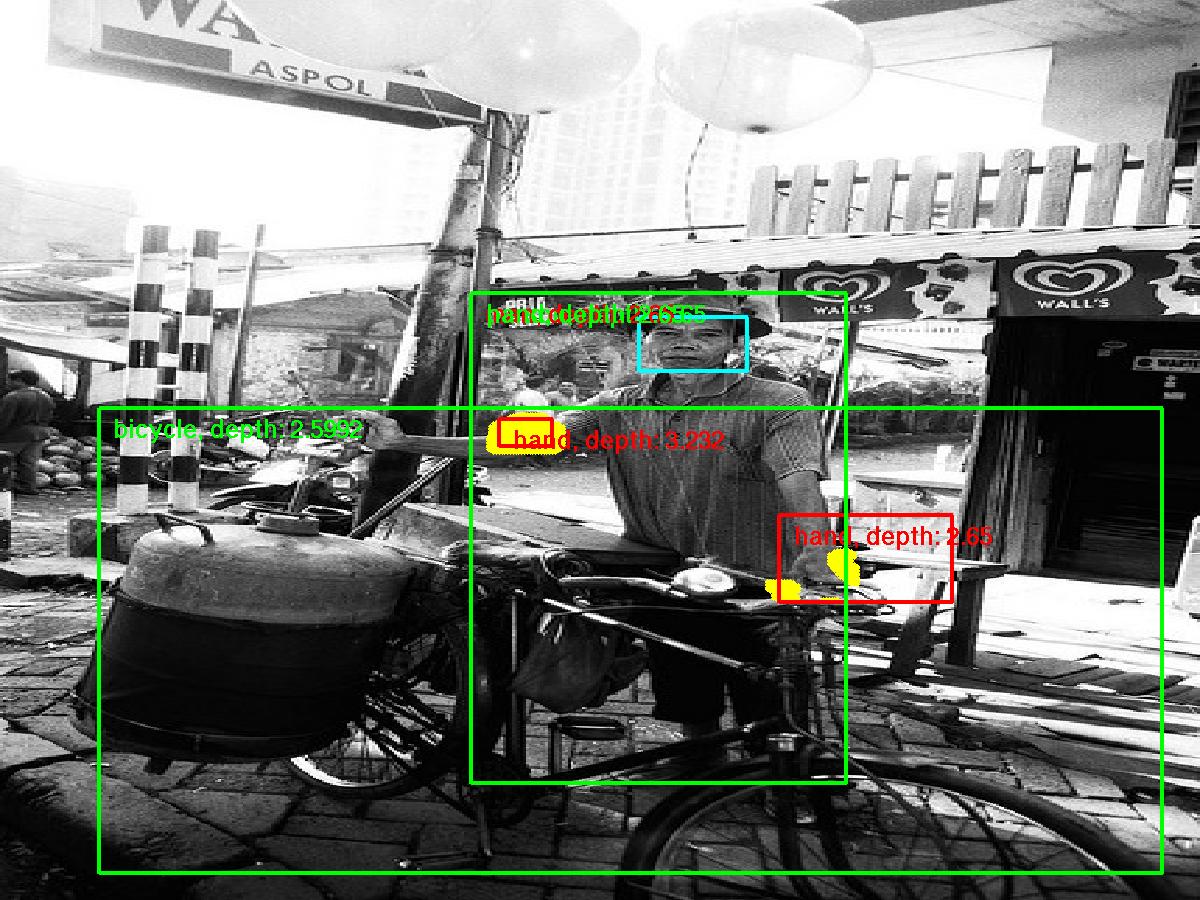}
} \\

\scriptsize{\textbf{Q:} How many people are there?} & \scriptsize{\textbf{Q:} What is on top of the television?} & \scriptsize{\textbf{Q:} What is the man holding?} \\
 \scriptsize{\textbf{A:} 2} & \scriptsize{\textbf{A:} cat} & \scriptsize{\textbf{A:}  bicycle} \\

\subfigure{
  \includegraphics[totalheight=0.165\textheight]{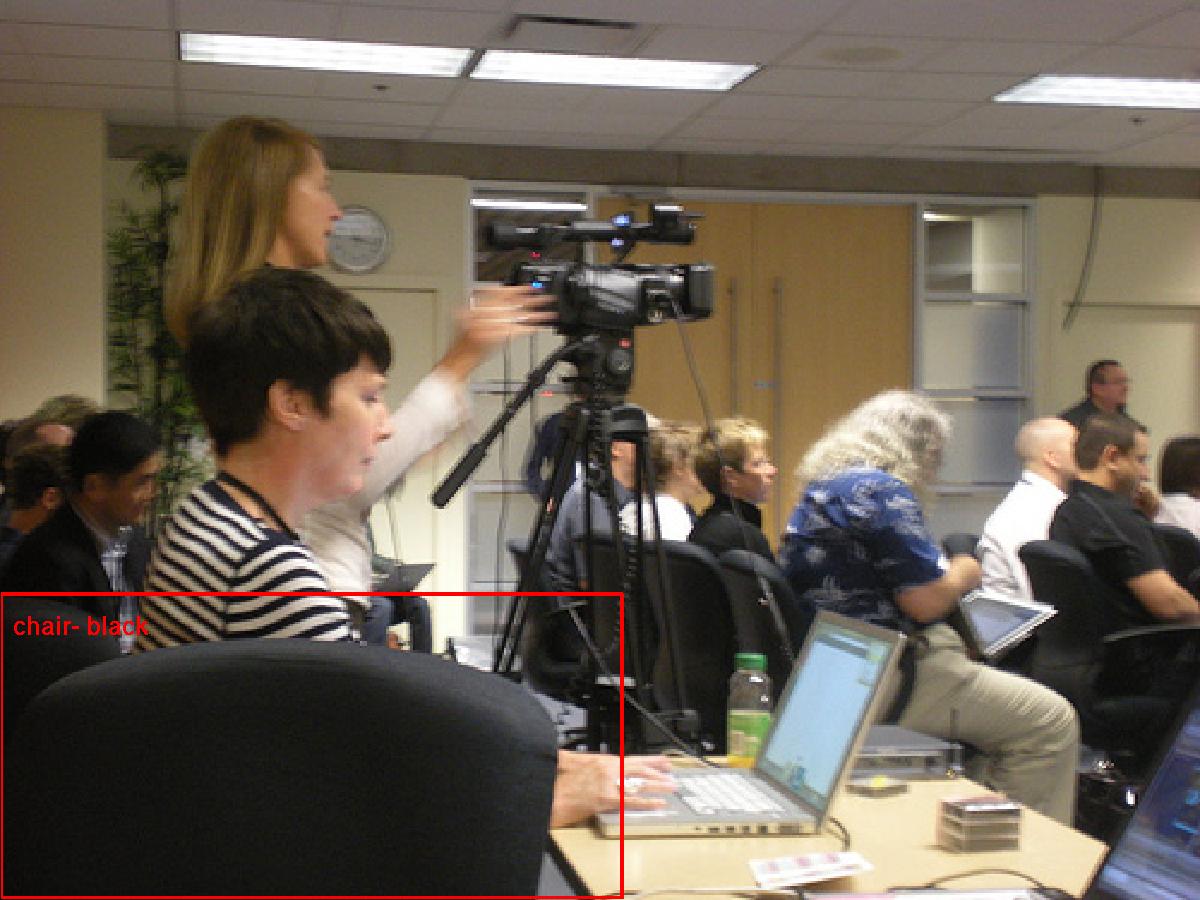}
} &
\subfigure{
  \includegraphics[totalheight=0.165\textheight]{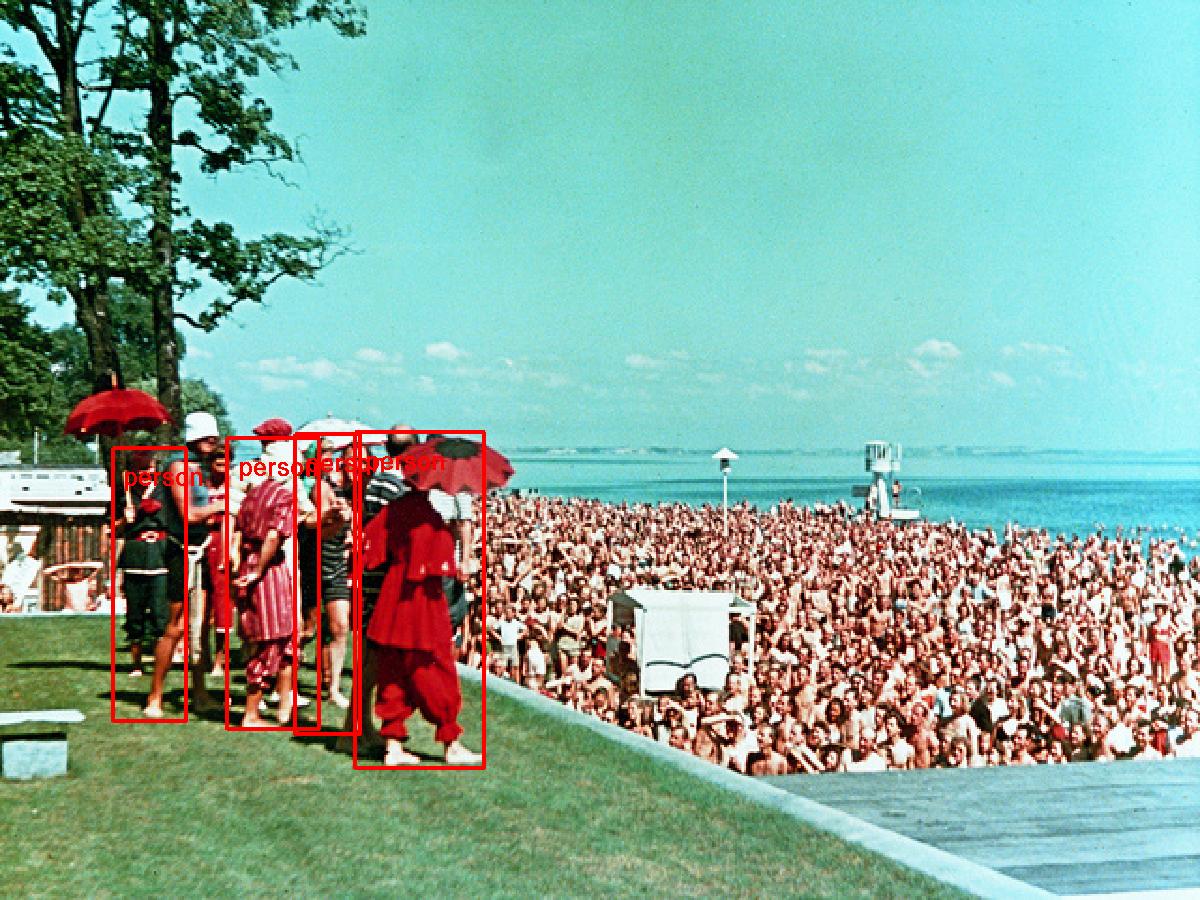}
}&
\subfigure{
  \includegraphics[totalheight=0.165\textheight]{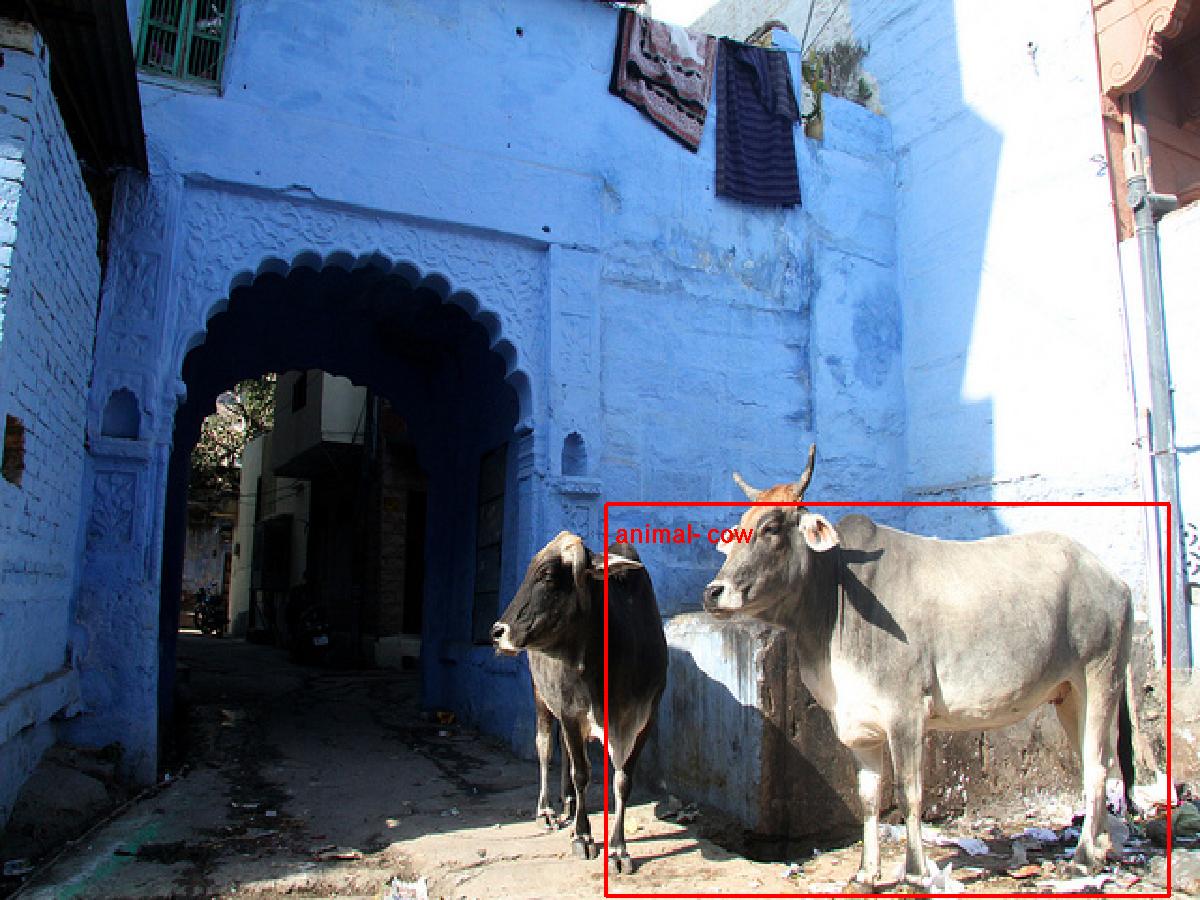}
} \\

\scriptsize{\textbf{Q:} What color is the chair?} & \scriptsize{\textbf{Q:}} \tinyPut{Are there lots of people in this photo?} & \scriptsize{\textbf{Q:} What kind of animals are shown?} \\
\scriptsize{\textbf{A:} black} & \scriptsize{\textbf{A:} yes} & \scriptsize{\textbf{A:} cow}
\end{tabular}
}
\caption[Examples for correct answers]{Examples for correct answers from the VQA dataset\cite{VQA} (short answers). [Object detection is based on faster R-CNN + DeepLab].}
\label{fig:goodRes}
\end{centering}
\end{figure}

Examples with proper answers are shown in Figure \ref{fig:goodRes}. It includes the use of ConceptNet \cite{Speer2013} in some cases to obtain prior knowledge regarding related classes (e.g. subclasses) and other commonsense knowledge. For example \textit{\textbf{'ride'}} subclasses: \textit{\{'bicycle', 'bus', 'boat', 'motorbike', 'train'\}}, \textit{\textbf{'transportation'}} subclasses: \textit{\{'train', 'boat', 'bicycle'\}}, \textit{\textbf{'animal'}} subclasses: \textit{\{'dog', 'horse', 'cat', 'bird', 'sheep', 'cow'\}}.

\begin{figure} [h!]
\begin{centering}
\scalebox{0.9}{
\setlength\tabcolsep{1.5pt} 
\renewcommand{\arraystretch}{0.8}
\begin{tabular}{p{5cm}p{5cm}p{5.5cm}}
\subfigure{
  \includegraphics[totalheight=0.165\textheight]{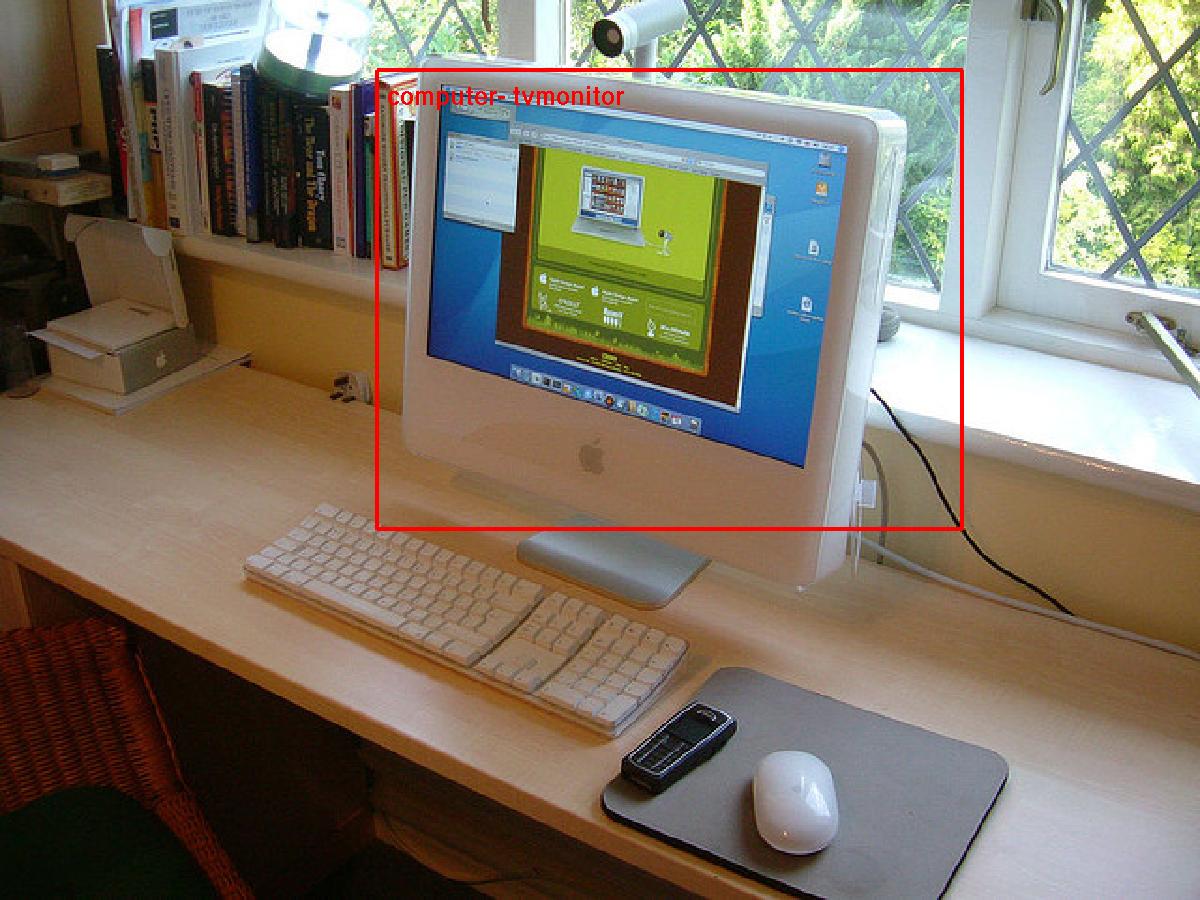}
} &
\subfigure{
  \includegraphics[totalheight=0.165\textheight]{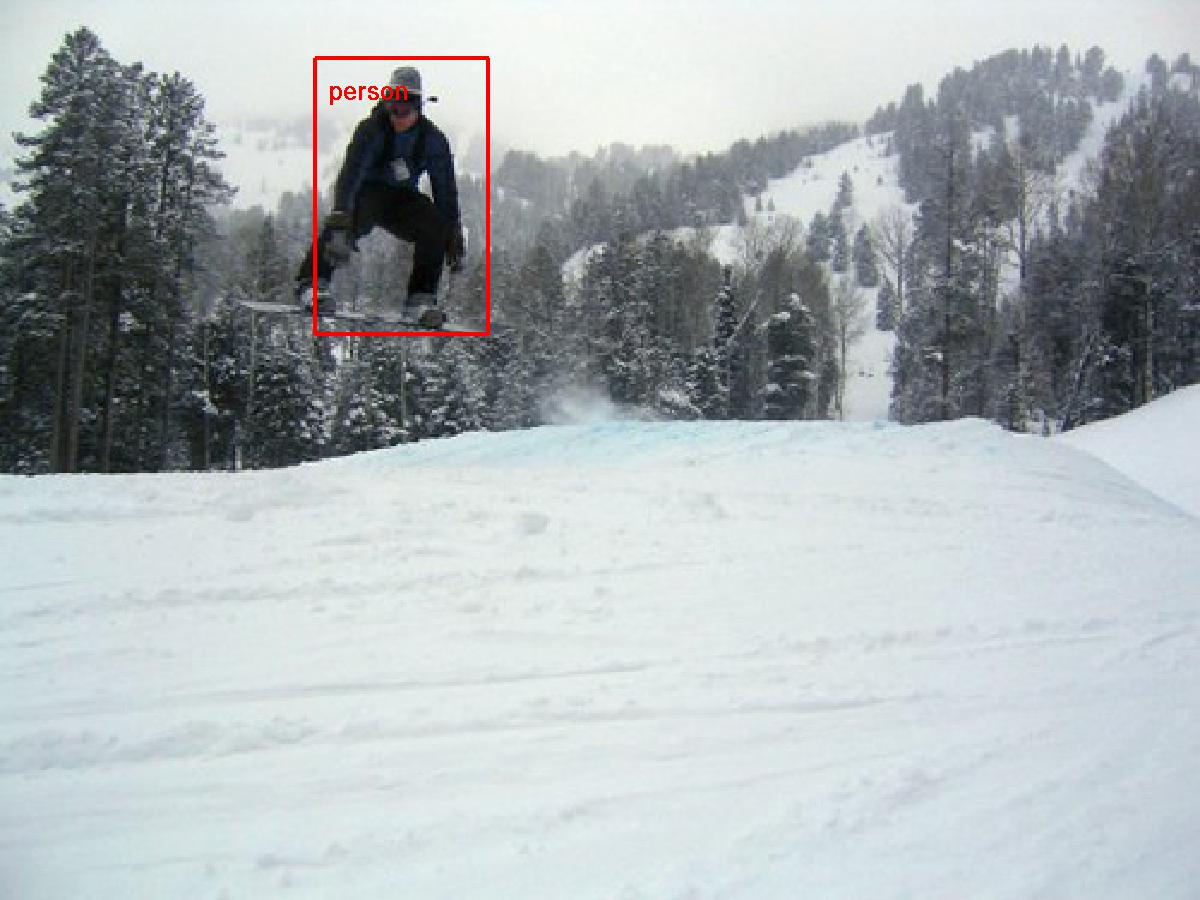}
} &
\subfigure{
  \includegraphics[totalheight=0.165\textheight]{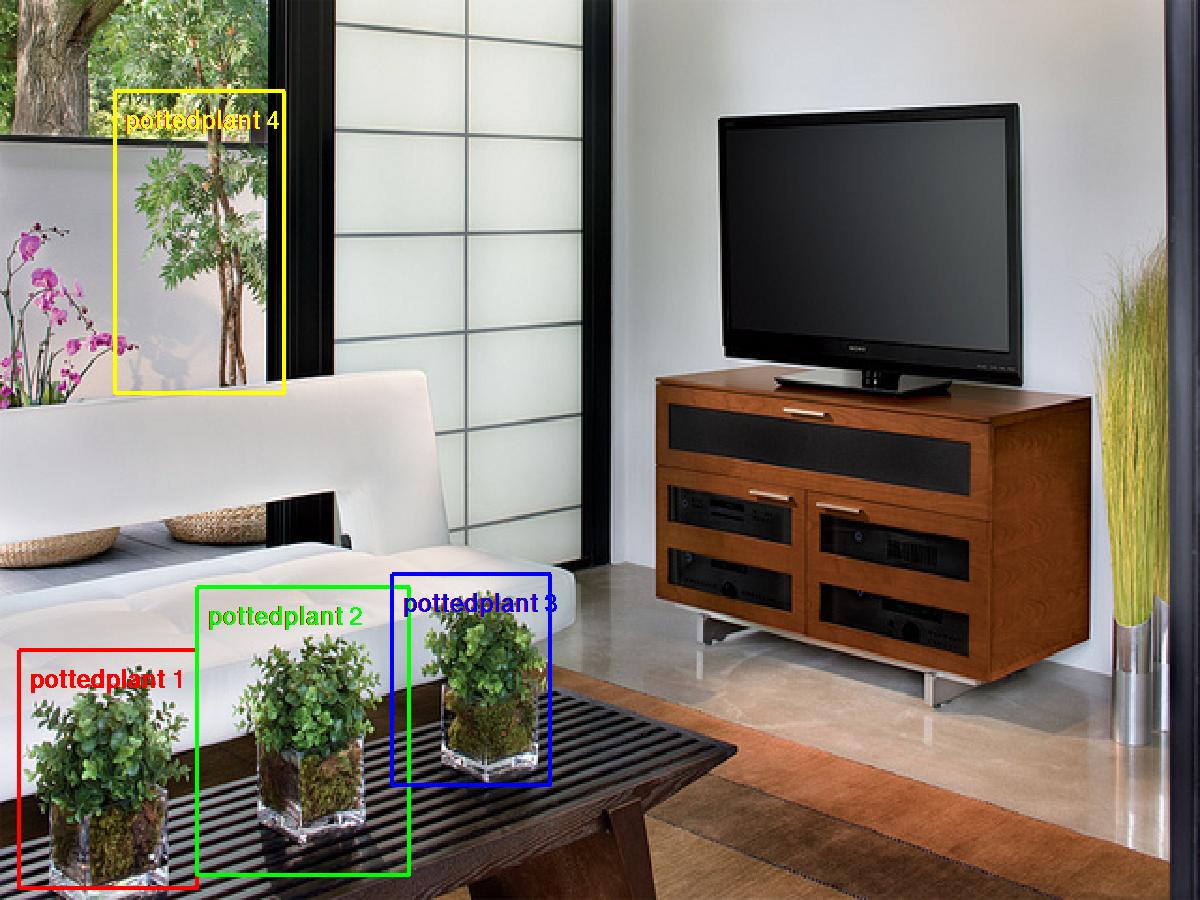}
} \\

\scriptsize{\textbf{Q:} What kind of computer is this?} & \scriptsize{\textbf{Q:} What is he on?} & \scriptsize{\textbf{Q:}} \tinyPut{How many plants can be seen in this picture?} \\
\scriptsize{\textbf{A:} tvmonitor} &  \scriptsize{\textbf{A:} nothing} &  \scriptsize{\textbf{A:} 4} \\

\subfigure{
  \includegraphics[totalheight=0.165\textheight]{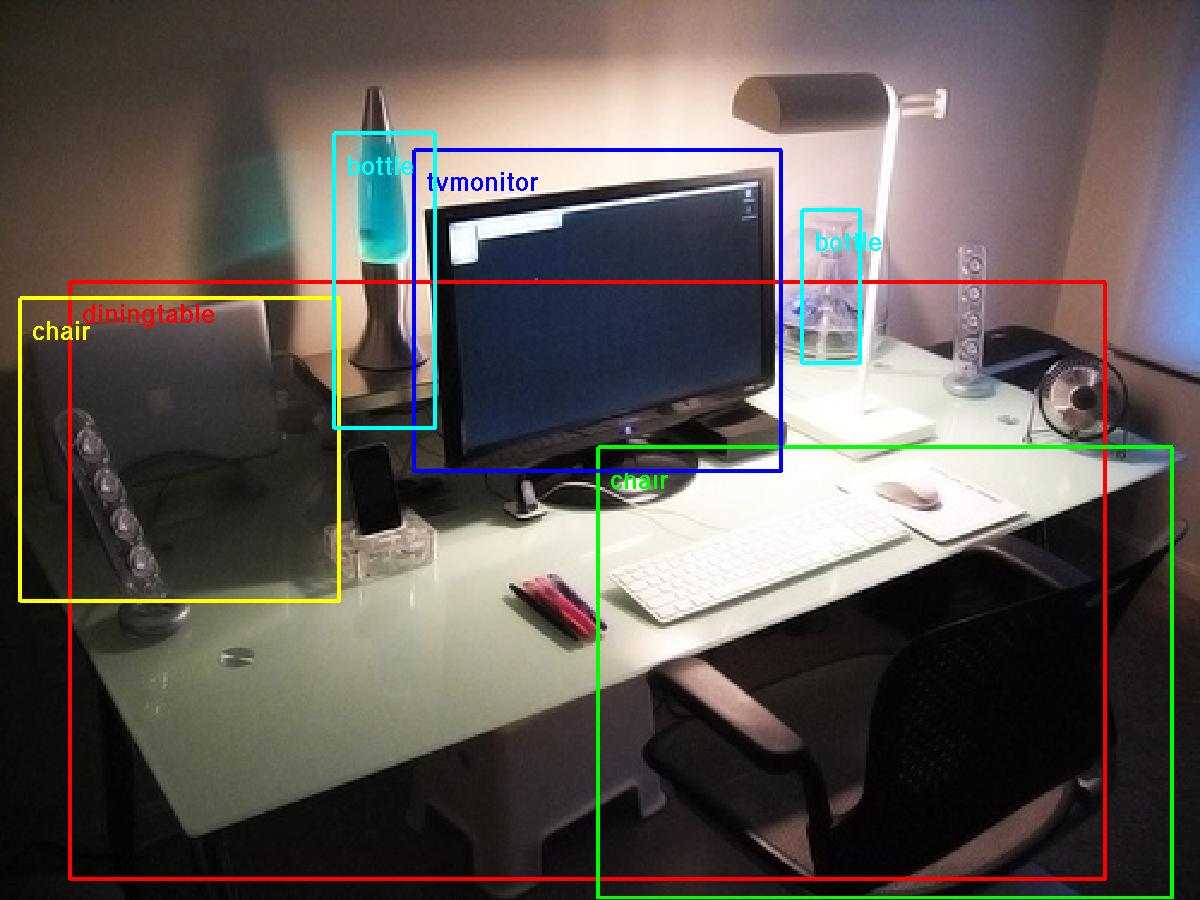}
} &
\subfigure{
  \includegraphics[totalheight=0.165\textheight]{COCO_val2014_000000220182_qa_how_many_on_table_class_dets}
} &
\subfigure{
  \includegraphics[totalheight=0.165\textheight]{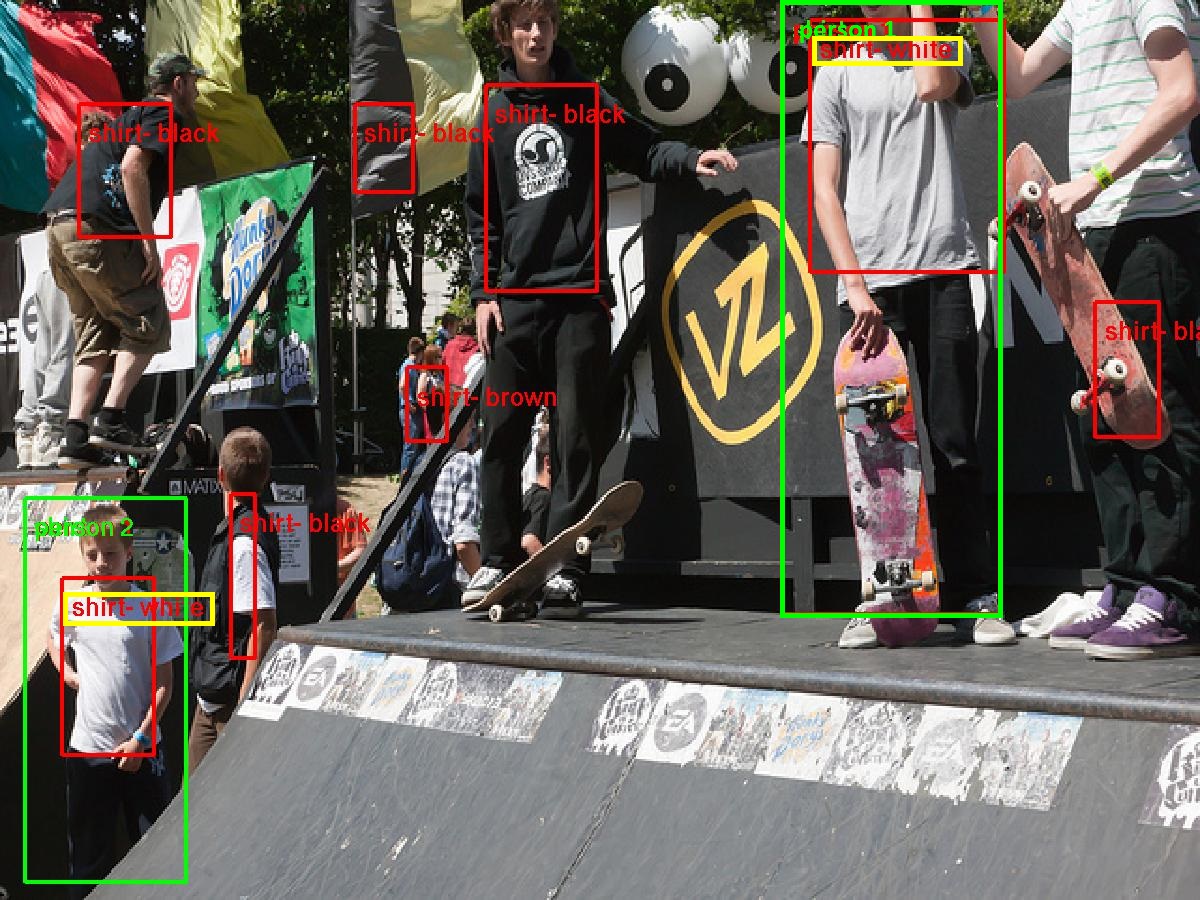}
} \\

\scriptsize{\textbf{Q:} What is on the computer monitor?} & \scriptsize{\textbf{Q:} How many flowers are on the table?} & \scriptsize{\textbf{Q:}} \tinyPut{How many people are wearing white shirts?} \\
\scriptsize{\textbf{A:} nothing} & \scriptsize{\textbf{A:} There is no diningtable} & \scriptsize{\textbf{A:} 2} \\

\subfigure{
  \includegraphics[totalheight=0.165\textheight]{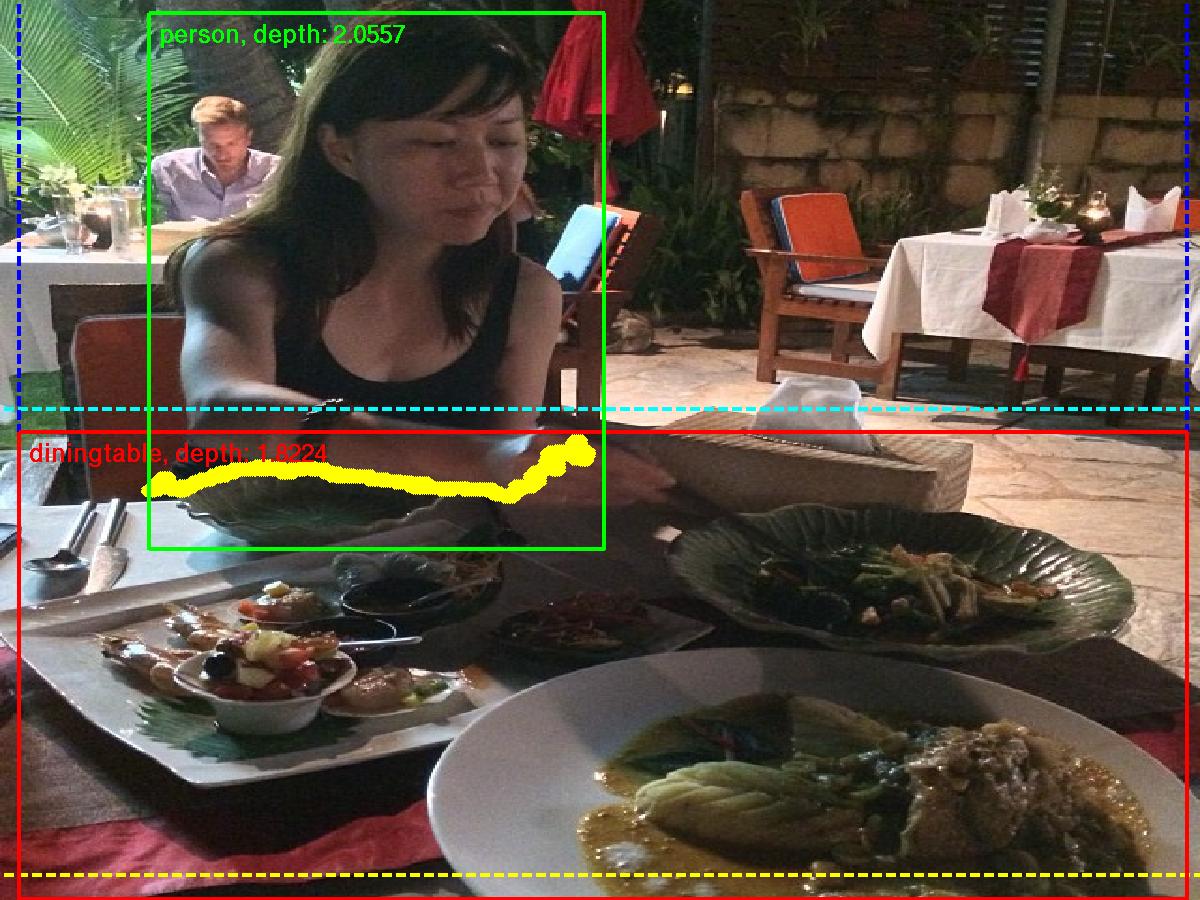}
} &
\subfigure{
  \includegraphics[totalheight=0.165\textheight]{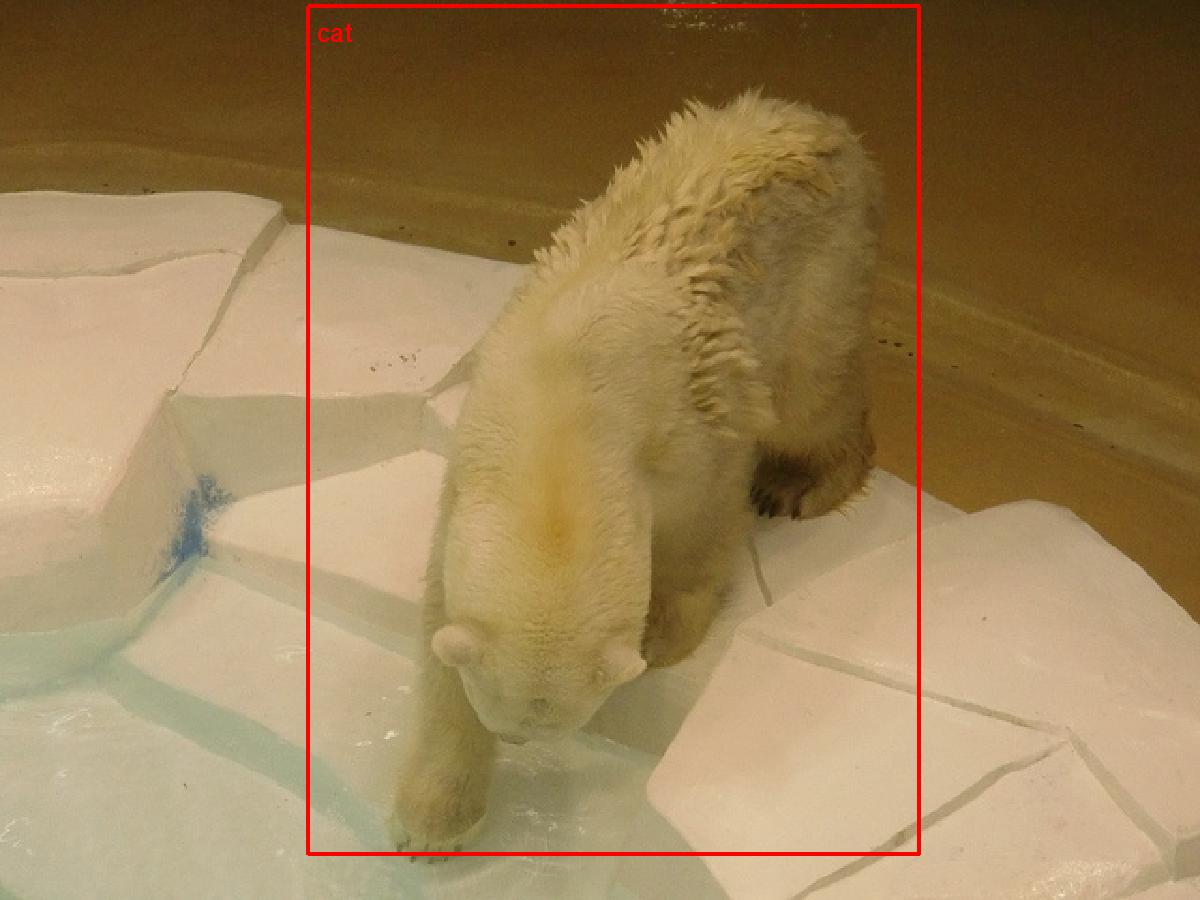}
} &
\subfigure{
  \includegraphics[totalheight=0.165\textheight]{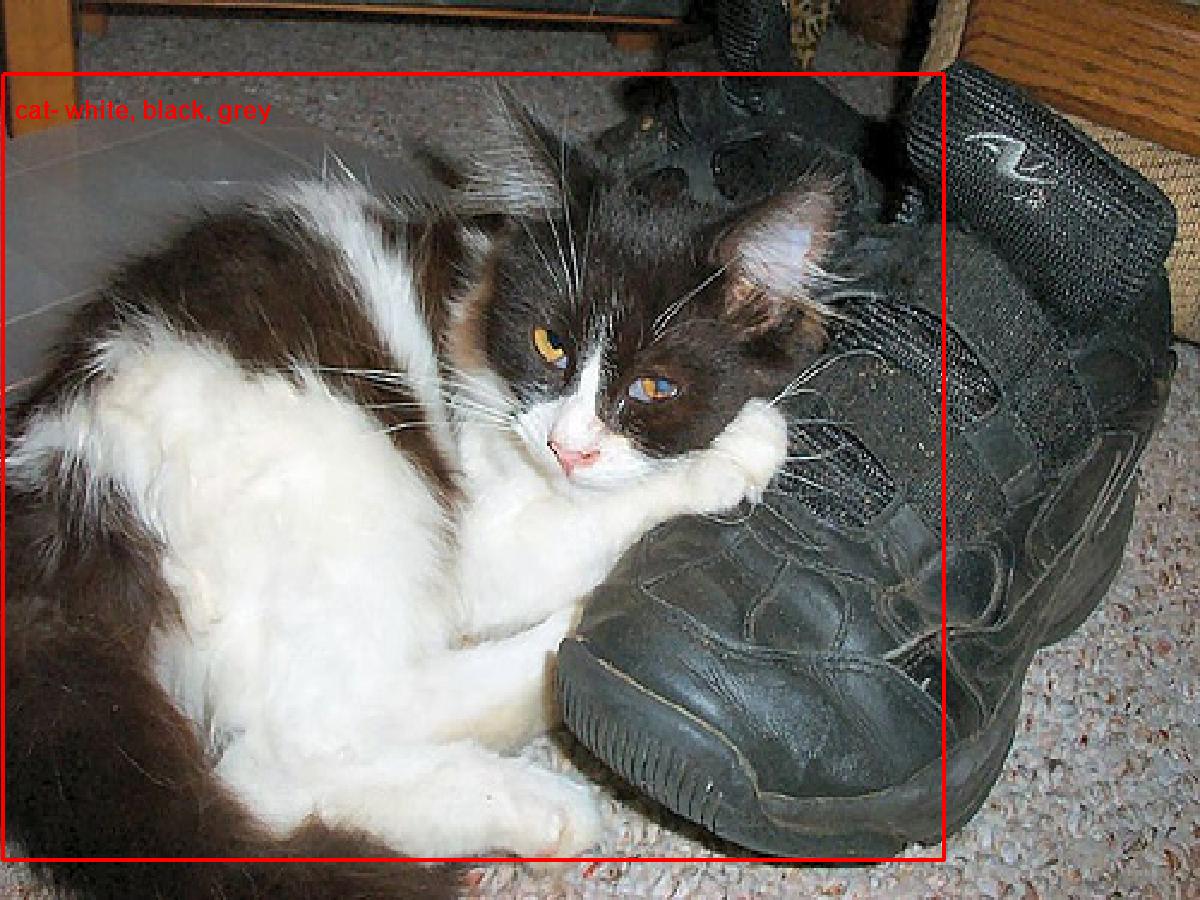}
} \\

\scriptsize{\textbf{Q:} What is on the table?} & \scriptsize{\textbf{Q:} Is this a cat?} & \scriptsize{\textbf{Q:} What colors is this cat?} \\
 \scriptsize{\textbf{A:} person} & \scriptsize{\textbf{A:} yes} & \scriptsize{\textbf{A:} white, black, grey} \\

\subfigure{
  \includegraphics[totalheight=0.165\textheight]{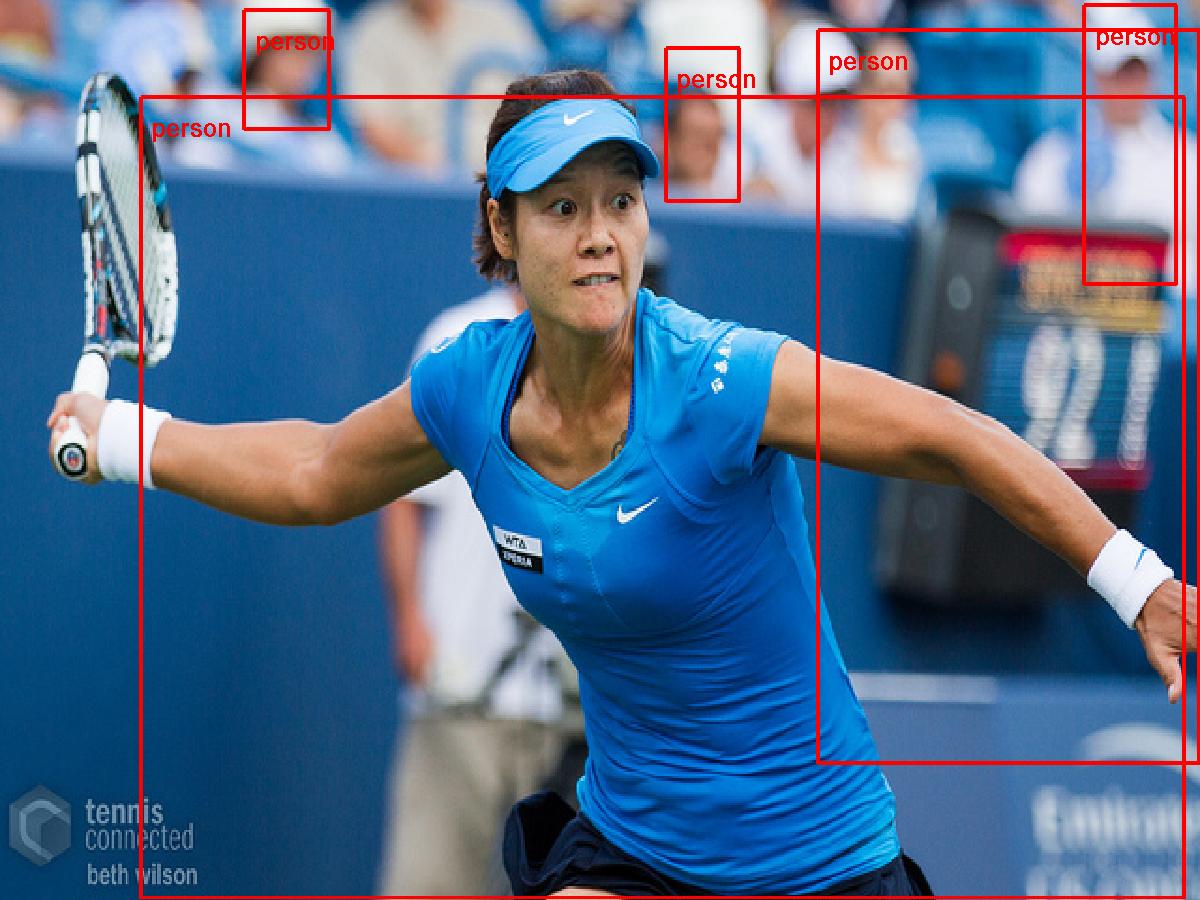}
} &
\subfigure{
  \includegraphics[totalheight=0.165\textheight]{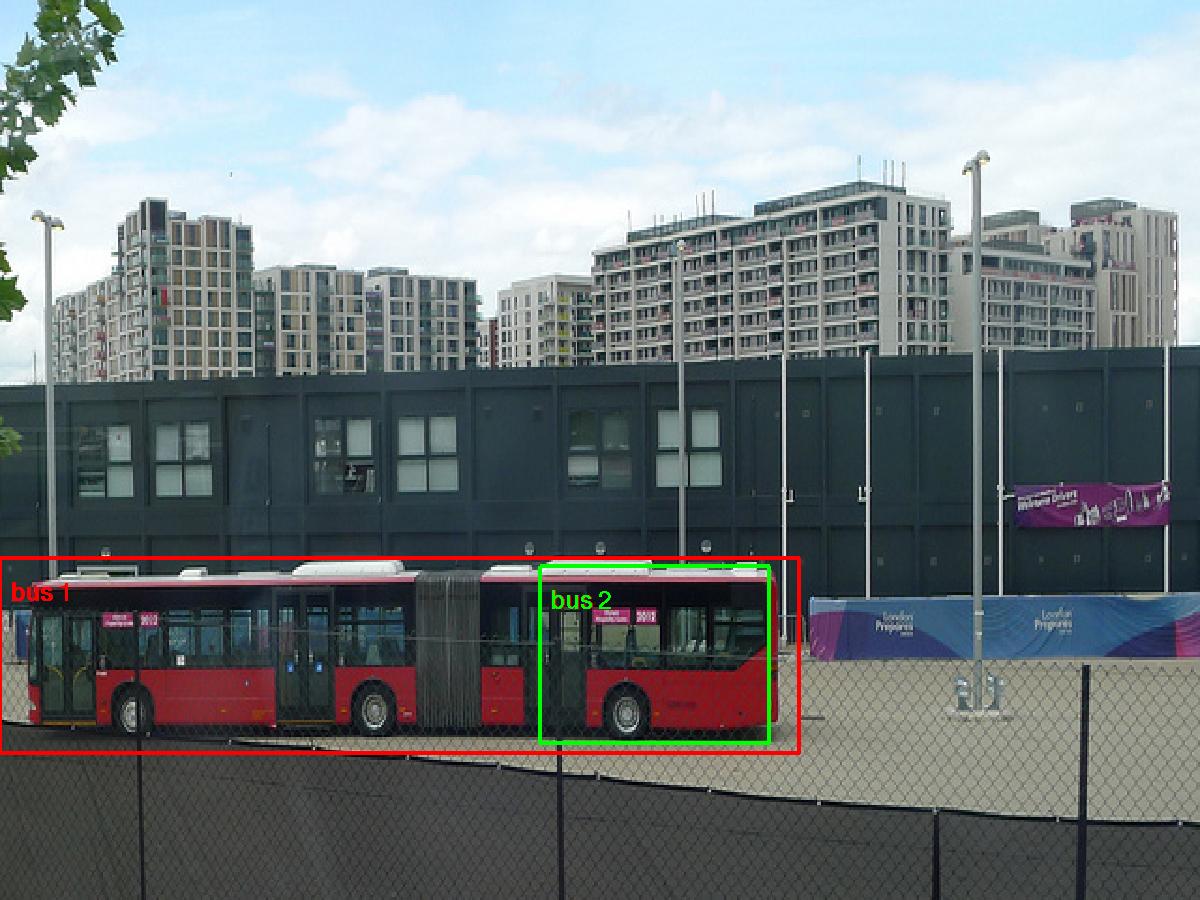}
}&
\subfigure{
  \includegraphics[totalheight=0.165\textheight]{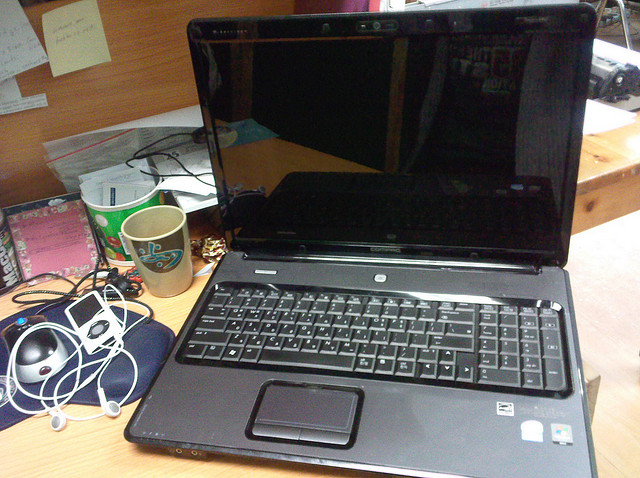}
} \\

\scriptsize{\textbf{Q:}} \tinyPut{What is the woman wearing on her wrists?} & \scriptsize{\textbf{Q:} How many buses are there?} & \scriptsize{\textbf{Q:} Is the screen on?} \\
\scriptsize{\textbf{A:} don't know} & \scriptsize{\textbf{A:} 2} & \scriptsize{\textbf{A:} unknown property 'on'}
\end{tabular}
}
\caption[Examples for incorrect answers]{Examples for incorrect answers from the VQA dataset\cite{VQA} (short answers). [Object detection is based on faster R-CNN + DeepLab].}
\label{fig:badRes}
\end{centering}
\end{figure}

Examples with wrong answers are shown in Figure \ref{fig:badRes}. The reasons for failures include detection failures, unknown visual elements, missing prior knowledge and other assumptions.

Further examination of the results provides some insights regarding additional sources of failure.

One element that adds "noise" to the system is the use of internet based external knowledge database. While providing essential information, retrieved data is also prone to errors and yields detection attempts of wrong objects. This is demonstrated by the results of queries of 'carpet' and the relation IsA which imply that the following may be a carpet: 'Barack Obama', 'book', 'monitor',' a plastic bag', 'a glass of water', etc. Another example for such an error is the retrieved relation \textit{'chair IsA door'}. A partial solution is using the associated weights that indicate the strength of each result. Some results may be misleading as they may refer to different meanings of the queried words. Following are examples for such results:
\begin{quote}
\textit{'\textbf{train} isA control'}

\textit{'\textbf{monitor} isA track'}

\textit{'\textbf{screen}\_door isA door'}
\end{quote}
In some cases the intersection of retrieved classes with recognizable objects is so small, that it may cause a wrong conclusion based on a very superficial check. An example for this is the question "Are these toys?", where the recognizable retrieved classes are 'bicycle', 'skateboard', 'frisbee', 'kite' and 'motorcycle' hence answering 'no' if none of them was detected.

\begin{figure} [h!]
\begin{centering}
\setlength\tabcolsep{1.5pt} 
\renewcommand{\arraystretch}{0.8}
\begin{tabular}{p{5cm}p{5cm}p{5cm}}
\subfigure{
  \includegraphics[totalheight=0.165\textheight]{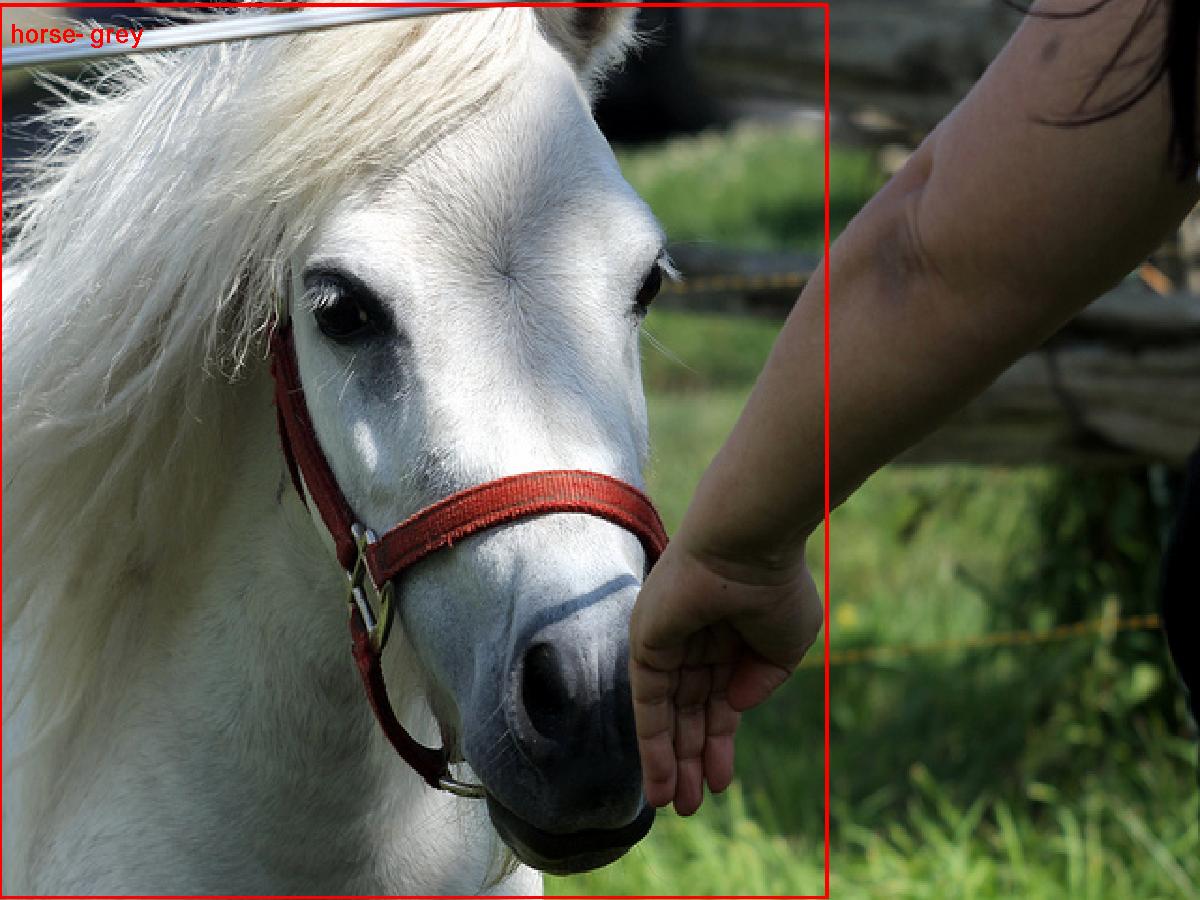}
} &
\subfigure{
  \includegraphics[totalheight=0.165\textheight]{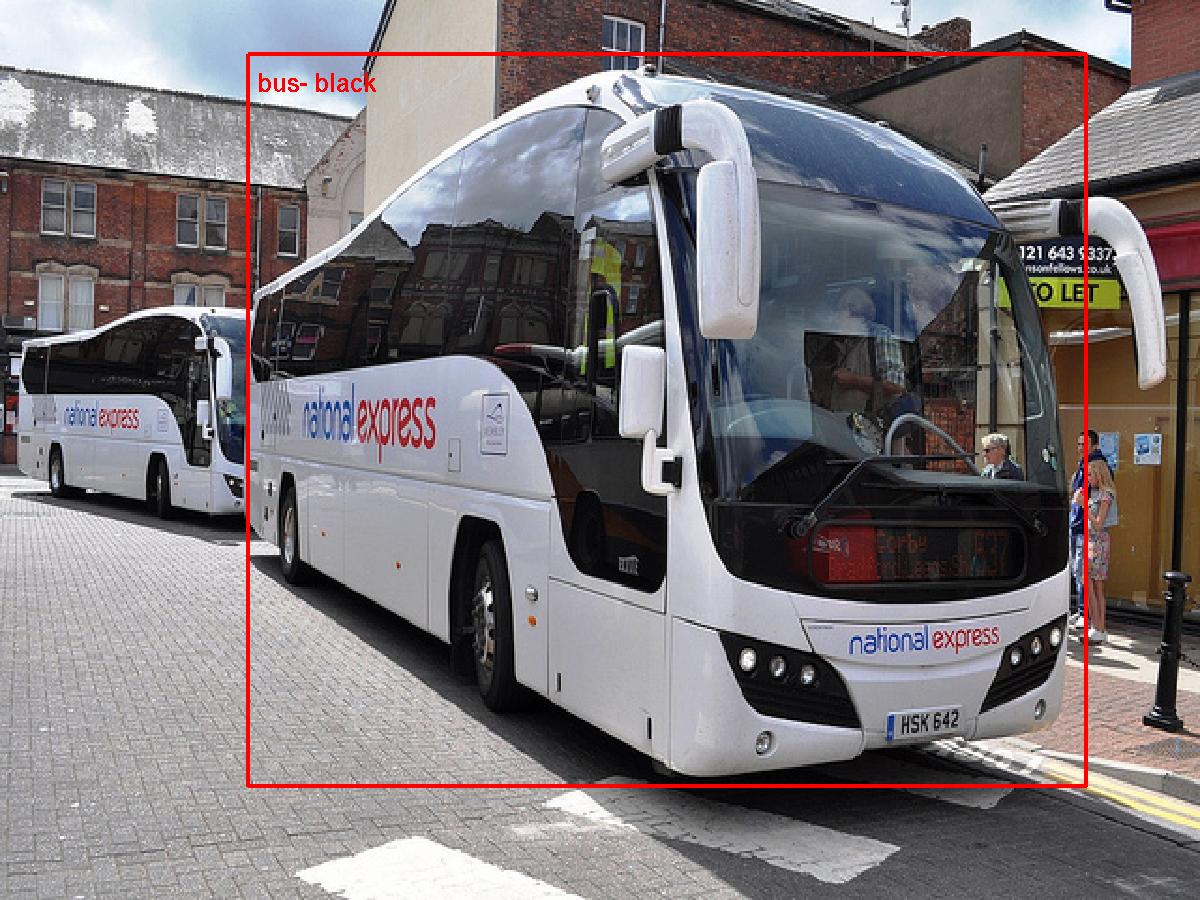}
} &
\subfigure{
  \includegraphics[totalheight=0.165\textheight]{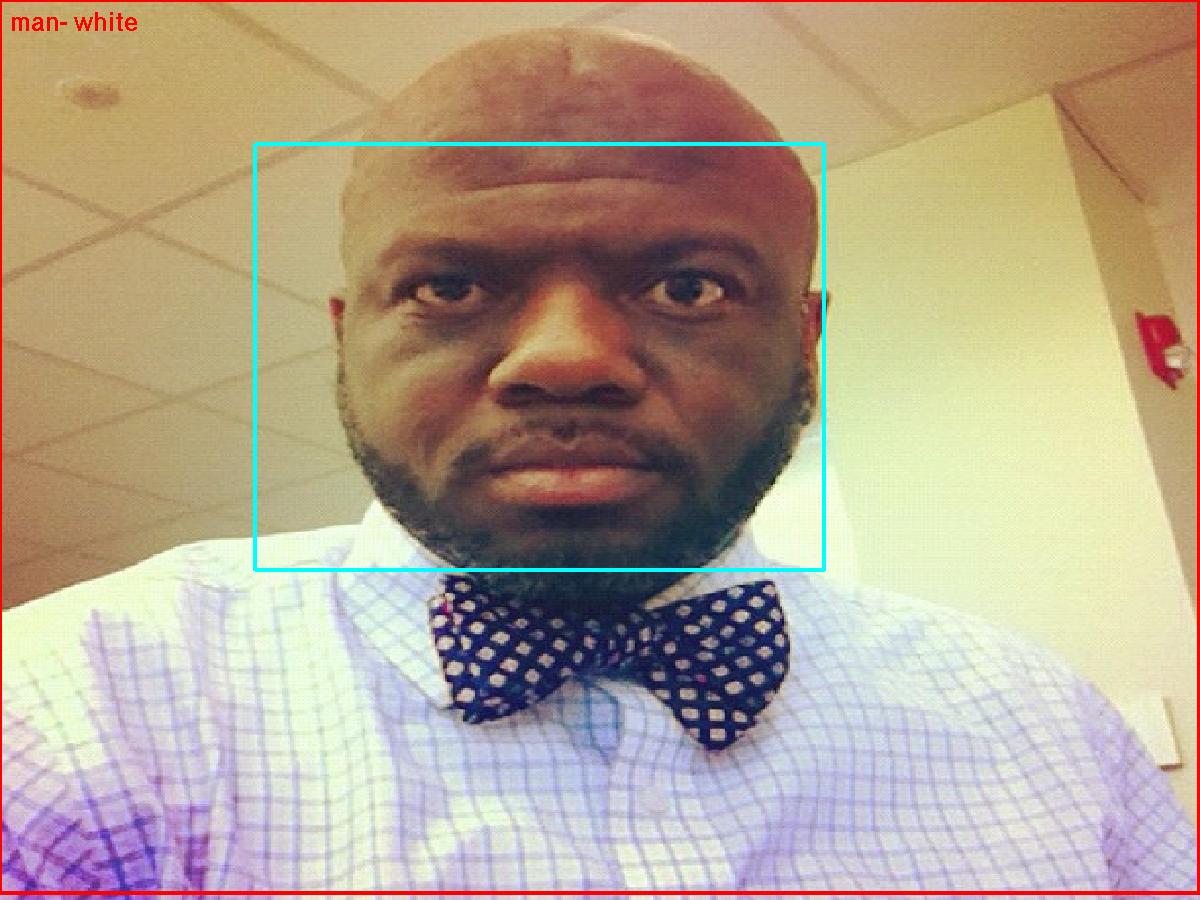}
} \\

\subfigure{
  \includegraphics[totalheight=0.165\textheight]{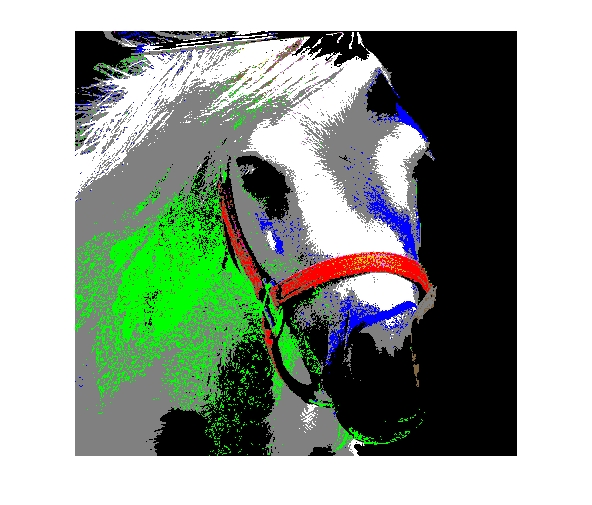}
} &
\subfigure{
  \includegraphics[totalheight=0.165\textheight]{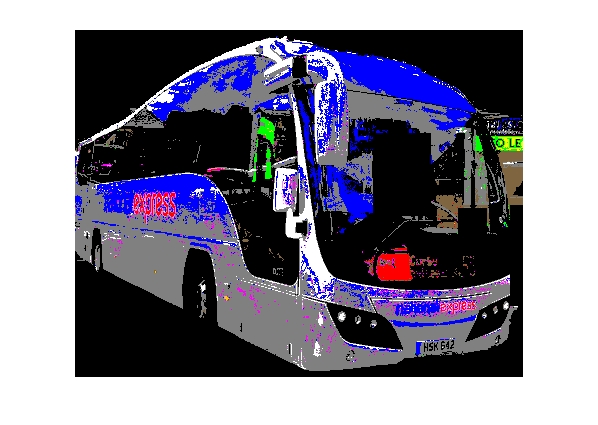}
} &
\subfigure{
  \includegraphics[totalheight=0.165\textheight]{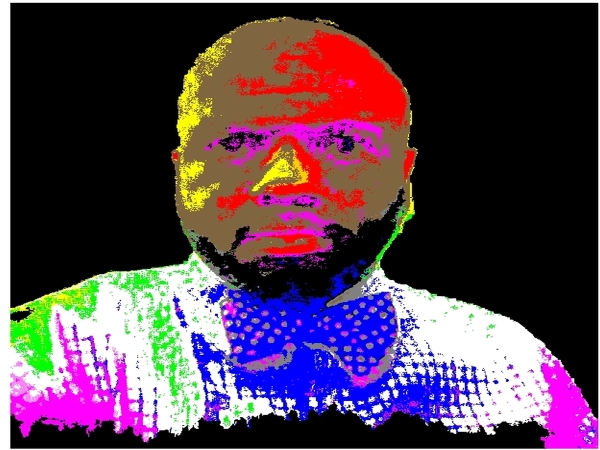}
} \\

\scriptsize{\textbf{Q:} What color is the horse?} & \scriptsize{\textbf{Q:} What color is the bus?} & \scriptsize{\textbf{Q:} Is the man white?} \\
\scriptsize{\textbf{A:} grey} &  \scriptsize{\textbf{A:} black} &  \scriptsize{\textbf{A:} yes}

\end{tabular}
\caption[Perceived color problems]{Demonstration of perceived color challenges. Each column corresponds to one example. For each example, the top image is the input image with markings of relevant results. The bottom image is a map of color names corresponding to the required object. Below the images, the question and corresponding answer are given. First column demonstrates classifications errors in the generated map of color names due to shading. Second column require ignoring the windows and wheels areas for an accurate answer. For the example of the third column, only specific area should be checked and colors should correspond to different names. [Object detection is based on faster R-CNN + DeepLab].}
\label{fig:colorProblems}
\end{centering}
\end{figure}

An interesting observation regarding the estimation of some visual elements is for the generation of color name maps \cite{van2007learning}, which is based on supervised learning (11 optional colors per pixel). When object colors are required, the map is generated for the object area in the image, and based on dominant colors the answer is provided. Retrieving object color may appear as a trivial task, as the intensity of original RGB image channels should provide the exact color of each pixel. However, using such methods fail to obtain the perceived color, as it is hardly related to levels of actual RGB channels. Hence, learning methods are incorporated to address this problem, and still there are many inaccuracies. In addition to these inaccuracies, the required process for obtaining perceived color of an object is not consistent. This can be seen in the examples of Figure \ref{fig:colorProblems}, where inquiring for the color of a person requires different color naming and focus on specific regions. The bus example also requires specific behavior, where the windows and wheels areas of the bus should be ignored.

As previously mentioned the parser sensitivity to phrasing and other issues such as its indifference to type of phrase coordinators ('and', 'or') causes representation failures or misrepresentations, which results with inability to provide a correct answer. For example when 'or' is used (e.g. ''Are the flowers yellow or white?'') the answer will be always 'no', as both options are required to be true. Hence, we get an answer which is irrelevant to the question.

Questions may be misinterpreted due to multiple meaning of words and phrases or subtle differences. As previously discussed this mainly effects the use of external knowledge database where a wide range of concepts may be used, which may lead to an unclear meaning of a concept (e.g. 'train'- vehicle vs. learn, 'monitor'- screen vs. supervise). Such confusions happen also for the question itself. An example for a misinterpreted question is ``What is the table/bus number?'', which is interpreted as ``What is the number of tables/buses?''

Currently, other than enhancing object detection by attention from question relations, details from the question are not used as hints for correctness of expressions. A case where such information may be further utilized is when the query is for a property of an object. In this case there may be a prior assumption or an increase in probability that such an object exists. Of course, an automatic assumption of existence is not desirable. However, reduction in classification thresholds, additional attempts using hints and other measures may be utilized to reflect the higher probability for the existence of such an object. For example, given the question ``What is the age of the man?'', the probability that a man indeed exist in the image should rise, and refuting this assumption should be performed only when the evidence is substantial.

\section{Discussion and Conclusions}

We have presented an approach to visual question answering that seeks to compose an answering procedure based on the 'abstract' structure of the query. We exploit the compositional nature of the question and represent it as a directed graph with objects represented as nodes and relations as edges. Each basic component of this graph representation is mapped to a dedicated basic procedure. The collection of these basic procedures are put together, along with additional required processes, into a complex procedure for the entire query.

This procedure incorporates query details and intermediate results and stores them in the graph nodes and a working memory module. The stored information completes the guidance to the procedure and allows handling different types of visual elements. Question relations are used as an attention source to enhance object detection. Querying for external common information is also handled by the procedure in order to complete the required prior knowledge needed to answer the question.

Breaking the answering process into basic meaningful components, corresponding to basic logic patterns, enables awareness at each step to the accomplished and unaccomplished parts of the task. This includes recognizing and reporting on failures and limitations, that in many cases are corrected and provided with valid alternatives. Elaborations to the answers are provided, according to the stored information. Since the building blocks include simple real world detectors, the system is modular and its improvement is not limited.

Human abilities motivate us to examine and handle some complicated attributes that are addressed naturally by humans, even though they may hardly appear in real queries. These attributes, such as 'odd man out', demonstrate representation challenges, that require extending the natural graph representation. Currently specific configuration is created to represent these attributes. Future upgrades may allow handling it more smoothly.

Evaluation of representation capabilities demonstrated that, even though potentially, our scheme can represent practically all queries, current state of the system is limited. The observed problems include limitations in vocabulary identification, sensitivity to phrasing and cases of grammatical similarity for different elements (e.g. 'wearing the same color' vs. 'wearing the same pants'). Additionally, some rare representation limitations exist, such as relations between more than two objects of different classes.

Even though the recognition abilities are currently limited due to scope of existing detectors, the system is self aware and mostly reply by specifying its limitation (which may trigger an addition of the desired detectors to the system). The representation limitations discussed in \ref{sec:representation} are a fundamental source of failures, which is added to incremental chances for errors of the used detectors. Our system does not exploit any language bias of the question. The answer is exclusively provided by the procedure evaluating the logic representation of the question. However, improvement is ongoing, as detectors keep improving and their scope keeps growing.

Current approaches to visual question answering use mostly end-to-end schemes that are very different than our approach. Although some methods include adaptive aspects, the optimization process is more likely to exploit language bias than the complex mechanisms required for proper answering. These methods maximize statistical results, but are likely to fail in addressing subtle, yet meaningful cases. This fits the analysis of current models, demonstrating the tendency to utilize only part of the question, provide same answers for different images and fail on novel forms. A combination of UnCoRd system and an end-to-end model may be beneficial in some cases, for example enhancing UnCoRd elaborations with "intuitive" answer in some cases (such as unknown visual elements).

We've integrated and examined various aspects of answering questions on images using our answering system. Much more research and investigation is required for all these aspects, as well as others. Future research will include learning the representation mapping and making it more robust, further investigating and improving the visual elements analyzers (e.g. combine the type of object when possible for property detection) and more.

\vskip 0.2in
\bibliography{ref}
\bibliographystyle{theapa}

\end{document}